\documentclass{article}

\PassOptionsToPackage{numbers, compress}{natbib}

 \usepackage[preprint]{neurips_2026}

\usepackage[utf8]{inputenc} %
\usepackage[T1]{fontenc}    %
\usepackage{hyperref}       %
\usepackage{url}            %
\usepackage{booktabs}       %
\usepackage{amsfonts}       %
\usepackage{nicefrac}       %
\usepackage{microtype}      %
\usepackage{xcolor}         %



\usepackage{graphicx}
\usepackage{subcaption}
\usepackage{booktabs,array}

\usepackage[dvipsnames]{xcolor}
\usepackage{amsmath, amssymb, mathtools, amsthm, arydshln, url, bbm, csquotes, multirow, balance, placeins, xspace}
\allowdisplaybreaks

\usepackage{hyperref}

\usepackage{comment}
\usepackage[capitalize,noabbrev,nameinlink]{cleveref}

\usepackage[textsize=tiny]{todonotes}

\usepackage{amsmath,amsfonts,bm}

\def\eqref#1{equation~\ref{#1}}

\def\1{\bm{1}}

\DeclareMathAlphabet{\mathsfit}{\encodingdefault}{\sfdefault}{m}{sl}
\SetMathAlphabet{\mathsfit}{bold}{\encodingdefault}{\sfdefault}{bx}{n}

\def\sN{{\mathbb{N}}}

\def\sR{{\mathbb{R}}}

\newcommand{\E}{\mathbb{E}}

\newcommand{\Var}{\mathrm{Var}}

\definecolor{TolMutedBlue}{HTML}{332288}
\definecolor{TolMutedGreen}{HTML}{117733}
\definecolor{TolMutedPurple}{HTML}{AA4499}

\definecolor{Paretogreen}{HTML}{4CAF50}
\definecolor{Streamlining}{HTML}{3A86FF}
\definecolor{CUP}{HTML}{F72585}
\definecolor{CVP}{HTML}{F72585}

\hypersetup{
    colorlinks=true,
    citecolor=TolMutedBlue,
    linkcolor=TolMutedGreen,
    urlcolor=TolMutedPurple,
}

\theoremstyle{plain}

\theoremstyle{definition}

\theoremstyle{remark}

\crefname{figure}{Fig.}{Figs.}
\Crefname{figure}{Figure}{Figures}
\crefname{table}{Tab.}{Tabs.}
\Crefname{table}{Table}{Tables}
\crefname{section}{Sec.}{Secs.}
\Crefname{section}{Section}{Sections}
\crefname{equation}{Eq.}{Eqs.}
\Crefname{equation}{Equation}{Equations}
\crefname{appendix}{App.}{App.}
\Crefname{appendix}{Appendix}{Appendix}

\newcommand{\myparagraph}[1]{\vspace{0.75em}\noindent\textbf{#1.}\hspace{0.75em}}

\newcommand{\cf}{\textit{cf.}\ }
\newcommand{\eg}{\textit{e.g.}\ }
\newcommand{\ie}{\textit{i.e.}\ }

\newcommand{\vpm}[2]{$#1{\scriptstyle\,\pm\,#2}$}
\newcommand{\bvpm}[2]{$\mathbf{#1}{\scriptstyle\,\pm\,#2}$}

\title{Calibrated Sampling-Free Uncertainty Estimation \\in Bayesian Deep Learning}

\author{%
  Tobias Jan Wieczorek\textsuperscript{1} \quad
  Leon de Andrade\textsuperscript{1} \quad
  Thomas Möllenhoff\textsuperscript{2} \quad
  Marcus Rohrbach\textsuperscript{1} \\[0.6em]
  \textsuperscript{1}TU Darmstadt \& hessian.AI, Darmstadt, Germany \\
  \textsuperscript{2}RIKEN Center for Advanced Intelligence Project, Tokyo, Japan \\[0.4em]
  \texttt{tobias.wieczorek@tu-darmstadt.de}
}
\begin{document}

\maketitle

\begin{abstract}
    Modern deep learning models remain notoriously prone to overconfidence, limiting their reliability in high-stakes applications. Bayesian methods aim to counter this by learning a distribution over model parameters, and recent advances now make this feasible for large-scale architectures at costs comparable to AdamW. However, a challenge remains at test time: predictions must be averaged across many forward passes with weights sampled from the posterior, which is prohibitively expensive. Variance propagation offers an efficient alternative, computing layer-wise analytical approximations of uncertainty in a single forward pass. While such techniques are effective for MLPs, their extension to modern architectures remains challenging, due to increased depth and diversity of layer types. To fill this gap, we propose Calibrated Variance Propagation (CVP), which introduces a new propagation method for normalization layers, combines it with recent techniques for handling activation functions, and absorbs residual error through a light calibration step. CVP yields comparably accurate uncertainty estimates to MC sampling across transformers and CNNs, at a fraction of the cost. Against prior variance propagation work, CVP improves coverage at 0.5\% risk from 8.2\% to 14.6\% with BEiT-3 on Visual Reasoning (NLVR2) and from 2.6\% to 10.8\% with ViLT on VQAv2, with gains extending to convolutional architectures.
\end{abstract}

\section{Introduction}\label{sec:intro}

Recent advances in deep learning based on the transformer architecture \cite{attention} have driven remarkable progress across language \cite{gpt2}, vision \cite{vit}, and multimodal understanding \cite{clip}, yet models still struggle to know when they are wrong. Rather than expressing uncertainty, even frontier models frequently exhibit confident hallucinations~\cite{why_llm_hallucinate}. This is a critical limitation for high-stakes applications such as disease detection~\cite{leibig2017}, legal reasoning~\cite{dahl2024legal}, and visual assistance for blind users \cite{vizwiz}. In such domains, models must know when to abstain, a capability formalized as \textit{selective prediction} \cite{chow1970seminal}. Equipping models with this skill remains an open problem, with scaling alone proving insufficient \cite{abstentionbench}.

\begin{figure}[!thbp]
    \centering
    \captionsetup[subfigure]{font=footnotesize}
    \begin{subfigure}[t]{0.32\textwidth}
        \centering
        \includegraphics[width=\textwidth]{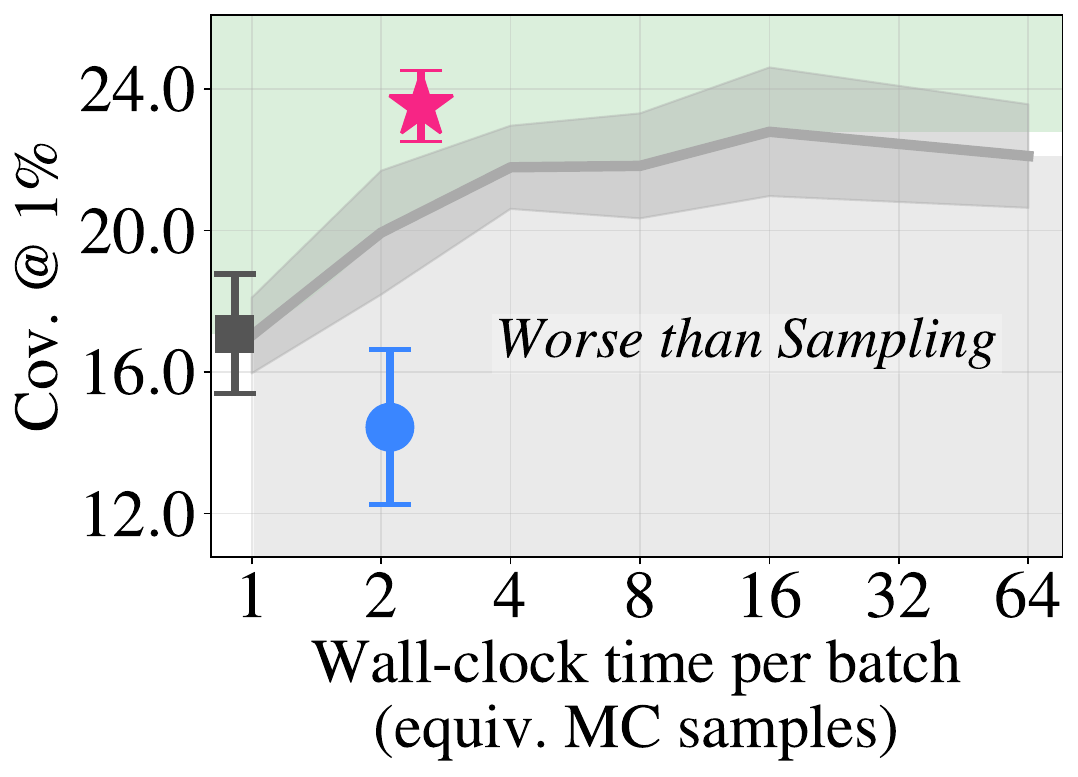}
        \caption{\textbf{Visual Reasoning} on NLVR2}
        \label{fig:teaser1}
    \end{subfigure}
    \hfill
    \begin{subfigure}[t]{0.32\textwidth}
        \centering
        \includegraphics[width=\textwidth]{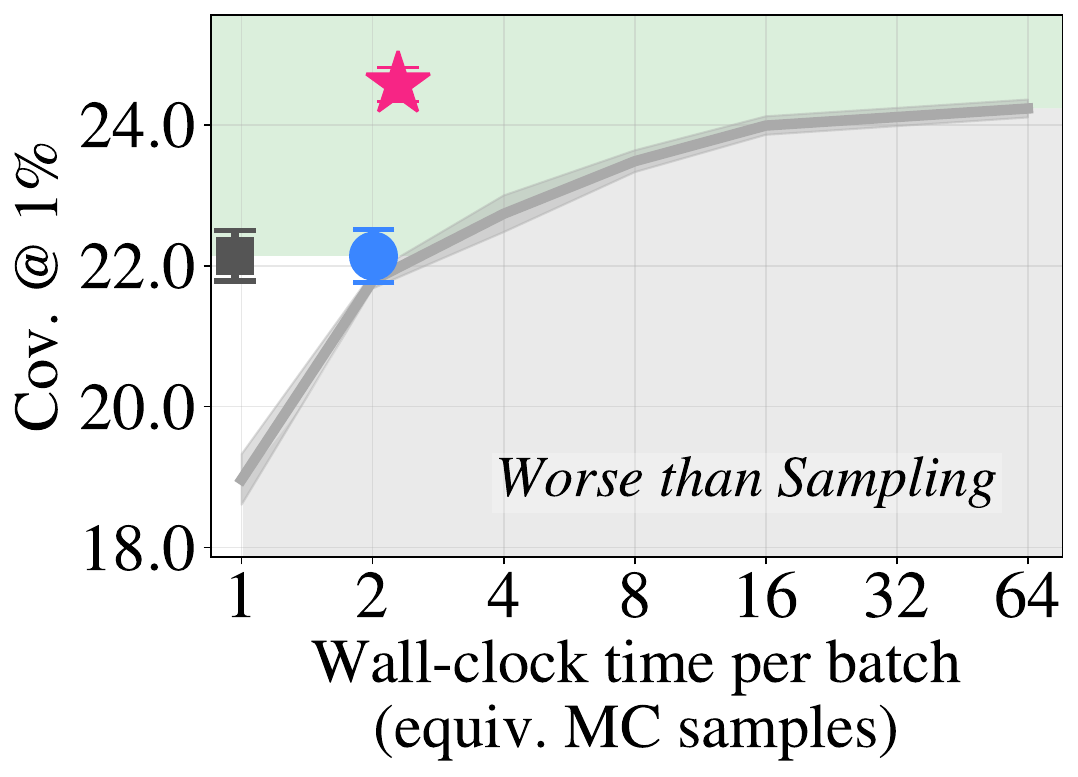}
        \caption{\textbf{VQA} on VQAv2}
        \label{fig:teaser2}
    \end{subfigure}
    \hfill
    \begin{subfigure}[t]{0.32\textwidth}
        \centering
        \includegraphics[width=\textwidth]{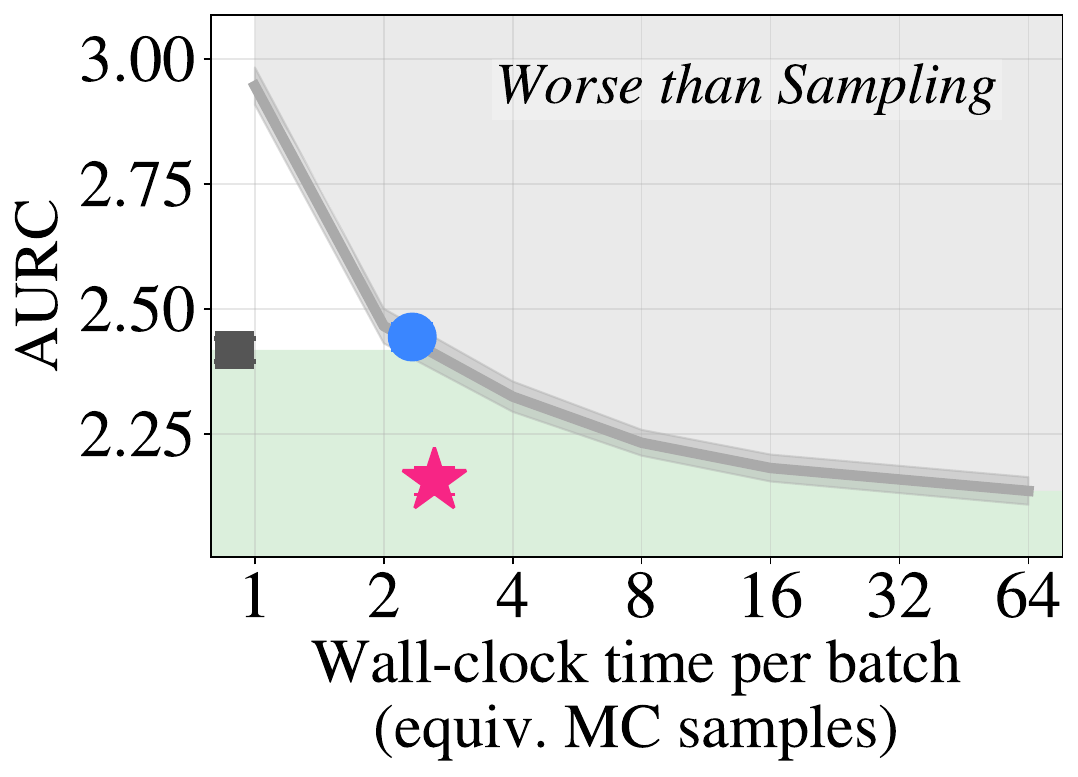}
        \caption{\textbf{Image Classification} on CIFAR 100}
        \label{fig:teaser3}
    \end{subfigure}
    \vspace{0.3em}
    \includegraphics[width=0.7\textwidth]{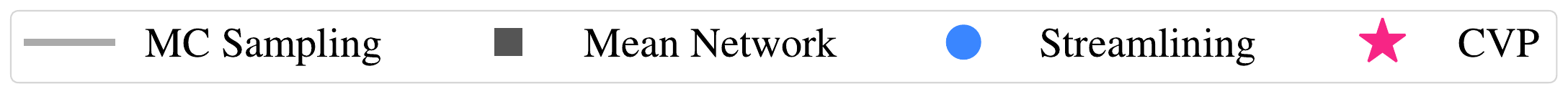}
    \caption{\textbf{CVP delivers consistent \colorbox{Paretogreen!20}{Pareto improvements} in selective prediction.} Variational learning yields a diagonal Gaussian posterior $\mathcal{N}(\mu,\sigma^2)$ over weights; the cheap \textit{mean network} sets each weight to its mean $\mu$ and discards the variance $\sigma^2$, while MC sampling draws weights from the posterior. Across vision and multimodal benchmarks, CVP Pareto-dominates both alternatives, while prior work (\textit{Streamlining} \cite{streamlining}) does not improve over the mean network despite added cost. For (a) and (b), the used model is BEiT-3 \cite{beit3}, for (c) it is ViT-B \cite{vit}.}
    \label{fig:teaser}
\end{figure}

Bayesian neural networks \cite{variational_nns,osawa2019practical} (BNNs) provide a natural route to uncertainty quantification, encoding parameter uncertainty directly into model weights using Bayesian / variational methods. Once impractical for modern architectures, this has been made viable by recent advances in variational learning \cite{ivon}, which yield a diagonal Gaussian posterior $\mathcal{N}(\mu,\sigma^2)$ over the weights at training costs on par with standard optimizers. The BNN bottleneck has thus shifted to test time: predicting with the \textit{mean network} (just using $\mu$) is cheap but ignores the variance ($\sigma^2$), while Monte-Carlo (MC) sampling requires prohibitively many forward passes.

Variance propagation offers an alternative that avoids repeated forward passes but still utilizes the learned posterior: layer-wise analytical rules propagate the mean and variance of intermediate activations through the model (\cf~\cref{fig:method_varprop}). To be useful in practice, however, it must Pareto-dominate the mean network and MC sampling on the efficiency-reliability tradeoff (\cf~\cref{fig:teaser}). Most existing variance propagation methods do not transfer to modern architectures, and the only recent work, \citet{streamlining} (hereafter \textit{Streamlining}), improves calibration but not selective prediction.

We address this gap by proposing \textit{Calibrated Variance Propagation} (CVP), a comprehensive variance propagation framework for modern architectures, including CNNs and transformers. CVP combines exact propagation rules for activation functions \cite{analytic_covar_prop,residual_varprop} with our new approximation for normalization layers such as LayerNorm \cite{layernorm}. To counter error accumulation with depth, we insert learnable scaling factors for the propagated variance at key layers, fitted in a lightweight calibration phase analogous to temperature scaling \cite{tempscaling}. CVP Pareto-dominates both the mean network and MC sampling on the efficiency-reliability tradeoff, and extends variance propgation to multimodal transformers for the first time. Our contributions are:

\begin{enumerate}
    \item \textbf{Variance propagation rule for normalization layers.} We develop a new variance propagation rule for normalization layers with substantially improved fidelity to MC sampling over linearization, and extensively validate it on synthetic data and model activations.
    \item \textbf{Per-layer calibration stage.} We introduce learnable per-layer scaling factors that correct systematic error accumulation, yielding improved uncertainty estimates in terms of calibration and particularly selective prediction.
    \item \textbf{Optimized implementation.} Our CVP implementation runs in a single pass at $\sim 2$-$3\times$ the cost of a deterministic forward pass across architectures, delivering substantial reliability gains without the repeated sampling of MC methods.
\end{enumerate}

\section{Related Work}\label{sec:related_work}

\myparagraph{Selective Prediction} The ability to abstain from uncertain predictions was first formalized by \citet{chow1970seminal} as a trade-off between coverage and risk on accepted examples - later works adapted it to deep learning \cite{el-yaniv2010foundations,geifman_selpred}. The quality of the abstention rule depends directly on the underlying confidence scores, thus reliable confidence estimates are central to selective prediction. However, deterministic networks are systematically overconfident and poorly calibrated \cite{tempscaling}, a problem that \citet{abstentionbench} show persists even in foundation models and cannot be solved by scaling alone. Post-hoc calibration techniques such as temperature scaling \cite{tempscaling} address miscalibration but help little with abstention \cite{reliable_vqa}. This motivates principled alternatives in the form of variational Bayesian methods \cite{variational_nns}, which encode input-dependent uncertainty through distributions over weights. Recently, such methods have been shown to deliver substantial gains in selective prediction for both visual \cite{varvqa} and audio question answering \cite{varaqa}. CVP builds on this line of work by targeting the test-time bottleneck of expensive MC sampling.

\myparagraph{Bayesian Deep Learning} Within variational Bayesian learning, early methods optimized weight distributions via SGD \cite{variational_nns,bayes_by_backprop}, but struggled to scale to modern architectures, with reported failure modes including posterior over-pruning \cite{trippe2018overpruning} and the limited expressiveness of mean-field families~\cite{foong2020expressiveness}. Practical alternatives have therefore dominated for years: deep ensembles \cite{deep_ensembles} train multiple independent networks, but their cost grows linearly with the ensemble size and is prohibitive for the latest large architectures; MC Dropout \cite{gal2016dropout} reinterprets dropout as approximate inference at no training overhead, and SWAG \cite{swag} fits a Gaussian to the SGD trajectory post-hoc. Recent natural-gradient methods have revived scalable Bayesian deep learning at a cost comparable to standard optimizers: the variational IVON optimizer \cite{ivon} in particular matches AdamW in accuracy while producing well-calibrated Gaussian posteriors at nearly identical training cost. IVON has been shown to empirically outperform both MC Dropout and SWAG in predictive uncertainty \cite{ivon}. While CVP is agnostic to the underlying posterior, our experiments focus on IVON, whose variances are learned end-to-end with the network and thus yield a posterior amenable to sampling and propagation.

\myparagraph{Variance Propagation} Propagating distributions analytically through neural networks dates back to \citet{frey1999}, who derived closed-form propagation rules for Gaussian inputs. Subsequent work has followed two main directions. The first propagates \emph{input} uncertainty through fixed-weight networks, with rules for diagonal Gaussians \cite{gast2018}, Gaussian mixtures \cite{split_and_merge,input_uncertainty_prop}, and general stable distributions \cite{stable_dist_prop}. CVP belongs to the second direction, which propagates Bayesian \emph{parameter} uncertainty as an alternative to MC sampling and treats input uncertainty as a special case. Within this direction, early methods used propagated uncertainty to perform approximate inference during training \cite{probabilistic_backprop,deterministic_varinf}; \citet{postels2019_dropoutprop} instead applied it at test time, replacing the sampling used for Dropout uncertainty. More recent contributions refine the per-layer approximations, including a general covariance rule for nonlinear activations \cite{analytic_covar_prop}, analytic approximations for softmax \cite{laplace_bridge,mean_field_softmax}, and exact moment matching for residual blocks in MLPs \cite{residual_varprop}. All these methods, however, target MLP-like models and do not handle the normalization layers central to modern architectures. Streamlining \cite{streamlining}, the only prior variance propagation method targeting large-scale models, linearizes LayerNorm and the activation functions, which leaves propagated activations identical to those of the mean network until the final layer (\cf~\cref{fig:method_cup}). We show that this leaves Streamlining no better than the mean network in selective prediction, despite its added cost. CVP closes this gap, delivering the first variance propagation framework that Pareto-dominates both the mean network and MC sampling on ResNets and transformers, and the first to extend variance propagation to multimodal settings.

\section{Preliminaries}\label{sec:background}

In Bayesian deep learning, predictions are obtained by integrating over the weights posterior $p(\theta \mid \mathcal{D})$:
\begin{equation}\label{eq:method_predictive}
    p(y \mid x, \mathcal{D}) = \int p(y|\theta,x)\, p(\theta \mid \mathcal{D})\, d\theta.
\end{equation}
The exact posterior $p(\theta \mid \mathcal{D}) \propto p(\mathcal{D} \mid \theta)\, p(\theta)$ is intractable for deep networks, as is the integral in~\cref{eq:method_predictive}. Two layers of approximation are therefore standard. First, the posterior is replaced during training by a variational approximation $q(\theta) = \mathcal{N}(\theta \mid \mu, \Sigma)$~\cite{variational_nns}. Second, the integral over $\theta$ is approximated at test time. The two standard choices are Monte Carlo (MC) sampling,
\begin{equation}\label{eq:method_mc}
    p(y \mid x, \mathcal{D}) \approx \frac{1}{M}\sum_{m=1}^{M} f_{\theta_m}(x), \qquad \theta_m \sim q(\theta),
\end{equation}
which needs $M$ forward passes, and the mean network $f_{\mu}$, which needs only one pass, but discards $\Sigma$.

\subsection{Variance Propagation}

\begin{figure}[t]
    \centering
    \includegraphics[width=0.9\textwidth]{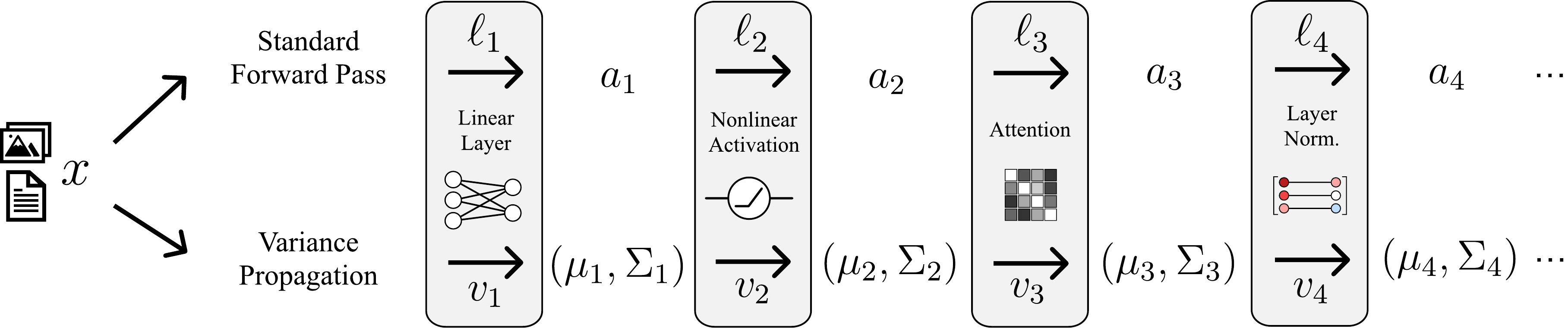}
    \caption{\textbf{Variance propagation as a chain of operators.} Each $v_i$ mirrors a layer $f_i$ of the original network, but instead of activations $a_i$, their means $\mu_i$ and variances $\Sigma_i$ are propagated. The shown sequence of layers is exemplaric.}
    \label{fig:method_varprop}
\end{figure}

Variance propagation sits between these two options by propagating means and variances analytically through the network in a single forward pass (\cref{fig:method_varprop}). The network $f_\theta = \ell_n^\theta \circ \cdots \circ \ell_1^\theta$ is replaced by a chain of operators $(v_n^\theta \circ \cdots \circ v_1^\theta)$, which operate on mean-variance pairs:

\begin{equation}\label{eq:method_chain}
    v_i: (\mu_i, \Sigma_i) \mapsto (\mu_{i+1}, \Sigma_{i+1}).
\end{equation}

\myparagraph{Diagonal covariances and independence} For tractability, we assume independence between $\theta$ and the activations $a$ at every layer (and between $\theta$ and the network input $x$), which is standard in the literature \cite{gast2018,deterministic_varinf}. We also propagate diagonal covariances only: for transformers, storing per-token full covariances for an activation tensor of $L$ tokens and model dimension $d$ would already require memory of $\mathcal{O}(L d^2)$, and cross-token covariances $\mathcal{O}(L^2 d^2)$ - a $10^3\times$ to $10^5\times$ memory increase over the $\mathcal{O}(L d)$ of the deterministic forward pass at typical\footnote{For example ViT-Base, where $D=768$ and $224 \times 224$ image size, giving $L=196$ patches at $16\times 16$ patch size.} $L, d$. In the following, we use $x$ and $y$ for the input and output of the layer under discussion, with moments $(\mu_x, \Sigma_x)$ and $(\mu_y, \Sigma_y)$.

\subsection{Existing Per-Layer Propagation Rules}

A complete variance propagation framework for transformers requires a rule for each component of the architecture: linear layers, element-wise activations, residual connections, multi-head attention, normalization layers, and the output softmax. Below, we review existing rules for all components except the normalization layers, which have so far only been addressed by linearization.

\myparagraph{Linear Layers} For a linear layer $y = Wx + b$ with diagonal Gaussians on $W$, $b$, and $x$, the output mean and variance follow directly from elementary identities (full derivation in \cref{sec:app_deriv_linearlayer}):
\begin{equation}\label{eq:method_linear}
    \mu_y = \mu_W \mu_x + \mu_b, \qquad
    \Sigma_y = (\Sigma_W + \mu_W^2)\, \Sigma_x + \Sigma_W\, \mu_x^2 + \Sigma_b.
\end{equation}
The same rule extends directly to convolutional layers \cite{gast2018}.

\myparagraph{Activation Functions} For element-wise activations $g$, the standard approach is the Delta method, which performs a first-order Taylor linearization of $g$ around $\mu_x$:
\begin{equation}\label{eq:method_delta}
    \mu_y = g(\mu_x), \qquad \Sigma_y = g'(\mu_x)^2\, \Sigma_x.
\end{equation}
It applies to any differentiable activation but degrades for inputs with large variance, particularly near regions of high curvature in $g$. Crucially, $\mu_y = g(\mu_x)$ does not depend on $\Sigma_x$, so the Delta method leaves the propagated mean unchanged from the mean network's.

\myparagraph{Residual Connections} For residual blocks $y = x + h(x)$, one typically assumes independence between the residual branch output $h(x)$ and the skip connection $x$, yielding
\begin{equation}
    \mu_y = \mu_x + \mu_{h(x)}, \qquad \Sigma_y = \Sigma_x + \Sigma_{h(x)}.
\end{equation}
If $h(x)$ is a linear layer, an exact rule can be derived \cite{residual_varprop}.

\begin{figure}[!th]
    \centering
    \includegraphics[width=\textwidth]{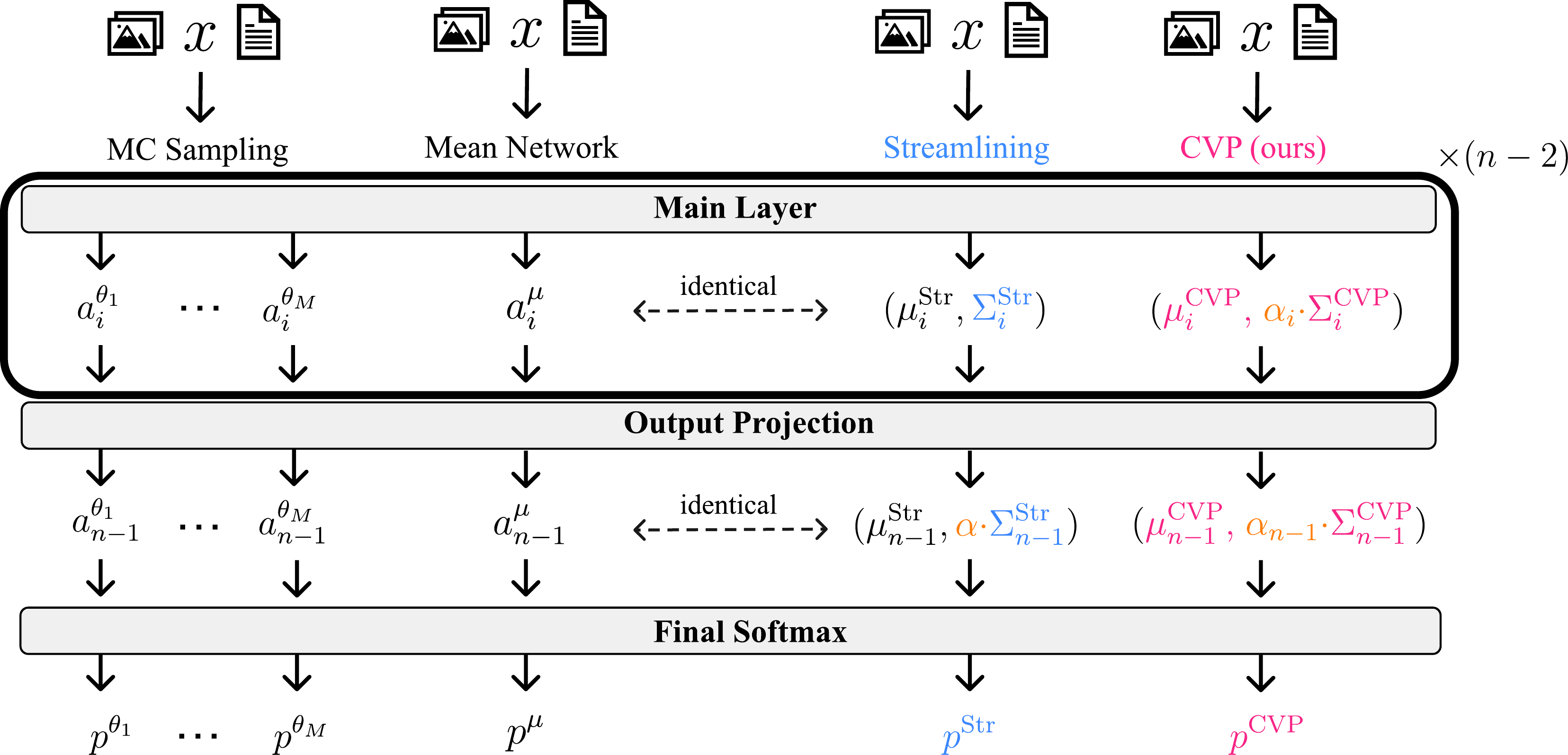}
    \caption{\textbf{CVP vs. Streamlining \cite{streamlining}.} Due to per-layer linearizations, Streamlining's propagated mean activation is identical to that of the mean network until the final layer ($\mu_i^\textrm{Str}=a_i^{\mu}$). CVP deviates from the mean network early on ($\mu_i^\textrm{CVP}\neq a_i^{\mu}$) and calibrates propagated variances with scaling factors $\alpha_i$, while Streamlining only calibrates at the logit level. Note that in practice, scaling factors are only inserted before key layers (\cf \cref{sec:method_calib}).}
    \label{fig:method_cup}
\end{figure}

\myparagraph{Attention} For multi-head self-attention $\mathrm{Attention}(Q,K,V) = \mathrm{softmax}(QK^\top / \sqrt{d_k})\, V$, the queries and keys both depend on the same input $x$ via $Q = x W_Q$ and $K = x W_K$, so the linear-layer rule does not apply: $QK^\top = x W_Q W_K^\top x^\top$ is quadratic in $x$. Streamlining \cite{streamlining} has addressed this by passing only the means through the query and key projections, treating the attention map
\begin{equation}\label{eq:method_attention_map}
    A = \mathrm{softmax}\!\left(\frac{\mu_x W_Q W_K^\top \mu_x^\top}{\sqrt{d_k}}\right)
\end{equation}
as deterministic. Variance is reintroduced at the value projection $V = x W_V$, where \Cref{eq:method_linear} gives moments $(\mu_V, \Sigma_V)$. The attention output $y = A V$ then satisfies $\mu_y = A\, \mu_V$ and  $\Sigma_y = (A \odot A)\, \Sigma_V$.

\myparagraph{Output softmax / sigmoid} The final softmax / sigmoid\footnote{Sigmoid is used \eg in the case of multi-label classification, such as VQA.} acts on a single logit vector. In contrast to the attention softmax, which operates on $L\times L$ matrices at every layer and head, MC sampling is tractable for the last-layer softmax due to the lower dimensionality. In practice, one draws $M$ samples from the propagated Gaussian over logits and averages the resulting softmax probabilities, which adds negligible overhead.

\section{Calibrated Variance Propagation}\label{sec:cvp}

At every layer, Streamlining's rules leave the propagated mean equal to the mean network's (\cref{fig:method_cup}): the Delta method gives $\mu_y = g(\mu_x)$ for activations, and the LayerNorm linearization gives $\mu_y = \mathrm{LayerNorm}(\mu_x)$. Only the variance reflects parameter uncertainty, which Streamlining rescales with a single scalar before the output softmax. This improves calibration but not selective prediction, since a uniform rescaling preserves the relative ordering of samples by uncertainty (\cref{fig:teaser}).

CVP differs from Streamlining in two per-layer choices: it replaces the Delta method with exact moments for nonlinear activations (\cref{sec:method_actfunc}) and the linearization of normalization layers with an expectation-based rule (\cref{sec:method_ln}). Both make the propagated mean depend on $\Sigma_x$ at every layer, so CVP's predictions diverge from the mean network's. A lightweight calibration step (\cref{sec:method_calib}) absorbs the approximation error that accumulates across layers.

\begin{table*}[ht]
  \centering
  \renewcommand{\arraystretch}{1.3}
  \setlength{\tabcolsep}{3.5pt}
  \footnotesize
  \caption{\textbf{CVP delivers the best uncertainty quality among methods of similar runtime on image classification.} We report mean and standard error (SEM) over 10 (ViT) and 5 (ResNet) seeds; bold marks the best per metric (before rounding). Wall-clock time is relative to a single MC sample.}
  \label{tab:default:vit_test_sem}
  \begin{tabular}{l c r r r r r r r}
    \toprule
    \multirow{2}{*}{Method} & Rel. & \multicolumn{1}{c}{\multirow{2}{*}{Acc.}} & \multicolumn{1}{c}{\multirow{2}{*}{AURC ($\downarrow$)}} & \multicolumn{1}{c}{\multirow{2}{*}{C@$\tfrac{1}{2}$\%}} & \multicolumn{1}{c}{\multirow{2}{*}{C@1\%}} & \multicolumn{1}{c}{\multirow{2}{*}{NLL ($\downarrow$)}} & \multicolumn{1}{c}{\multirow{2}{*}{ECE ($\downarrow$)}} & \multicolumn{1}{c}{\multirow{2}{*}{Brier ($\downarrow$)}} \\
     & Time &  &  &  &  &  &  &  \\
    \midrule
    \multicolumn{9}{l}{\textit{\textbf{ImageNet (ResNet-50)}}} \\
    Mean Network & 1.0$\times$ & \vpm{77.2}{0.0} & \vpm{6.26}{0.01} & \vpm{7.1}{1.0} & \vpm{23.7}{0.2} & \vpm{0.89}{0.00} & \vpm{3.48}{0.04} & \vpm{0.32}{0.00} \\
    \quad + Temp. Sc. & 1.0$\times$ & \vpm{77.2}{0.0} & \vpm{6.28}{0.01} & \vpm{10.2}{0.5} & \vpm{23.5}{0.2} & \vpm{0.88}{0.00} & \vpm{1.34}{0.02} & \vpm{0.32}{0.00} \\
    \hdashline
    2 samples & 2.0$\times$ & \vpm{76.4}{0.0} & \vpm{6.69}{0.01} & \vpm{9.6}{0.5} & \vpm{22.2}{0.2} & \vpm{0.91}{0.00} & \bvpm{1.18}{0.04} & \vpm{0.33}{0.00} \\
    4 samples & 4.0$\times$ & \vpm{76.9}{0.0} & \vpm{6.49}{0.01} & \vpm{10.3}{0.5} & \vpm{23.0}{0.2} & \vpm{0.89}{0.00} & \vpm{1.60}{0.03} & \vpm{0.32}{0.00} \\
    \hdashline
    \textcolor{Streamlining}{Streamlining} & 2.6$\times$ & \vpm{77.2}{0.0} & \vpm{6.28}{0.01} & \vpm{10.1}{0.5} & \vpm{23.3}{0.2} & \vpm{0.88}{0.00} & \vpm{1.43}{0.02} & \vpm{0.32}{0.00} \\
    \hdashline
    \textcolor{CVP}{CVP} & 2.9$\times$ & \bvpm{77.5}{0.0} & \bvpm{6.18}{0.00} & \bvpm{10.8}{0.5} & \bvpm{23.8}{0.3} & \bvpm{0.86}{0.00} & \vpm{1.40}{0.05} & \bvpm{0.31}{0.00} \\
    \midrule
    \multicolumn{9}{l}{\textit{\textbf{CIFAR-100 (ViT-Base)}}} \\
    Mean Network & 0.9$\times$ & \vpm{87.2}{0.1} & \vpm{2.42}{0.02} & \vpm{39.2}{1.0} & \vpm{52.8}{0.5} & \vpm{0.49}{0.00} & \vpm{4.64}{0.06} & \vpm{0.19}{0.00} \\
    \quad + Temp. Sc. & 0.9$\times$ & \vpm{87.2}{0.1} & \vpm{2.46}{0.02} & \vpm{37.4}{1.0} & \vpm{51.9}{0.5} & \vpm{0.47}{0.00} & \vpm{2.80}{0.05} & \vpm{0.19}{0.00} \\
    \hdashline
    2 samples & 2.0$\times$ & \vpm{87.0}{0.1} & \vpm{2.47}{0.03} & \vpm{43.4}{1.2} & \vpm{54.2}{0.7} & \vpm{0.49}{0.00} & \vpm{2.69}{0.07} & \vpm{0.19}{0.00} \\
    4 samples & 4.0$\times$ & \vpm{87.5}{0.1} & \vpm{2.32}{0.03} & \vpm{44.9}{0.7} & \vpm{56.0}{0.5} & \vpm{0.46}{0.00} & \vpm{1.99}{0.06} & \vpm{0.18}{0.00} \\
    \hdashline
    \textcolor{Streamlining}{Streamlining} & 2.3$\times$ & \vpm{87.2}{0.1} & \vpm{2.48}{0.02} & \vpm{36.7}{1.7} & \vpm{51.3}{0.5} & \vpm{0.48}{0.00} & \vpm{3.01}{0.07} & \vpm{0.19}{0.00} \\
    \hdashline
    \textcolor{CVP}{CVP} & 2.6$\times$ & \bvpm{88.1}{0.1} & \bvpm{2.16}{0.03} & \bvpm{45.1}{1.2} & \bvpm{56.9}{0.7} & \bvpm{0.42}{0.00} & \bvpm{1.91}{0.08} & \bvpm{0.18}{0.00} \\
    \midrule
    \multicolumn{9}{l}{\textit{\textbf{CIFAR-10 (sub-tiny ViT)}}} \\
    Mean Network & 0.9$\times$ & \vpm{79.9}{0.1} & \vpm{5.84}{0.04} & \vpm{16.8}{1.2} & \vpm{26.6}{0.6} & \vpm{0.63}{0.00} & \vpm{6.92}{0.10} & \vpm{0.29}{0.00} \\
    \quad + Temp. Sc. & 0.9$\times$ & \vpm{79.9}{0.1} & \vpm{5.78}{0.04} & \vpm{16.7}{1.0} & \vpm{27.7}{0.7} & \vpm{0.58}{0.00} & \vpm{1.11}{0.06} & \vpm{0.28}{0.00} \\
    \hdashline
    2 samples & 2.0$\times$ & \vpm{78.3}{0.1} & \vpm{6.70}{0.02} & \vpm{15.3}{1.3} & \vpm{22.6}{1.0} & \vpm{0.64}{0.00} & \vpm{2.79}{0.12} & \vpm{0.31}{0.00} \\
    4 samples & 4.0$\times$ & \vpm{79.3}{0.1} & \vpm{6.14}{0.03} & \vpm{16.1}{1.3} & \vpm{25.2}{0.7} & \vpm{0.61}{0.00} & \vpm{1.21}{0.06} & \vpm{0.29}{0.00} \\
    \hdashline
    \textcolor{Streamlining}{Streamlining} & 2.0$\times$ & \vpm{79.8}{0.1} & \vpm{5.71}{0.04} & \vpm{19.9}{0.9} & \vpm{28.9}{0.4} & \vpm{0.58}{0.00} & \vpm{1.06}{0.07} & \vpm{0.28}{0.00} \\
    \hdashline
    \textcolor{CVP}{CVP} & 2.2$\times$ & \bvpm{80.3}{0.1} & \bvpm{5.52}{0.03} & \bvpm{20.2}{0.8} & \bvpm{28.9}{0.6} & \bvpm{0.56}{0.00} & \bvpm{0.91}{0.06} & \bvpm{0.28}{0.00} \\
    \bottomrule
  \end{tabular}
\end{table*}

\subsection{Exact activation moments}\label{sec:method_actfunc}

For element-wise activations $g$, CVP replaces the Delta method (\cref{eq:method_delta}) with the exact moments
\begin{equation}\label{eq:method_exact_act}
    \mu_y = \mathbb{E}[g(x)], \qquad \Sigma_y = \mathrm{Var}[g(x)],
\end{equation}
under the assumption $x \sim \mathcal{N}(\mu_x, \mathrm{diag}(\Sigma_x))$. For the activations used in the architectures we evaluate, exact moments are available: GELU \cite{residual_varprop}, sigmoid \cite{bishop, sigmoid_moment_matching}, tanh (via $\tanh(x) = 2\sigma(2x) - 1$), and ReLU \cite{frey1999}. Full expressions are in \cref{sec:app_deriv_actfunc}. Unlike the Delta method, exact moments make $\mu_y$ depend on $\Sigma_x$, so the propagated mean diverges from the mean network's at every nonlinearity.

\begin{table*}[ht]
  \centering
  \renewcommand{\arraystretch}{1.25}
  \setlength{\tabcolsep}{3pt}
  \footnotesize
  \caption{\textbf{CVP also gives the best uncertainty estimates among methods of similar runtime in the multimodal domain.} We report mean SEM over 5 seeds; bold marks the best per metric (before rounding). Wall-clock time is relative to a single MC sample.}
  \label{tab:default:multimodal_test_sem}
  \begin{tabular}{l c r r r r r r r}
    \toprule
    \multirow{2}{*}{Method} & Rel. & \multicolumn{1}{c}{\multirow{2}{*}{Acc.}} & \multicolumn{1}{c}{\multirow{2}{*}{AURC ($\downarrow$)}} & \multicolumn{1}{c}{\multirow{2}{*}{C@$\tfrac{1}{2}$\%}} & \multicolumn{1}{c}{\multirow{2}{*}{C@1\%}} & \multicolumn{1}{c}{\multirow{2}{*}{NLL / BCE ($\downarrow$)}} & \multicolumn{1}{c}{\multirow{2}{*}{ECE ($\downarrow$)}} & \multicolumn{1}{c}{\multirow{2}{*}{Brier ($\downarrow$)}} \\
     & Time &  &  &  &  &  &  &  \\
    \midrule
    \multicolumn{9}{l}{\textit{\textbf{VQAv2 (BEiT-3-base)}}} \\
    Mean Network & 1.0$\times$ & \vpm{73.8}{0.0} & \vpm{7.67}{0.02} & \vpm{13.8}{0.4} & \vpm{22.1}{0.4} & \vpm{2.72}{0.00} & \vpm{3.83}{0.10} & \vpm{0.40}{0.00} \\
    \quad + Temp. Sc. & 1.0$\times$ & \vpm{73.8}{0.0} & \vpm{7.67}{0.02} & \vpm{13.8}{0.4} & \vpm{22.1}{0.4} & \vpm{2.72}{0.00} & \vpm{3.74}{0.11} & \vpm{0.40}{0.00} \\
    \hdashline
    2 samples & 2.0$\times$ & \vpm{73.3}{0.0} & \vpm{7.94}{0.02} & \vpm{13.7}{0.3} & \vpm{21.8}{0.1} & \vpm{2.74}{0.00} & \vpm{2.53}{0.07} & \vpm{0.40}{0.00} \\
    4 samples & 4.0$\times$ & \vpm{73.5}{0.1} & \vpm{7.76}{0.02} & \vpm{15.6}{0.2} & \vpm{22.7}{0.3} & \vpm{2.71}{0.00} & \bvpm{2.15}{0.06} & \vpm{0.40}{0.00} \\
    \hdashline
    \textcolor{Streamlining}{Streamlining} & 2.0$\times$ & \vpm{73.8}{0.0} & \vpm{7.67}{0.02} & \vpm{13.8}{0.4} & \vpm{22.1}{0.4} & \vpm{2.72}{0.00} & \vpm{3.79}{0.10} & \vpm{0.40}{0.00} \\
    \hdashline
    \textcolor{CVP}{CVP} & 2.3$\times$ & \bvpm{73.8}{0.1} & \bvpm{7.57}{0.03} & \bvpm{17.3}{0.3} & \bvpm{24.6}{0.2} & \bvpm{2.70}{0.00} & \vpm{3.17}{0.08} & \bvpm{0.40}{0.00} \\
    \midrule
    \multicolumn{9}{l}{\textit{\textbf{NLVR2 (BEiT-3-base)}}} \\
    Mean Network & 0.9$\times$ & \vpm{83.4}{0.1} & \vpm{5.61}{0.13} & \vpm{8.5}{2.7} & \vpm{17.1}{1.7} & \vpm{0.41}{0.01} & \vpm{5.92}{0.68} & \vpm{0.24}{0.00} \\
    \quad + Temp. Sc. & 0.9$\times$ & \vpm{83.4}{0.1} & \vpm{5.61}{0.13} & \vpm{8.5}{2.7} & \vpm{17.1}{1.7} & \vpm{0.36}{0.00} & \bvpm{1.21}{0.04} & \vpm{0.23}{0.00} \\
    \hdashline
    2 samples & 2.0$\times$ & \vpm{82.9}{0.2} & \vpm{5.78}{0.17} & \vpm{10.7}{2.1} & \vpm{19.9}{1.8} & \vpm{0.40}{0.01} & \vpm{4.48}{0.65} & \vpm{0.24}{0.00} \\
    4 samples & 4.0$\times$ & \vpm{83.1}{0.3} & \vpm{5.55}{0.14} & \vpm{14.0}{2.6} & \vpm{21.8}{1.2} & \vpm{0.39}{0.01} & \vpm{3.93}{0.57} & \vpm{0.24}{0.00} \\
    \hdashline
    \textcolor{Streamlining}{Streamlining} & 2.1$\times$ & \vpm{83.4}{0.1} & \vpm{5.66}{0.14} & \vpm{8.2}{2.2} & \vpm{14.4}{2.2} & \vpm{0.37}{0.00} & \vpm{1.38}{0.11} & \vpm{0.23}{0.00} \\
    \hdashline
    \textcolor{CVP}{CVP} & 2.5$\times$ & \bvpm{83.8}{0.2} & \bvpm{5.24}{0.10} & \bvpm{14.6}{2.6} & \bvpm{23.5}{1.0} & \bvpm{0.36}{0.00} & \vpm{1.27}{0.19} & \bvpm{0.23}{0.00} \\
    \midrule
    \multicolumn{9}{l}{\textit{\textbf{VQAv2 (ViLT)}}} \\
    Mean Network & 1.0$\times$ & \vpm{69.2}{0.1} & \vpm{10.53}{0.03} & \vpm{2.4}{1.5} & \vpm{13.3}{0.5} & \vpm{3.10}{0.00} & \vpm{7.72}{0.29} & \vpm{0.48}{0.00} \\
    \quad + Temp. Sc. & 1.0$\times$ & \vpm{69.2}{0.1} & \vpm{10.53}{0.03} & \vpm{2.5}{1.4} & \vpm{13.3}{0.5} & \vpm{3.10}{0.00} & \vpm{7.68}{0.29} & \vpm{0.48}{0.00} \\
    \hdashline
    2 samples & 2.0$\times$ & \vpm{68.0}{0.0} & \vpm{11.43}{0.03} & \vpm{7.6}{0.3} & \vpm{13.2}{0.4} & \vpm{3.13}{0.00} & \vpm{2.90}{0.03} & \vpm{0.48}{0.00} \\
    4 samples & 4.0$\times$ & \vpm{68.6}{0.1} & \vpm{10.87}{0.03} & \vpm{10.5}{0.4} & \vpm{15.7}{0.2} & \vpm{3.07}{0.00} & \bvpm{2.06}{0.03} & \vpm{0.47}{0.00} \\
    \hdashline
    \textcolor{Streamlining}{Streamlining} & 2.0$\times$ & \vpm{69.2}{0.1} & \vpm{10.53}{0.03} & \vpm{2.6}{1.5} & \vpm{13.3}{0.5} & \vpm{3.10}{0.00} & \vpm{7.70}{0.29} & \vpm{0.48}{0.00} \\
    \hdashline
    \textcolor{CVP}{CVP} & 2.2$\times$ & \bvpm{69.3}{0.1} & \bvpm{10.30}{0.06} & \bvpm{10.8}{0.2} & \bvpm{17.4}{0.2} & \bvpm{3.02}{0.00} & \vpm{3.69}{0.13} & \bvpm{0.46}{0.00} \\
    \bottomrule
  \end{tabular}
\end{table*}

\subsection{An expectation-based rule for normalization layers}\label{sec:method_ln}

In the following, LayerNorm is the running example, but the technique also translates directly to \eg BatchNorm \cite{batchnorm} and FRN \cite{frn}. Let $D$ be the model dimension, then LayerNorm acts as

\begin{equation}\label{eq:method_layernorm}
    y = \frac{x - m(x)}{\sqrt{s^2(x) + \epsilon}} \odot \gamma + \beta, \quad m(x) = \frac{1}{D}\sum_{i=1}^D x_i, \quad s^2(x) = \frac{1}{D}\sum_{i=1}^D \bigl(x_i - m(x)\bigr)^2,
\end{equation}

Streamlining \cite{streamlining} replaces $s^2(x)$ with $s^2(\mu_x)$, which reduces the propagated mean to $\mathrm{LayerNorm}(\mu_x)$. We instead replace $s^2(x)$ with its expectation under the input distribution:

\begin{equation}\label{eq:method_layernorm_s2}
    \E[s^2(x)] = s^2(\mu_x) + \left(1 - \tfrac{1}{D}\right)\bar{\Sigma}_x,
\end{equation}

where $\bar{\Sigma}_x = \tfrac{1}{D}\sum_i \Sigma_{x,i}$. This renders LayerNorm affine in $x$, so its moments follow from the linear rule  in \cref{eq:method_linear}. Full derivations are in \cref{sec:app_deriv_layernorm}, and we validate the approximation in \cref{sec:exp_layernorm}.

\subsection{Per-layer calibration}\label{sec:method_calib}
Per-layer approximation errors, primarily from the diagonal-covariance and Gaussian-input assumptions, compound with depth. We absorb them with scalars $\alpha_i > 0$ that multiplicatively rescale the propagated variance after key layers (\cf \Cref{fig:method_cup}). Optimal placement of these scalars remains somewhat open, but we found it best to insert them before \emph{mean-shifting} layers, where the output mean depends on the input variance, namely activation functions and normalization layers. The $\alpha_i$ are fitted by minimizing NLL on a held-out set with all other parameters frozen \cite{tempscaling, streamlining}. %

\section{Experiments}\label{sec:experiments}

We evaluate CVP on transformers and ResNets trained via variational learning. Across uni- and multimodal settings, CVP consistently Pareto-improves over the mean network and MC sampling on selective prediction and calibration at small constant overhead, while Streamlining does not.

\begin{figure}[!thbp]
    \centering
    \begin{subfigure}[t]{0.9\textwidth}
        \centering
        \includegraphics[width=\textwidth]{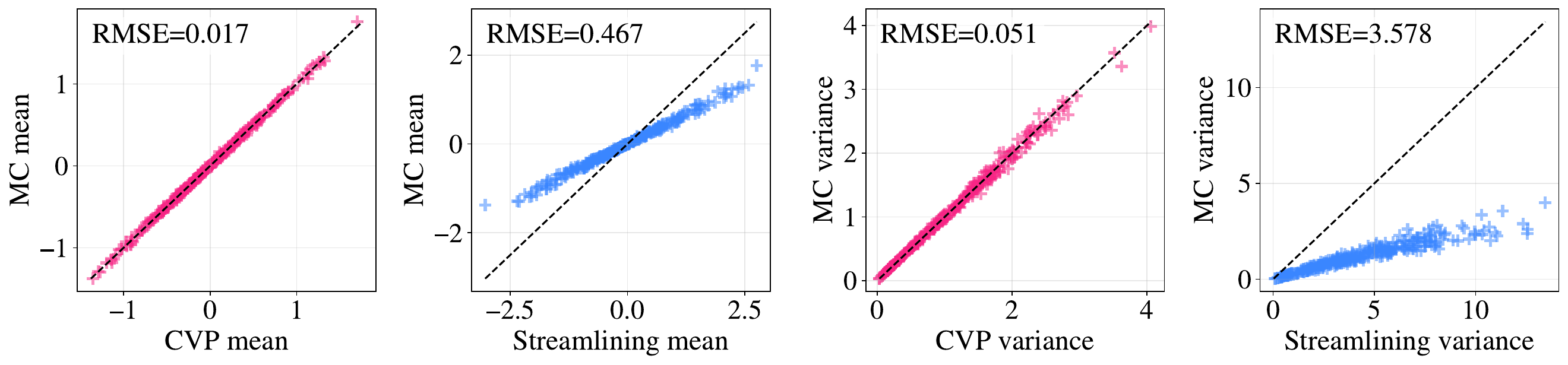}
        \caption{Synthetic Data with $\mu_x \sim \mathcal{N}(0, I)$ and $\Sigma_x\sim\mathcal{U}(0, 5)$.}
        \label{fig:layernorm_synthetic}
    \end{subfigure}
    \hfill
    \begin{subfigure}[t]{0.9\textwidth}
        \centering
        \includegraphics[width=\textwidth]{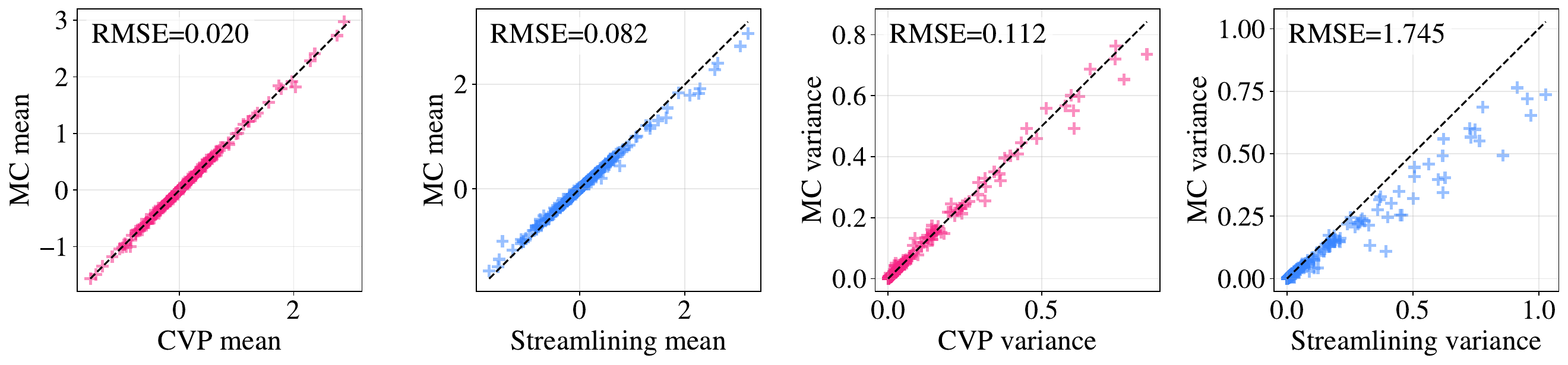}
        \caption{Real Data: ViT-Base on CIFAR-100}
        \label{fig:layernorm_real}
    \end{subfigure}

    \caption{\textbf{Validation of our LayerNorm approximation.} On synthetic and real data, our approximation (\cref{eq:method_layernorm_s2}) tracks MC Sampling more closely than the linearization employed by Streamlining. In both cases, every dot corresponds to a single post-LayerNorm activation.}
    \label{fig:layernorm_main}
\end{figure}

\myparagraph{Models and datasets} All models are trained using IVON \cite{ivon}, with hyperparameters tuned on a held-out validation split that is also used to fit the calibration scalars $\alpha_i$. Full hyperparameters and experimental details are in \cref{sec:app_hyperparams}. We evaluate on unimodal and multimodal settings:
\begin{itemize}
    \item \textbf{Image classification.} A 1.8M-parameter ViT trained from scratch on CIFAR-10 \cite{krizhevsky2009} (standard ViTs \cite{vit} saturate this task), ViT-Base fine-tuned on CIFAR-100, and the original ResNet-50 ImageNet checkpoints from \citet{ivon}.
    \item \textbf{VQA and Visual Reasoning.} ViLT \cite{vilt} on VQAv2 \cite{vqav2}, and BEiT-3 \cite{beit3} on VQAv2 and NLVR2 \cite{nlvr2}, reusing the IVON checkpoints from \citet{varvqa}.
\end{itemize}

\myparagraph{Baselines} We compare CVP against four reference points: the \emph{mean network} (\ie a single forward pass with $\theta=\mu$); \emph{MC sampling} with $M$ samples from $q(\theta)$ averaged after softmax; \emph{Streamlining} \cite{streamlining}, the closest variance propagation method; and \emph{temperature scaling} \cite{tempscaling} applied to the mean network's logits. Streamlining, CVP and temperature scaling receive identical calibration data. We note that our CVP implementation uses efficient operations and compilation to consistently stay in the $2$-$3\times$ runtime regime across many different architectures. For a fair runtime comparison, we also compile the deterministic models and MC sampling throughout.

\myparagraph{Metrics} We evaluate two complementary aspects of uncertainty quality. For calibration, we report Expected Calibration Error (ECE) \cite{ece,tempscaling}, the Brier score \cite{brier1950}, and negative log-likelihood. For ranking quality, we report Coverage at Risk,
\begin{equation}\label{eq:cov_at_risk}
    C@R = \max_{\gamma}\, C(\gamma) \quad \text{s.t.} \quad r(\gamma) \leq R,
\end{equation}
the highest fraction of inputs the model can answer at confidence threshold $\gamma$ while keeping the empirical risk $r(\gamma)$ below $R$ \cite{geifman_selpred}, and the threshold-free Area under the Risk-Coverage curve (AURC). We focus on high-stakes regimes (low $R$), where models are most differentiated. Additional risk levels, Adaptive Calibration Error \cite{ace}, and Effective Reliability \cite{reliable_vqa} are reported in \cref{sec:app_resultstables}.

\subsection{Image Classification with ViTs and ResNets}\label{sec:exp_vit}
With the single exception of ECE for the ResNet, CVP matches or outperforms all baselines across every metric, including the more expensive MC sampling with 4 samples (\cref{tab:default:vit_test_sem}). The gains are particularly marked in selective prediction: for AURC, CVP wins by $0.16$ over the next-best method on CIFAR-100 ViT-Base and by $0.19$ on CIFAR-10 sub-tiny ViT.

\begin{table}[ht]
  \centering
  \renewcommand{\arraystretch}{1.3}
  \setlength{\tabcolsep}{3.5pt}
  \footnotesize
  \caption{Ablation study: macro-average over all six benchmarks. SEMs combined assuming independence across benchmarks.}
  \label{tab:ablations:avg_test_sem}
  \begin{tabular}{l r r r r r r r}
    \toprule
    \multirow{2}{*}{Method} & \multicolumn{1}{c}{\multirow{2}{*}{Acc.}} & \multicolumn{1}{c}{\multirow{2}{*}{AURC ($\downarrow$)}} & \multicolumn{1}{c}{\multirow{2}{*}{C@$\tfrac{1}{2}$\%}} & \multicolumn{1}{c}{\multirow{2}{*}{C@1\%}} & \multicolumn{1}{c}{\multirow{2}{*}{NLL / BCE ($\downarrow$)}} & \multicolumn{1}{c}{\multirow{2}{*}{ECE ($\downarrow$)}} & \multicolumn{1}{c}{\multirow{2}{*}{Brier ($\downarrow$)}} \\
     &  &  &  &  &  &  &  \\
    \midrule
    \textcolor{CVP}{CVP} & \bvpm{78.8}{0.0} & \bvpm{6.20}{0.02} & \bvpm{19.1}{0.4} & \bvpm{28.7}{0.3} & \bvpm{1.33}{0.00} & \vpm{2.19}{0.04} & \bvpm{0.31}{0.00} \\
    \quad $-$ Normalization & \vpm{78.6}{0.0} & \vpm{6.34}{0.02} & \vpm{16.3}{0.4} & \vpm{26.0}{0.4} & \vpm{1.35}{0.00} & \bvpm{2.06}{0.04} & \vpm{0.31}{0.00} \\
    \quad $-$ Exact Activation & \vpm{78.6}{0.0} & \vpm{6.29}{0.02} & \vpm{17.4}{0.6} & \vpm{28.2}{0.3} & \vpm{1.37}{0.00} & \vpm{3.52}{0.09} & \vpm{0.32}{0.00} \\
    \quad $-$ Per-Layer Calib. & \vpm{78.3}{0.1} & \vpm{6.48}{0.03} & \vpm{16.6}{0.5} & \vpm{26.4}{0.4} & \vpm{1.35}{0.00} & \vpm{2.24}{0.05} & \vpm{0.31}{0.00} \\
    \quad $-$ All Calib. & \vpm{78.4}{0.1} & \vpm{6.46}{0.03} & \vpm{17.2}{0.6} & \vpm{27.3}{0.3} & \vpm{1.36}{0.00} & \vpm{4.06}{0.11} & \vpm{0.32}{0.00} \\
    \hdashline
    \textcolor{Streamlining}{Streamlining} & \vpm{78.4}{0.0} & \vpm{6.39}{0.02} & \vpm{15.4}{0.5} & \vpm{25.7}{0.4} & \vpm{1.35}{0.00} & \vpm{2.83}{0.06} & \vpm{0.32}{0.00} \\
    \quad + Per-Layer Calib. & \vpm{78.4}{0.0} & \vpm{6.35}{0.02} & \vpm{15.6}{0.5} & \vpm{26.8}{0.4} & \vpm{1.35}{0.00} & \vpm{2.87}{0.06} & \vpm{0.32}{0.00} \\
    \quad $-$ All Calib. & \vpm{78.4}{0.0} & \vpm{6.39}{0.02} & \vpm{14.9}{0.6} & \vpm{25.9}{0.3} & \vpm{1.37}{0.00} & \vpm{5.24}{0.13} & \vpm{0.32}{0.00} \\
    \bottomrule
  \end{tabular}
\end{table}

\subsection{VQA and Visual Reasoning with Multimodal Transformers}\label{sec:exp_vqa}
The same pattern carries over to multimodal models (\cref{tab:default:multimodal_test_sem}) - except for ECE in some cases, CVP matches or outperforms all baselines, with the largest gains in selective prediction. On NLVR2 BEiT-3, Streamlining even falls behind the mean network despite $>2\times$ the cost, and they match almost exactly on VQA. CVP, by contrast, consistently improves AURC over all methods, including MC sampling (\eg $0.57$ better than 4 samples on VQAv2 ViLT). The gap to Streamlining is particularly large on the high-stakes coverages $C@\frac{1}{2}\%$ and $C@1\%$.

\subsection{Validation of the LayerNorm Approximation}\label{sec:exp_layernorm}

We validate our LayerNorm approximation (\cref{eq:method_layernorm_s2}) in two regimes.

\emph{a) Synthetic data.} We sample inputs $x \sim \mathcal{N}(\mu_x, \Sigma_x)$ with $\mu_x \sim \mathcal{N}(0, I)$ and per-element variances drawn uniformly from $\mathcal{U}(0, 5)$. We keep the scale $\gamma$ and bias $\beta$ deterministic, since stochastic affine parameters reduce to a linear transformation handled exactly by \Cref{eq:method_linear}. The MC ground truth is computed from $M=4000$ samples. \cref{fig:layernorm_synthetic} shows that CVP's predicted output mean and variance align tightly with MC, while the linearization underestimates the variance. In \cref{sec:app_layernorm} we show that this gap widens with increasing input variance, whereas CVP stays close to the diagonal.

\emph{b) Real data.} We extract pre-LayerNorm activation distributions $(\mu_x, \Sigma_x)$ from a CVP forward pass on trained transformers via forward hooks, and run MC sampling ($M=200$), our rule, and Streamlining's linearization on each. Since each model contains many LayerNorm modules, we compute the RMSE over all of them combined, and subsample activations for the plots. \Cref{fig:layernorm_real} shows the resuls for CIFAR-100 (ViT-Base), results for the other models are similar (\cf \cref{sec:app_layernorm}).

\subsection{Ablations}\label{sec:exp_ablation}

We isolate the contribution of each CVP component by replacing it with the corresponding Streamlining choice or removing it entirely: the LayerNorm rule (replaced by $s^2(\mu_x)$), the exact activation moments (replaced by the Delta method), the per-layer calibration (replaced by a single output-level $\alpha$ as in Streamlining), and the calibration step entirely. We additionally test the converse direction by adding our per-layer calibration scheme to Streamlining's rules. Macro-averaged results across all six benchmarks are reported in \cref{tab:ablations:avg_test_sem}: removing any component degrades selective prediction (AURC, $C@\frac{1}{2}\%$, $C@1\%$), and the per-layer calibration applied to Streamlining narrows the gap to CVP, but does not close it. More detailed ablation results are in \cref{sec:app_ablations}.

\section{Discussion}\label{sec:discussion}
For uncertainty estimates to be useful in real-time, safety-critical systems, they must be available alongside the prediction, with no budget for repeated inference. CVP meets this requirement, delivering consistent Pareto improvements over the mean network and MC sampling on selective prediction across uni- and multimodal Bayesian transformers in a single forward pass. The two technical contributions, a closed-form rule for normlization layers and the per-layer calibration, allow variance propagation to provide reliable uncertainty estimates at this efficiency. We see two natural extensions: on the methodological side, richer approximations such as block-structured or low-rank covariances could improve fidelity at moderate added cost, while variance pruning could move CVP further toward the cost of the mean network. On the application side, future work could investigate whether the propagated moments transfer to tasks beyond selective prediction, such as out-of-distribution detection, active learning, or uncertainty-guided generation.

\myparagraph{Limitations} Like MC sampling, variance propagation incurs memory overhead at test time: a second set of parameter variances $\Sigma$ is stored alongside the means $\mu$, and each activation tensor carries its variance, roughly doubling the memory footprint of a forward pass. The per-layer calibration further requires a small held-out split, as does classical temperature scaling. While CVP is in principle agnostic to the underlying posterior, our experiments only cover variational posteriors from IVON; extending to other Gaussian approximations such as Laplace requires additional engineering and is left to future work. Our evaluation also covers only encoder-style architectures with classification or VQA-style heads; autoregressive generation is left to future work.

\newpage
\myparagraph{Acknowledgements} This research was partially funded by an Alexander von Humboldt Professorship in Multimodal Reliable AI, sponsored by Germany’s Federal Ministry for Research, Technology and Space and by a LOEWE-Spitzen-Professur (LOEWE/4a//519/05.00.002(0010)/93). TM was supported by the Bayes duality project, JST CREST Grant Number JPMJCR2112 and acknowledges the support of JSPS KAKENHI Grant Number 26H02541. The work has benefited from the Excellence Cluster “Reasonable AI” by the Deutsche Forschungsgemeinschaft (DFG, German Research Foundation) under Germany's Excellence Strategy – EXC-3057. For compute, we gratefully acknowledge support from the hessian.AI Service Center (funded by the Federal Ministry of Research, Technology and Space, BMFTR, grant no. 16IS22091) and the hessian.AI Innovation Lab (funded by the Hessian Ministry for Digital Strategy and Innovation, grant no. S-DIW04/0013/003).

\bibliographystyle{abbrvnat}
\bibliography{main}

\crefalias{section}{appendix}
\crefalias{subsection}{appendix}
\crefalias{subsubsection}{appendix}

\newpage
\appendix
\section{Full derivations of Variance Propagation rules}\label{sec:appendix_derivations}

\subsection{Notation and Prerequisites}\label{sec:app_notation}

Throughout the derivations below, we make use of the following well-known results. Let $\alpha, \alpha_i, \beta \in \sR$ be constants and let $X, X_i, Y \in \sR$ be random variables.

\begin{enumerate}
    \item Linearity of expectation:
    \begin{equation}\label{eq:statement1}
        \E\!\left[\sum_i \alpha_i X_i + \beta\right] = \sum_i \alpha_i\, \E[X_i] + \beta \tag{I}
    \end{equation}
    
    \item Expectation of products of \textbf{independent} variables:
    \begin{equation}\label{eq:statement2}
        \E\!\left[\prod_i X_i\right] = \prod_i \E[X_i] \tag{II}
    \end{equation}
    
    \item Variance--expectation relation:
    \begin{equation}\label{eq:statement3}
        \Var[X] = \E[X^2] - \E[X]^2 \tag{III}
    \end{equation}
    
    \item Variance scaling:
    \begin{equation}\label{eq:statement4}
        \Var[\alpha X + \beta] = \alpha^2\, \Var[X] \tag{IV}
    \end{equation}
    
    \item Variance of sums of \textbf{independent} variables:
    \begin{equation}\label{eq:statement5}
        \Var\!\left[\sum_i X_i\right] = \sum_i \Var[X_i] \tag{V}
    \end{equation}
\end{enumerate}

These statements require finite expectations of the $X_i$; \cref{eq:statement3,eq:statement4,eq:statement5} additionally require finite second moments, and \cref{eq:statement2,eq:statement5} require independence of the $X_i$.

We use the notation $\mu_z = \E[z]$ for the mean, $\Sigma_z = \Var[z]$ for the variance, and $\sigma_z = \sqrt{\Sigma_z}$ for the standard deviation. Following the diagonal-covariance assumption from \cref{sec:background}, all expectations and variances of vectors and matrices are understood to be element-wise. The element-wise (Hadamard) product of two matrices is denoted $\odot$, while $\cdot$ denotes the matrix--vector (dot) product.

Several activation function approximations below assume that the input is Gaussian-distributed. We therefore also define the standard Gaussian PDF $\phi(x)$ and CDF $\Phi(x)$:
\begin{equation}
    \phi(x) = \frac{1}{\sqrt{2\pi}}\, e^{-x^2/2},
\end{equation}
\begin{equation}
    \Phi(x) = \int_{-\infty}^x \mathcal{N}(\theta\,|\,0, 1)\, d\theta = \frac{1}{2}\!\left[1 + \mathrm{erf}\!\left(\frac{x}{\sqrt{2}}\right)\right].
\end{equation}

\subsection{Linear Layer}\label{sec:app_deriv_linearlayer}

Let the input and output dimensions be $M, N \in \sN$. The derivation assumes independence between the sampled weights $W \in \sR^{N \times M}$, the bias $b \in \sR^N$, and the input $x \in \sR^M$, as well as diagonal covariances for $W$, $x$, and $b$.

For the output mean and variance,
\begin{alignat*}{2}
    \mu_y    &= \E[W \cdot x + b] \\
             &= \E[W \cdot x] + \E[b] \qquad && \text{(I)} \\
             &= \E[W] \cdot \E[x] + \E[b] \qquad && \text{(I), (II)} \\
             &= \mu_W \cdot \mu_x + \mu_b, \\[12pt]
    \Sigma_y &= \Var[W \cdot x + b] \\
             &= \Var[W \cdot x] + \Sigma_b \qquad && \text{(V)} \\
             &= \E[(W \odot W) \cdot x^2] - \E[W \cdot x]^2 + \Sigma_b \qquad && \text{(III)} \\
             &= \E[W \odot W] \cdot \E[x^2] - (\mu_W \odot \mu_W) \cdot \mu_x^2 + \Sigma_b \qquad && \text{(II)} \\
             &= (\Sigma_W + \mu_W \odot \mu_W)(\Sigma_x + \mu_x^2) - (\mu_W \odot \mu_W) \mu_x^2 + \Sigma_b \qquad && \text{(III)}.
\end{alignat*}

Expanding the product and cancelling the $(\mu_W \odot \mu_W) \mu_x^2$ terms gives
\begin{equation}\label{eq:app_linear}
    \Sigma_y = (\Sigma_W + \mu_W \odot \mu_W) \cdot \Sigma_x + \Sigma_W \cdot \mu_x^2 + \Sigma_b.
\end{equation}

\subsection{Activation Functions}\label{sec:app_deriv_actfunc}

In the following, we assume $x \in \sR$, but as activation functions are typically applied element-wise, the results immediately generalize to higher dimensions. The Delta method requires only the first and second moments of $x$ to exist. The moment-matching strategies below additionally assume that the input is Gaussian-distributed.

\myparagraph{Delta Method} This approach propagates uncertainty through any differentiable nonlinear function $g(x)$ via local linearization around the input mean:
\begin{equation}
    g_{\textrm{local}}(x) = g(\mu_x) + g'(\mu_x)(x - \mu_x).
\end{equation}
The resulting moments are
\begin{alignat*}{2}
    \mu_y    &= \E[g_{\textrm{local}}(x)] \\
             &= g(\mu_x) + g'(\mu_x)(\mu_x - \mu_x) \qquad && \text{(I)} \\
             &= g(\mu_x), \\[8pt]
    \Sigma_y &= \Var[g_{\textrm{local}}(x)] \\
             &= g'(\mu_x)^2 \cdot \Sigma_x \qquad && \text{(IV)}.
\end{alignat*}
The Delta method is a first-order Taylor approximation and is therefore inaccurate for activations with high curvature or inputs with large variance. The moment-matching approaches below derive more exact moments under a Gaussian-input assumption.

\myparagraph{Sigmoid} For the logistic sigmoid $\sigma(x) = 1/(1 + e^{-x})$, \citet{bishop} proposed an approximation based on the close similarity of $\sigma(x)$ and $\Phi(x)$. They introduced a scaling parameter $\lambda$ to match $\Phi(\lambda x)$ to $\sigma(x)$ by requiring identical slopes at the origin, yielding $\lambda = \sqrt{\pi/8}$. Using $\Phi(\lambda x) \approx \sigma(x)$, the propagated mean is
\begin{equation}\label{eq:bishop_sigmoid_derivations}
    \begin{aligned}
        \mu_y &= \int \sigma(x)\, \mathcal{N}(x \,|\, \mu_x, \Sigma_x)\, dx \\
              &\approx \int \Phi(\lambda x)\, \mathcal{N}(x \,|\, \mu_x, \Sigma_x)\, dx \\
              &= \Phi\!\left(\frac{\lambda \mu_x}{\sqrt{1 + \lambda^2 \Sigma_x}}\right) \\
              &\approx \sigma\!\left(\frac{\mu_x}{\sqrt{1 + \lambda^2 \Sigma_x}}\right) = \sigma\!\left(\frac{\mu_x}{\beta_x}\right),
    \end{aligned}
\end{equation}
with $\beta_x = \sqrt{1 + \lambda^2 \Sigma_x} \geq 1$. The third equality follows from explicit integration; we omit the algebra for brevity.

\citet{sigmoid_moment_matching} arrives at the same form $\Phi(\lambda x) \approx \sigma(x)$ but with $\lambda = \sqrt{3/\pi^2}$, obtained by matching the standard logistic distribution against the standard normal via their CDFs. They additionally derive an expression for the propagated variance:
\begin{alignat*}{2}
    \Sigma_y &= \E[\sigma(x)^2] - \E[\sigma(x)]^2 \\
             &= \E[\sigma(x) - \sigma(x)(1 - \sigma(x))] - \E[\sigma(x)]^2 \\
             &= \E[\sigma(x) - \sigma'(x)] - \E[\sigma(x)]^2 \qquad && \sigma'(x) = \sigma(x)(1 - \sigma(x)) \\
             &= \E[\sigma(x)](1 - \E[\sigma(x)]) - \E[\sigma'(x)] \qquad && \text{(I)} \\
             &= \sigma\!\left(\frac{\mu_x}{\beta_x}\right)\!\left(1 - \sigma\!\left(\frac{\mu_x}{\beta_x}\right)\right) - \E[\sigma'(x)].
\end{alignat*}
The remaining term $\E[\sigma'(x)]$ is approximated by again using $\sigma(x) \approx \Phi(\lambda x)$, which gives $\sigma'(x) \approx \mathcal{N}(x \,|\, 0, 1/\lambda^2)$. After algebraic manipulation,
\begin{equation}
    \Sigma_y = \sigma\!\left(\frac{\mu_x}{\beta_x}\right)\!\left(1 - \sigma\!\left(\frac{\mu_x}{\beta_x}\right)\right)\!\left(1 - \frac{1}{\beta_x}\right).
\end{equation}

\myparagraph{Tanh} The hyperbolic tangent satisfies $\tanh(x) = 2\sigma(2x) - 1$. We can therefore reduce tanh moment matching to the sigmoid case: given an input with moments $(\mu_x, \Sigma_x)$, we evaluate the sigmoid moments above on the rescaled input $(2\mu_x, 4\Sigma_x)$, and combine them according to (I) and (IV):
\begin{alignat*}{2}
    \mu_y    &= 2\, \E[\sigma(2x)] - 1, \\[12pt]
    \Sigma_y &= 4\, \Var[\sigma(2x)],
\end{alignat*}
where $\E[\sigma(2x)]$ and $\Var[\sigma(2x)]$ are obtained from \cref{eq:bishop_sigmoid_derivations} and the variance expression below it, with $\mu_x$ replaced by $2\mu_x$ and $\Sigma_x$ replaced by $4\Sigma_x$.

\myparagraph{ReLU and Heaviside} For ReLU and the Heaviside step function $H(x) = \mathbb{1}_{x \geq 0}$, closed-form expressions for both $\mu_y$ and $\Sigma_y$ under Gaussian input were first reported by \citet{frey1999}:
\begin{alignat*}{4}
    g(x) &= H(x)              &&\implies \quad & \mu_y    &= \Phi\!\left(\frac{\mu_x}{\sigma_x}\right),                                       &\qquad&  \\
         &                    &&              & \Sigma_y &= \mu_y(1 - \mu_y),                                                                  &\qquad&  \\[12pt]
    g(x) &= \mathrm{ReLU}(x)  &&\implies \quad & \mu_y    &= \mu_x \Phi\!\left(\frac{\mu_x}{\sigma_x}\right) + \sigma_x \phi\!\left(\frac{\mu_x}{\sigma_x}\right),  &\qquad&  \\
         &                    &&              & \Sigma_y &= \Sigma_x \Phi\!\left(\frac{\mu_x}{\sigma_x}\right) + \mu_y(\mu_x - \mu_y).         &\qquad&
\end{alignat*}

\myparagraph{GELU} For the Gaussian Error Linear Unit $\mathrm{GELU}(x) = x \Phi(x)$, \citet{analytic_covar_prop} derived the closed-form mean
\begin{equation}
    \mu_y = \mu_x \Phi\!\left(\frac{\mu_x}{\sqrt{1 + \Sigma_x}}\right) + \frac{\Sigma_x}{\sqrt{1 + \Sigma_x}}\, \phi\!\left(\frac{\mu_x}{\sqrt{1 + \Sigma_x}}\right),
\end{equation}
and an infinite-series expression for $\Sigma_y$ based on derivatives of $\mu_y$ with respect to $\mu_x$. \citet{residual_varprop} subsequently derived an exact closed form for $\Sigma_y$ that avoids the series truncation, which we use throughout. We refer the reader to Appendix D of \citet{residual_varprop} for the full expression.

\subsection{Layer Normalization}\label{sec:app_deriv_layernorm}

Recall the LayerNorm operation: given input activations $x \in \sR^D$,
\begin{equation}\label{eq:app_layernorm}
    y = \frac{x - m(x)}{\sqrt{s^2(x) + \epsilon}} \odot \gamma + \beta,
\end{equation}
where $m(x) = \tfrac{1}{D}\sum_d x_d$ and $s^2(x) = \tfrac{1}{D}\sum_d (x_d - m(x))^2$ are the across-dimension mean and variance, with elementwise scale $\gamma \in \sR^D$ and shift $\beta \in \sR^D$. The denominator depends on $x$, making LayerNorm nonlinear. We address this by replacing $s^2(x)$ with its expectation $\E[s^2(x)]$ under the input distribution, turning \cref{eq:app_layernorm} into an affine transformation of $x$.

\paragraph{Centering matrix.} For ease of notation, we write the centering operation $x - m(x)$ as a matrix product $M_\mathrm{center} \cdot x$ with
\begin{align}
    M_\mathrm{center}                         &= I_{D \times D} - \tfrac{1}{D} \mathbf{1}_{D \times D}, \\
    M_\mathrm{center} \odot M_\mathrm{center} &= \left(1 - \tfrac{2}{D}\right) I_{D \times D} + \tfrac{1}{D^2} \mathbf{1}_{D \times D}, \label{eq:m_center_identity}
\end{align}
where $I_{D \times D}$ is the identity matrix and $\mathbf{1}_{D \times D}$ is the all-ones matrix. The LayerNorm equation becomes
\begin{equation}\label{eq:app_layernorm_compact}
    y = \frac{M_\mathrm{center} \cdot x}{\sqrt{\E[s^2(x)] + \epsilon}} \odot \gamma + \beta.
\end{equation}

\paragraph{Computing $\E[s^2(x)]$.} Writing $(M_\mathrm{center} \cdot x)_d$ for the $d$-th element of the centered vector,
\begin{alignat*}{2}
    \E[s^2(x)] &= \E\!\left[\tfrac{1}{D}\sum_d (x_d - m(x))^2\right] \\
               &= \tfrac{1}{D}\sum_d \E\!\left[(M_\mathrm{center} \cdot x)_d^2\right] \qquad && \text{(I)} \\
               &= \tfrac{1}{D}\sum_d \E\!\left[(M_\mathrm{center} \cdot x)_d\right]^2 + \Var\!\left[(M_\mathrm{center} \cdot x)_d\right] \qquad && \text{(III)} \\
               &= \tfrac{1}{D}\sum_d (M_\mathrm{center} \cdot \mu_x)_d^2 + \tfrac{1}{D}\sum_d (M_\mathrm{center} \odot M_\mathrm{center} \cdot \Sigma_x)_d.
\end{alignat*}

The first term is, by definition, the LayerNorm variance statistic $s^2$ evaluated at $\mu_x$:
\begin{equation}
    \tfrac{1}{D}\sum_d (M_\mathrm{center} \cdot \mu_x)_d^2 = s^2(\mu_x).
\end{equation}
For the second term, \cref{eq:m_center_identity} gives $(M_\mathrm{center} \odot M_\mathrm{center} \cdot \Sigma_x)_d = (1 - 2/D)\Sigma_{x,d} + \bar{\Sigma}/D$, where $\bar{\Sigma} = \tfrac{1}{D}\sum_d \Sigma_{x,d}$. Averaging over $d$,
\begin{equation}
    \tfrac{1}{D}\sum_d (M_\mathrm{center} \odot M_\mathrm{center} \cdot \Sigma_x)_d = (1 - 2/D)\bar{\Sigma} + \bar{\Sigma}/D = \left(1 - \tfrac{1}{D}\right) \bar{\Sigma}.
\end{equation}

Combining the two terms,
\begin{equation}\label{eq:app_layernorm_s2}
    \E[s^2(x)] = s^2(\mu_x) + \left(1 - \tfrac{1}{D}\right) \bar{\Sigma}.
\end{equation}

\paragraph{Output mean.} With $\E[s^2(x)] + \epsilon$ now a constant under the input distribution, $\mu_y$ follows from the linearity and independence of $x$, $\gamma$, and $\beta$:
\begin{alignat*}{2}
    \mu_y &= \E\!\left[\frac{M_\mathrm{center} \cdot x}{\sqrt{\E[s^2(x)] + \epsilon}} \odot \gamma + \beta\right] \\
          &= \frac{M_\mathrm{center} \cdot \mu_x}{\sqrt{\E[s^2(x)] + \epsilon}} \odot \mu_\gamma + \mu_\beta \qquad && \text{(I), (II)} \\
          &= \frac{\mu_x - m(\mu_x)}{\sqrt{s^2(\mu_x) + \left(1 - \tfrac{1}{D}\right)\bar{\Sigma} + \epsilon}} \odot \mu_\gamma + \mu_\beta.
\end{alignat*}

\paragraph{Output variance.} The variance computation requires care because $\gamma_d$ multiplies the entire sum $(M_\mathrm{center} \cdot x)_d$, not each $x_j$ individually. For a fixed output dimension $d$, let $A = (M_\mathrm{center} \cdot x)_d$ and $B = \gamma_d$, with $A$ and $B$ independent. Then $y_d = (\E[s^2(x)] + \epsilon)^{-1/2}\, A B + \beta_d$, and using the variance-of-product identity for independent random variables,
\begin{alignat*}{2}
    \Var[A B] &= \E[A]^2\, \Var[B] + \Var[A]\, \E[B]^2 + \Var[A]\, \Var[B] \\
              &= \E[A]^2\, \Sigma_{\gamma,d} + \Var[A]\, (\mu_{\gamma,d}^2 + \Sigma_{\gamma,d}).
\end{alignat*}
Substituting $\E[A] = (M_\mathrm{center} \cdot \mu_x)_d$ and $\Var[A] = (M_\mathrm{center} \odot M_\mathrm{center} \cdot \Sigma_x)_d$, then dividing by $\E[s^2(x)] + \epsilon$ and adding $\Sigma_{\beta,d}$, gives
\begin{equation}\label{eq:app_layernorm_variance}
    \Sigma_y = \frac{(M_\mathrm{center} \cdot \mu_x)^2 \odot \Sigma_\gamma + (M_\mathrm{center} \odot M_\mathrm{center} \cdot \Sigma_x) \odot (\Sigma_\gamma + \mu_\gamma^2)}{\E[s^2(x)] + \epsilon} + \Sigma_\beta.
\end{equation}

\section{Efficiency-Reliability Pareto plots}\label{sec:app_paretoplots}
\begin{figure}[!h]
    \centering
    \begin{subfigure}[t]{0.32\textwidth}
        \centering
        \includegraphics[width=\textwidth]{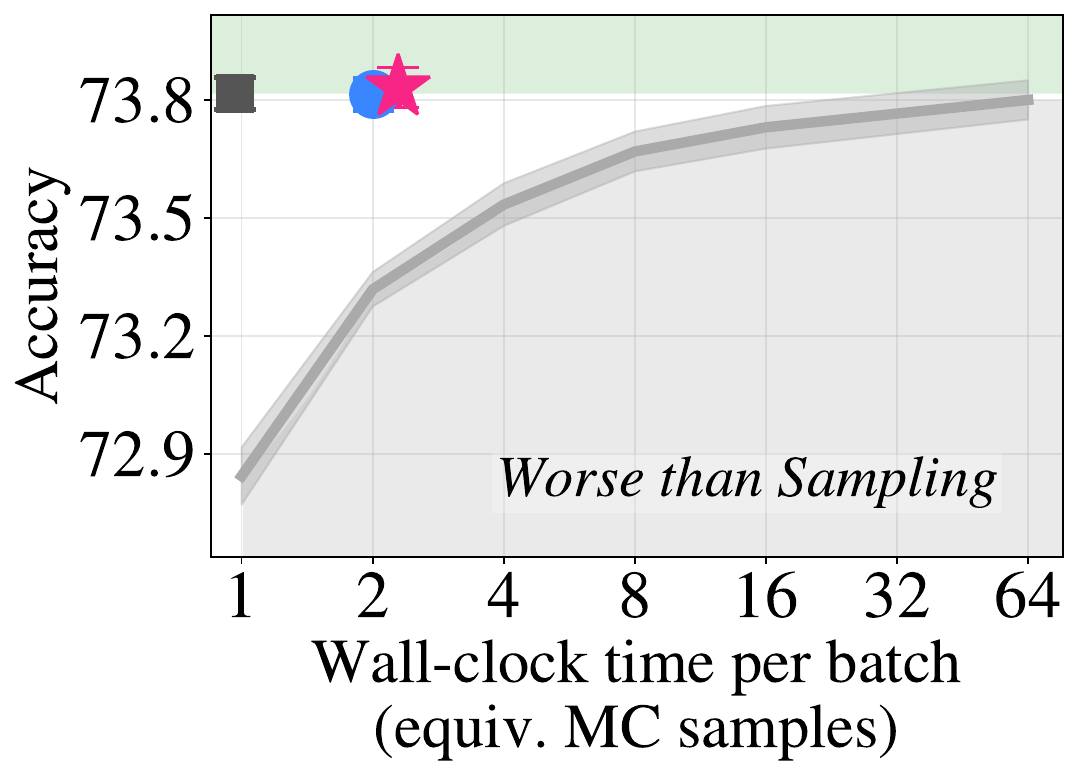}
    \end{subfigure}
    \hfill
    \begin{subfigure}[t]{0.32\textwidth}
        \centering
        \includegraphics[width=\textwidth]{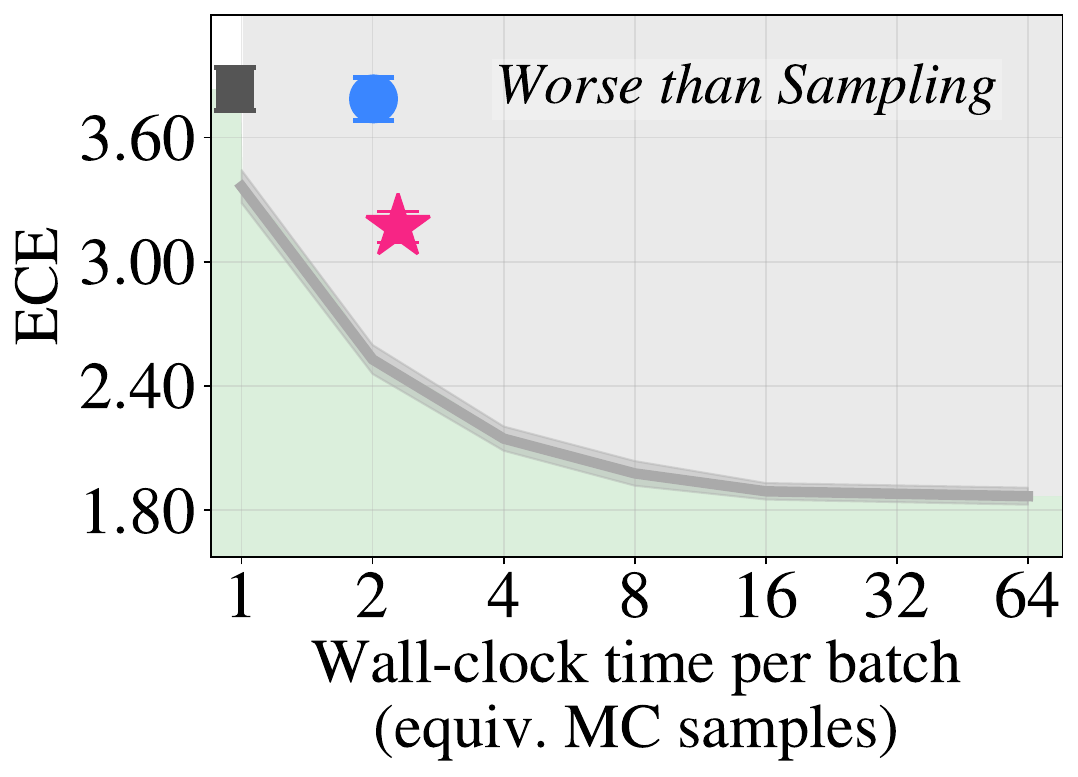}
    \end{subfigure}
    \hfill
    \begin{subfigure}[t]{0.32\textwidth}
        \centering
        \includegraphics[width=\textwidth]{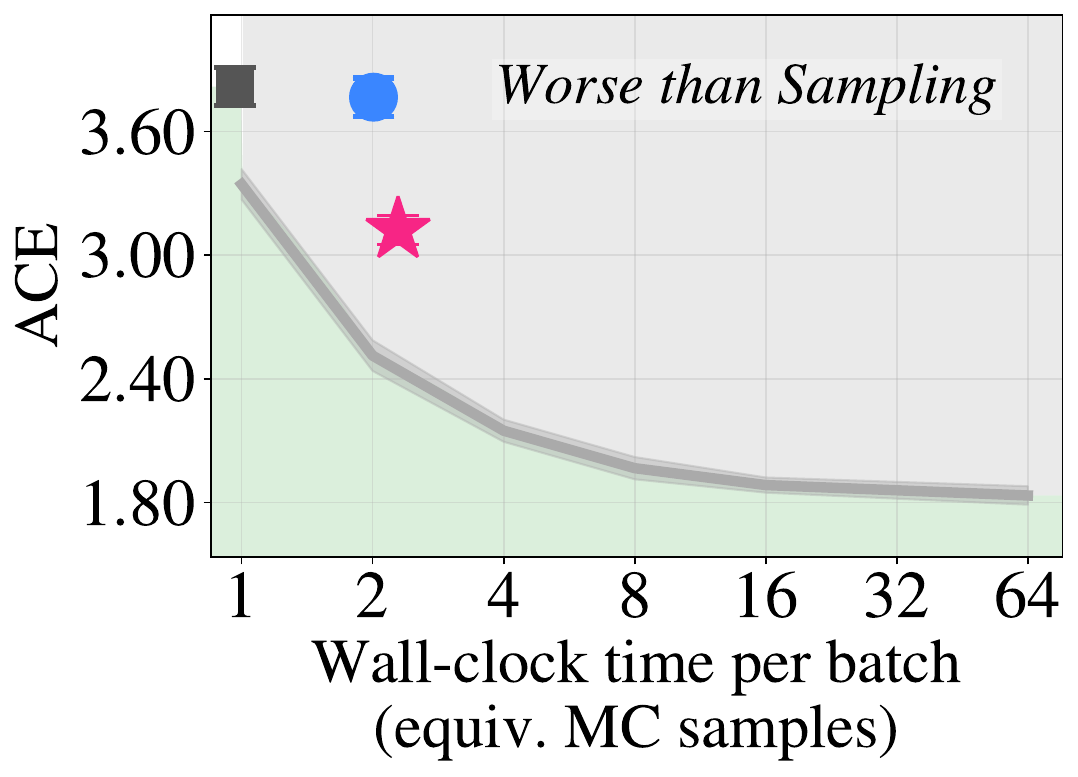}
    \end{subfigure}

    \vspace{1em}  %

    \begin{subfigure}[t]{0.32\textwidth}
        \centering
        \includegraphics[width=\textwidth]{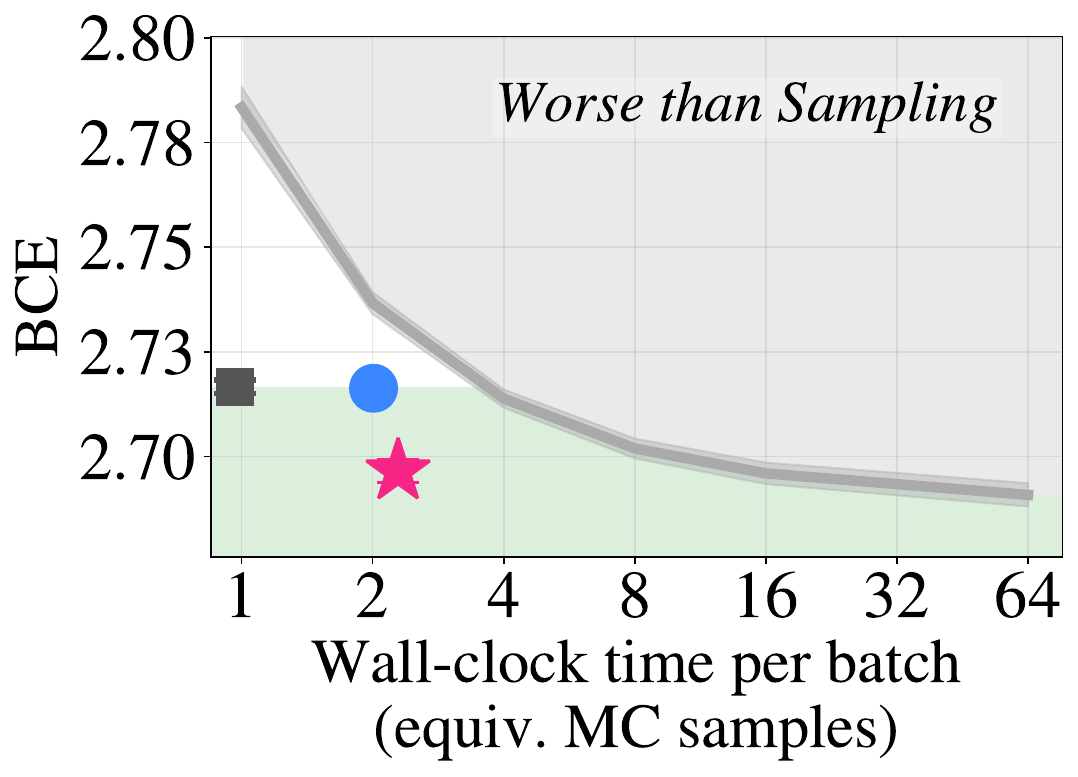}
    \end{subfigure}
    \hfill
    \begin{subfigure}[t]{0.32\textwidth}
        \centering
        \includegraphics[width=\textwidth]{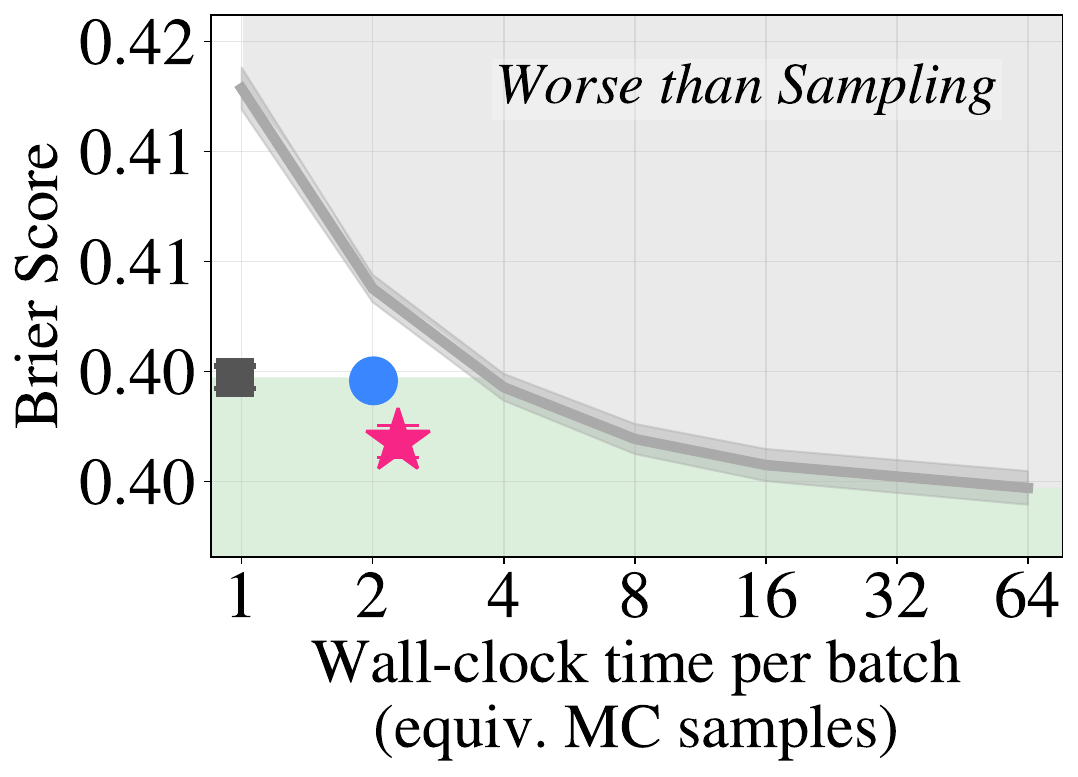}
    \end{subfigure}
    \hfill
    \begin{subfigure}[t]{0.32\textwidth}
        \centering
        \includegraphics[width=\textwidth]{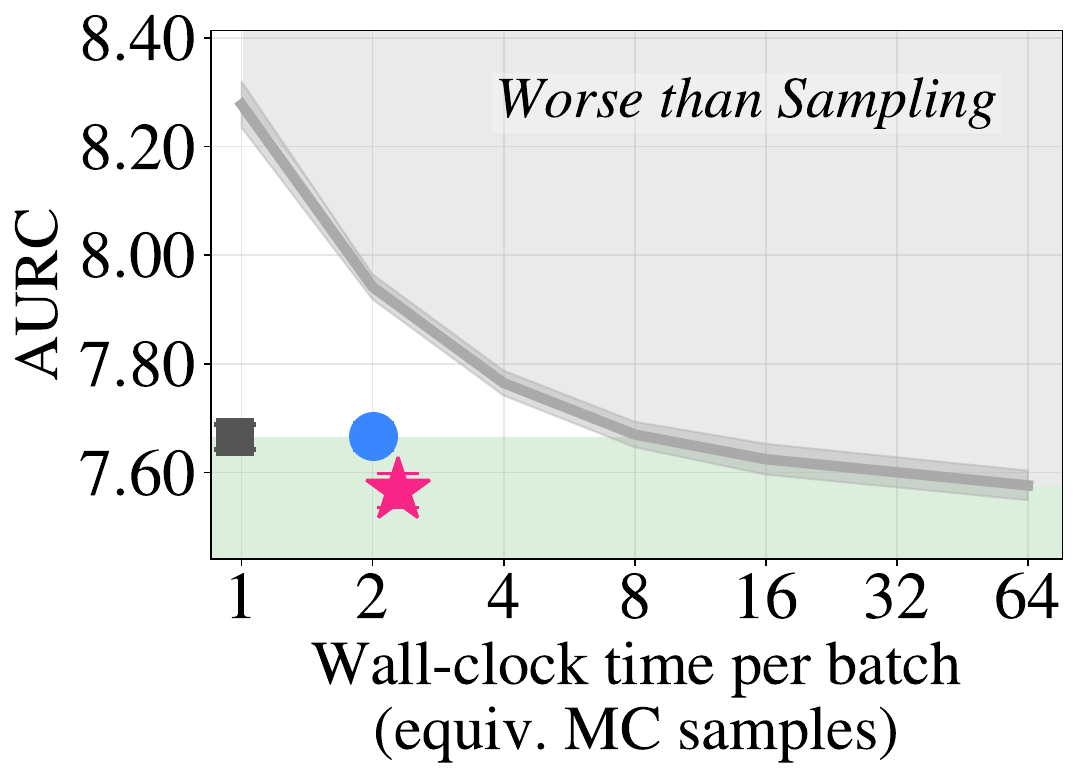}
    \end{subfigure}

    \vspace{1em}

    \begin{subfigure}[t]{0.32\textwidth}
        \centering
        \includegraphics[width=\textwidth]{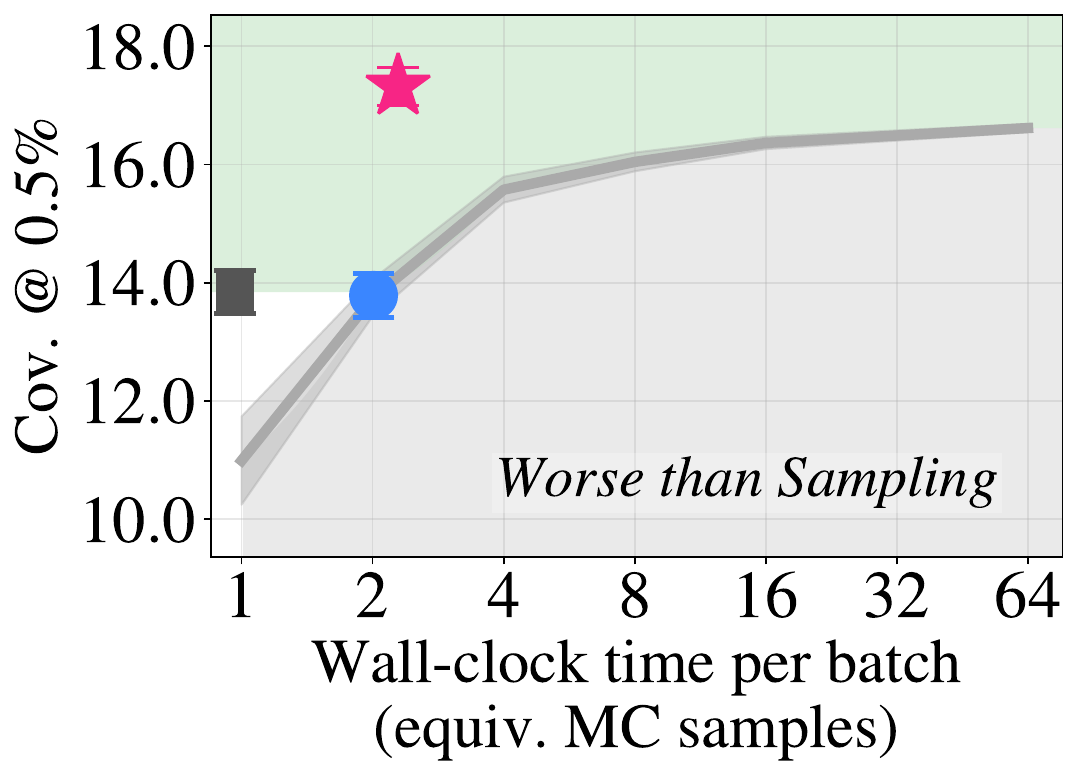}
    \end{subfigure}
    \hfill
    \begin{subfigure}[t]{0.32\textwidth}
        \centering
        \includegraphics[width=\textwidth]{new_figures/metrics/beitvqa/C_0.01_errors.pdf}
    \end{subfigure}
    \hfill
    \begin{subfigure}[t]{0.32\textwidth}
        \centering
        \includegraphics[width=\textwidth]{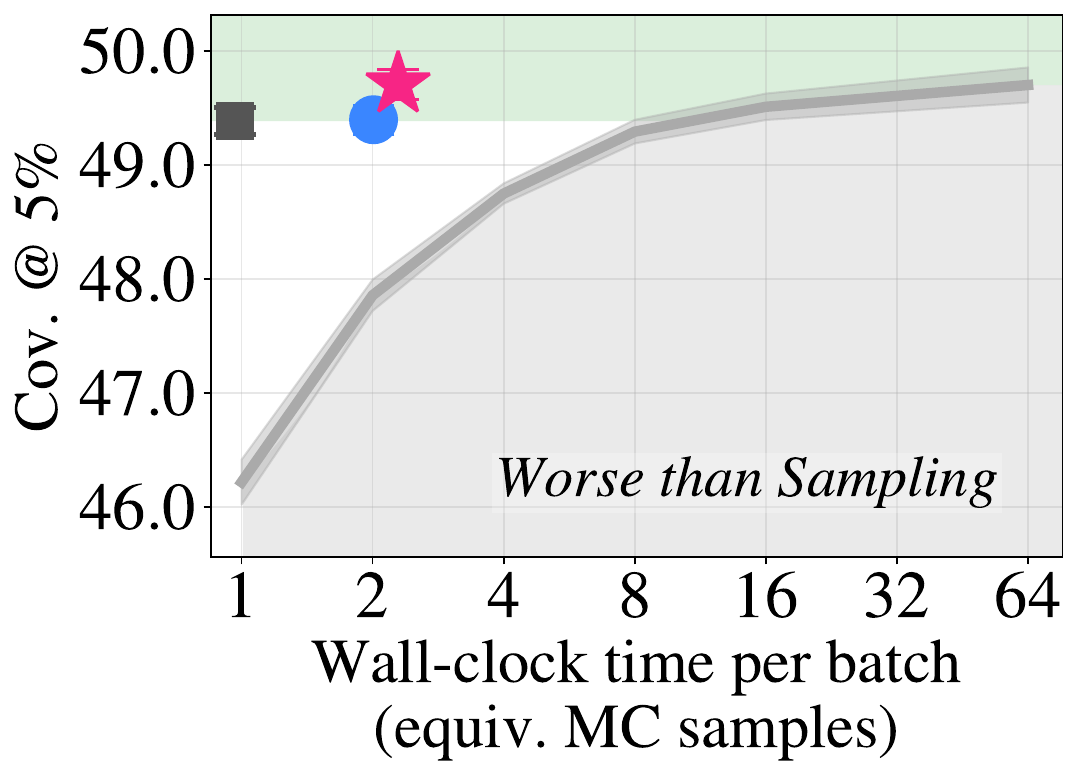}
    \end{subfigure}

    \vspace{1em}

    \begin{subfigure}[t]{0.32\textwidth}
        \centering
        \includegraphics[width=\textwidth]{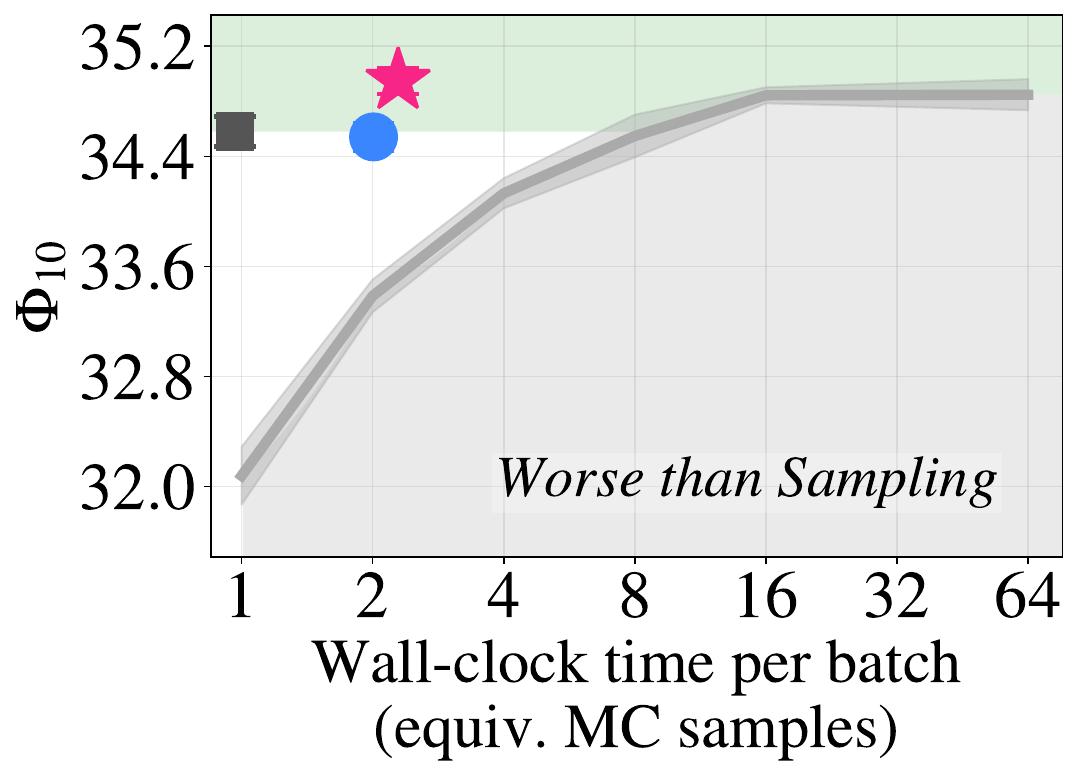}
    \end{subfigure}
    \hfill
    \begin{subfigure}[t]{0.32\textwidth}
        \centering
        \includegraphics[width=\textwidth]{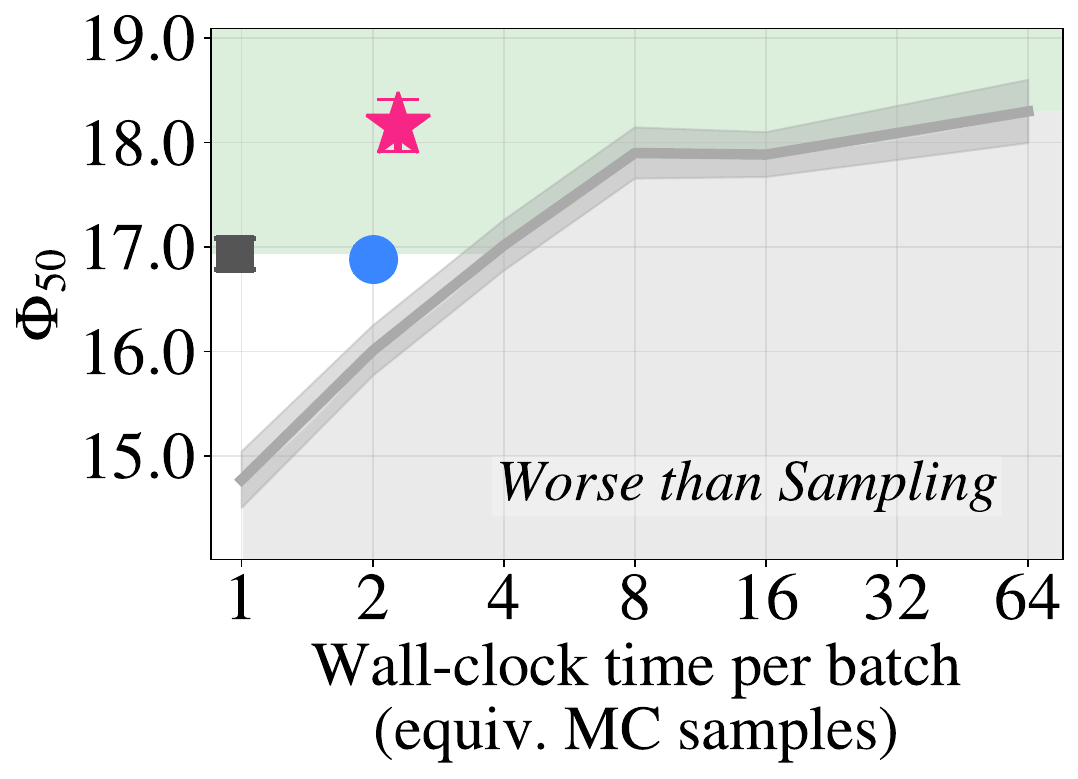}
    \end{subfigure}
    \hfill
    \begin{subfigure}[t]{0.32\textwidth}
        \centering
        \includegraphics[width=\textwidth]{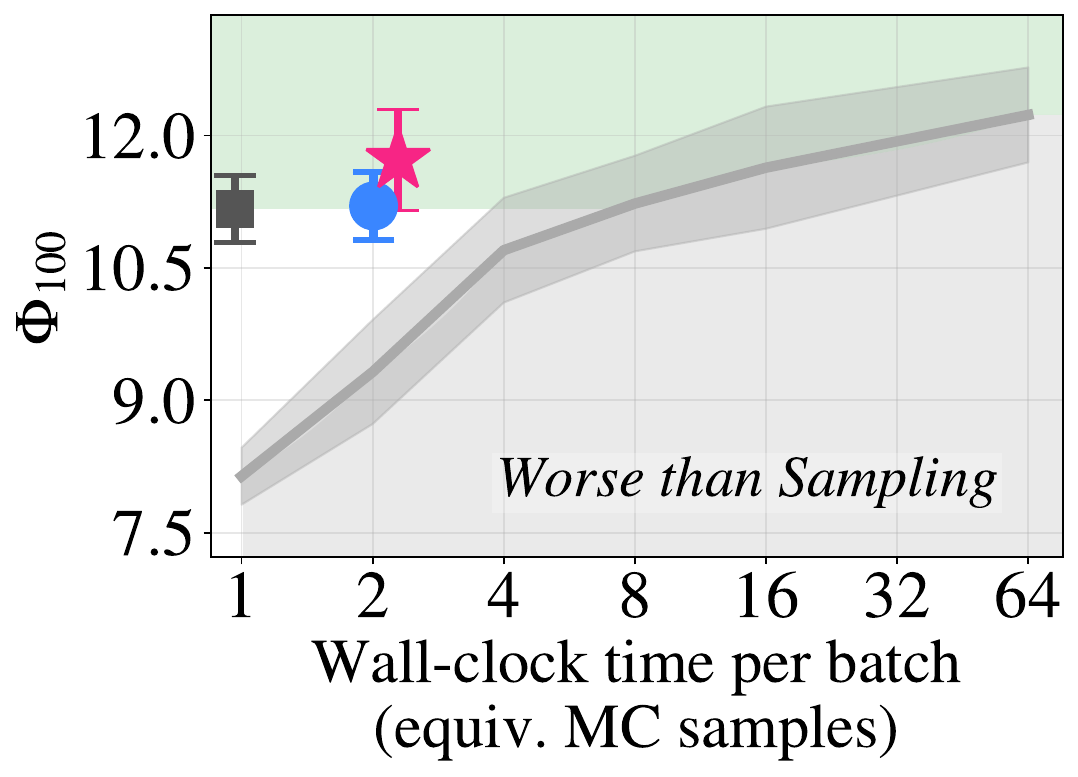}
    \end{subfigure}

    \vspace{0.3em}
    \includegraphics[width=0.6\textwidth]{new_figures/metrics/legend_horizontal.pdf}
    \caption{Results on Accuracy, Calibration, Loss, the Brier Score and Selective Prediction for BEiT-Base on VQAv2. The \colorbox{Paretogreen!20}{Pareto-dominating region} (vs. the mean network and MC Sampling) is highlighted in green. SEM (Standard Error of the Mean) across 5 random seeds is shown.}
    \label{fig:full_beitvqa}
\end{figure}

\begin{figure}[t]
    \centering
    \begin{subfigure}[t]{0.32\textwidth}
        \centering
        \includegraphics[width=\textwidth]{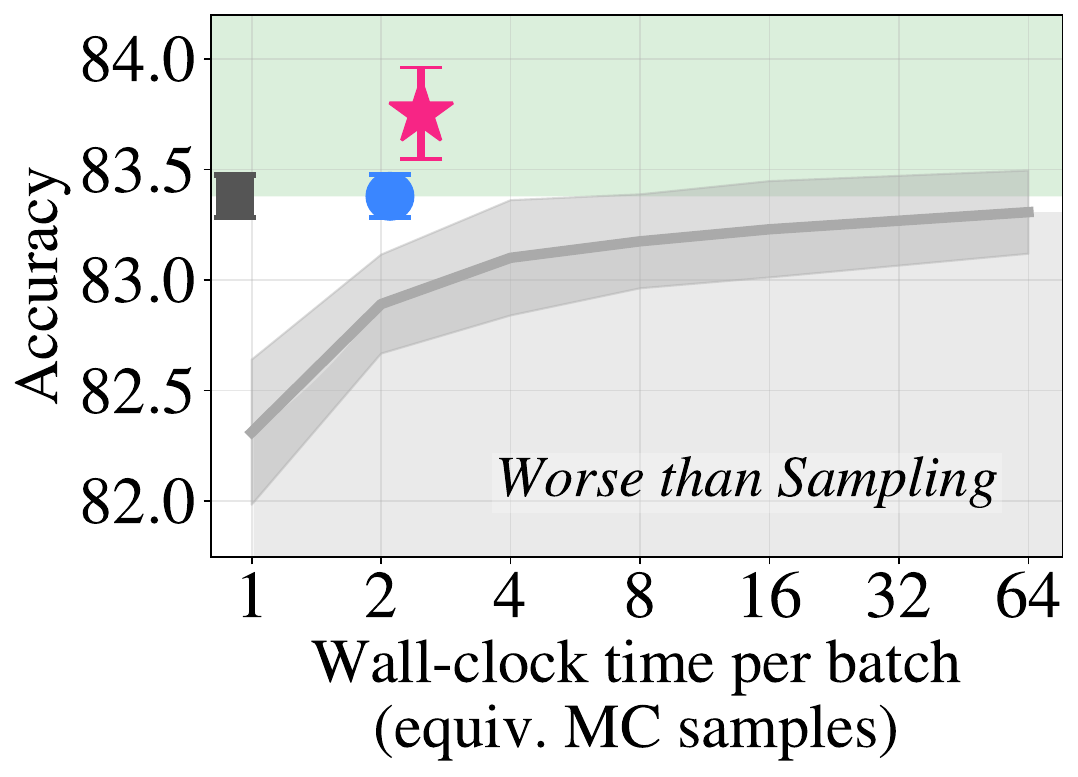}
    \end{subfigure}
    \hfill
    \begin{subfigure}[t]{0.32\textwidth}
        \centering
        \includegraphics[width=\textwidth]{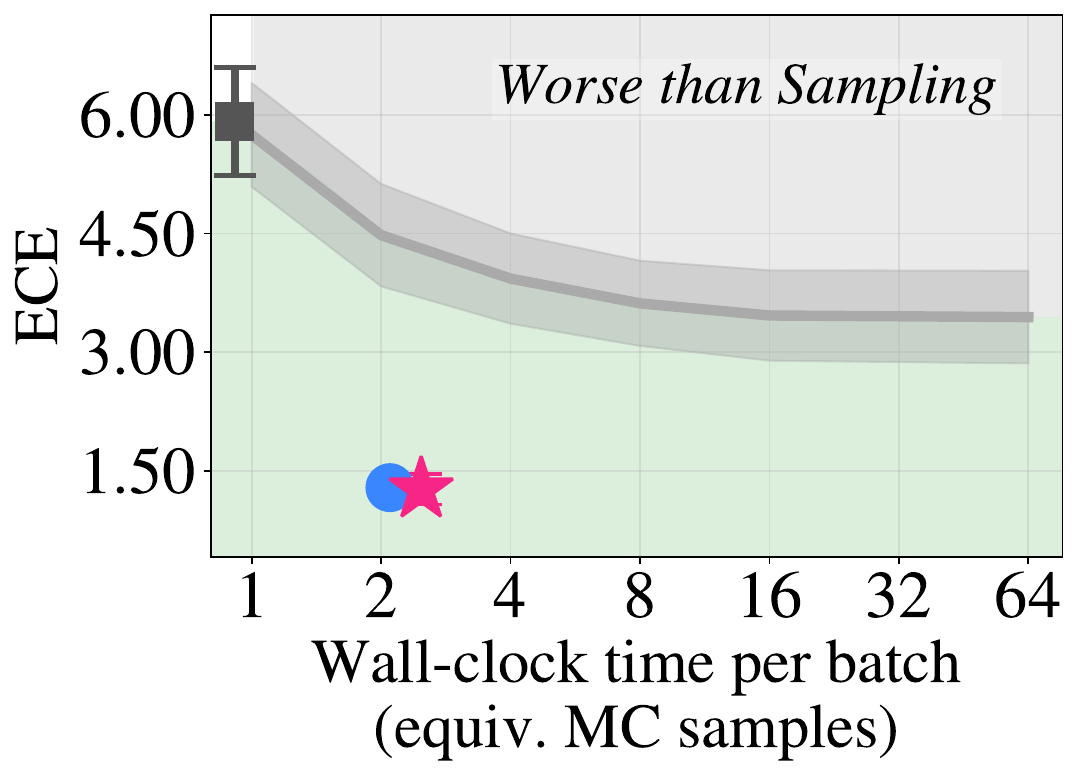}
    \end{subfigure}
    \hfill
    \begin{subfigure}[t]{0.32\textwidth}
        \centering
        \includegraphics[width=\textwidth]{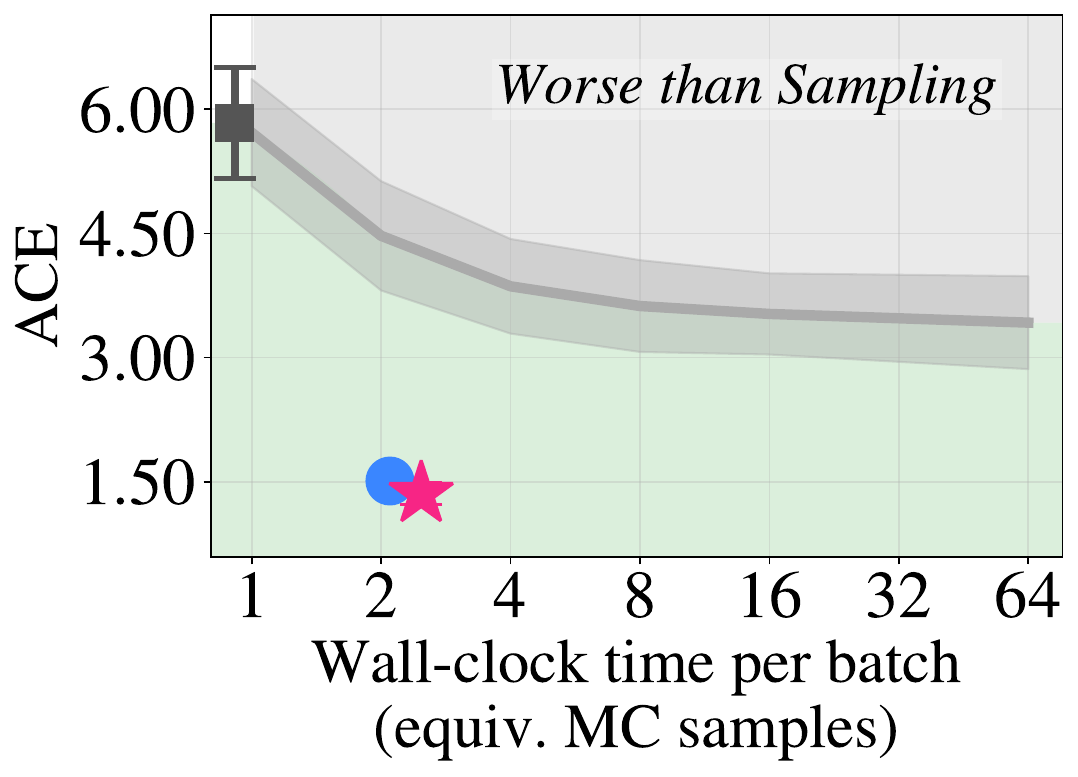}
    \end{subfigure}

    \vspace{1em}  %

    \begin{subfigure}[t]{0.32\textwidth}
        \centering
        \includegraphics[width=\textwidth]{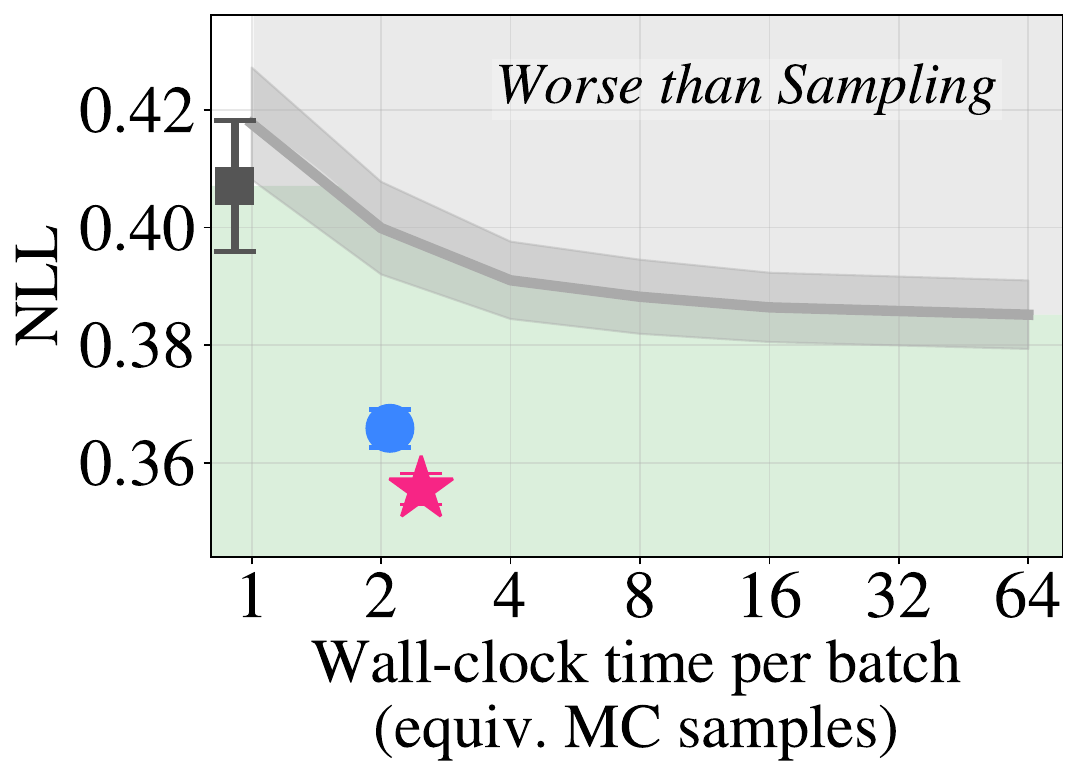}
    \end{subfigure}
    \hfill
    \begin{subfigure}[t]{0.32\textwidth}
        \centering
        \includegraphics[width=\textwidth]{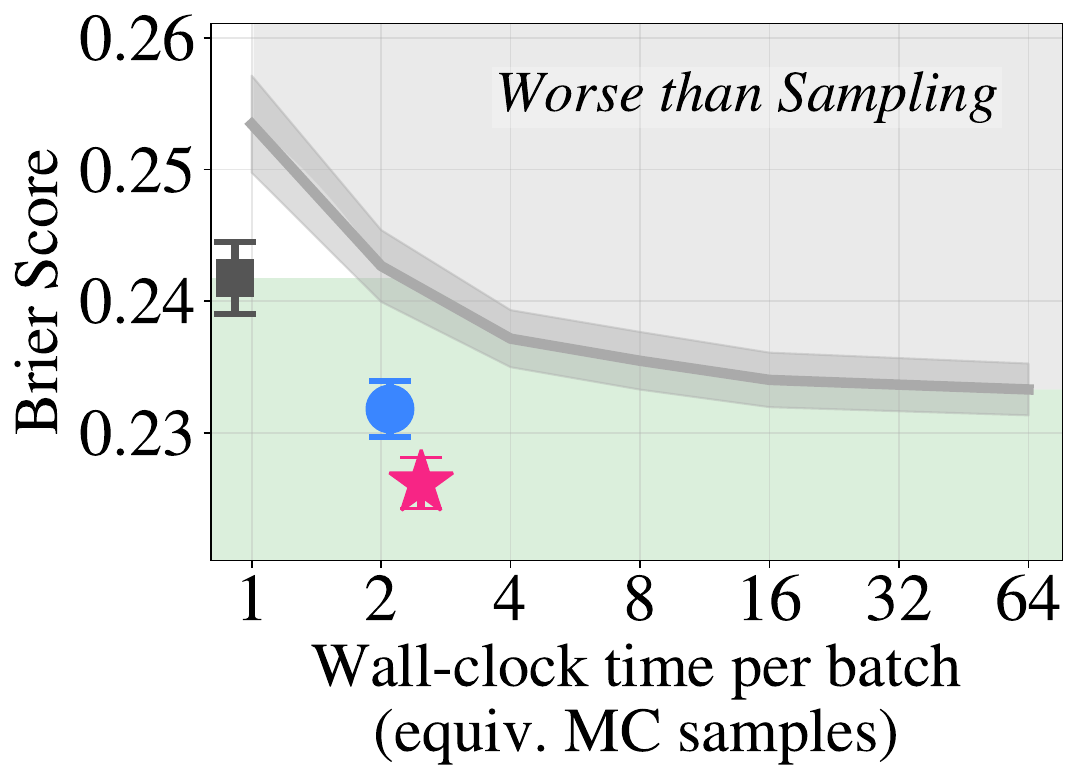}
    \end{subfigure}
    \hfill
    \begin{subfigure}[t]{0.32\textwidth}
        \centering
        \includegraphics[width=\textwidth]{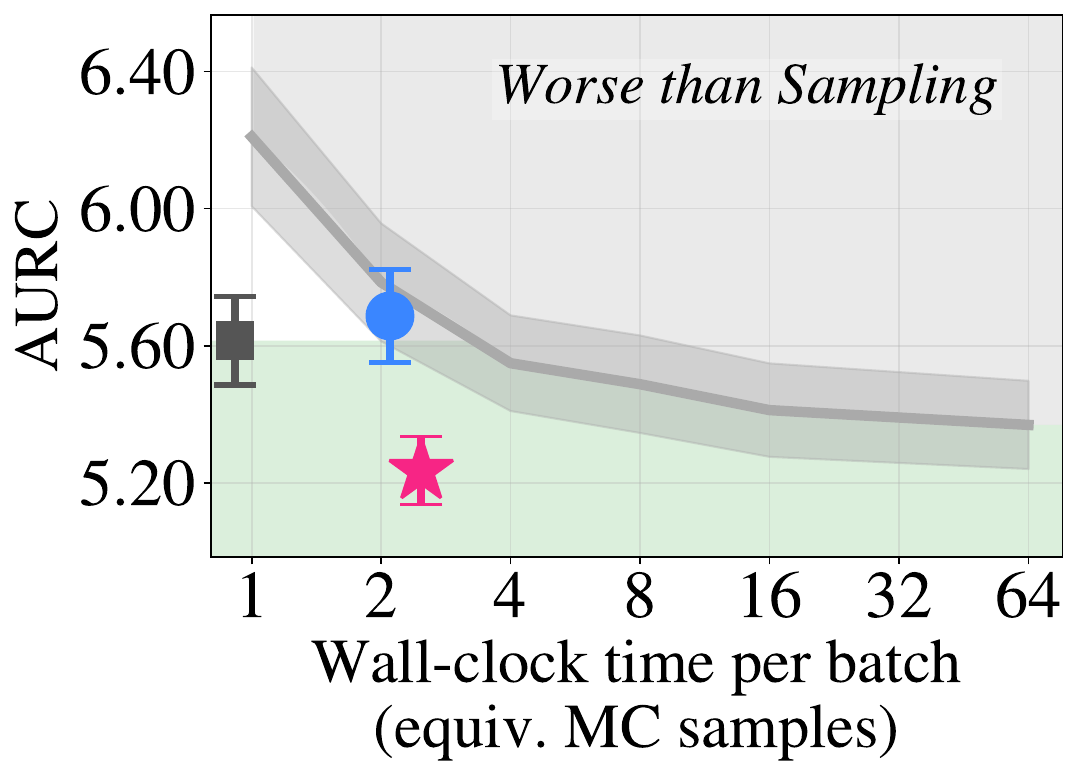}
    \end{subfigure}

    \vspace{1em}

    \begin{subfigure}[t]{0.32\textwidth}
        \centering
        \includegraphics[width=\textwidth]{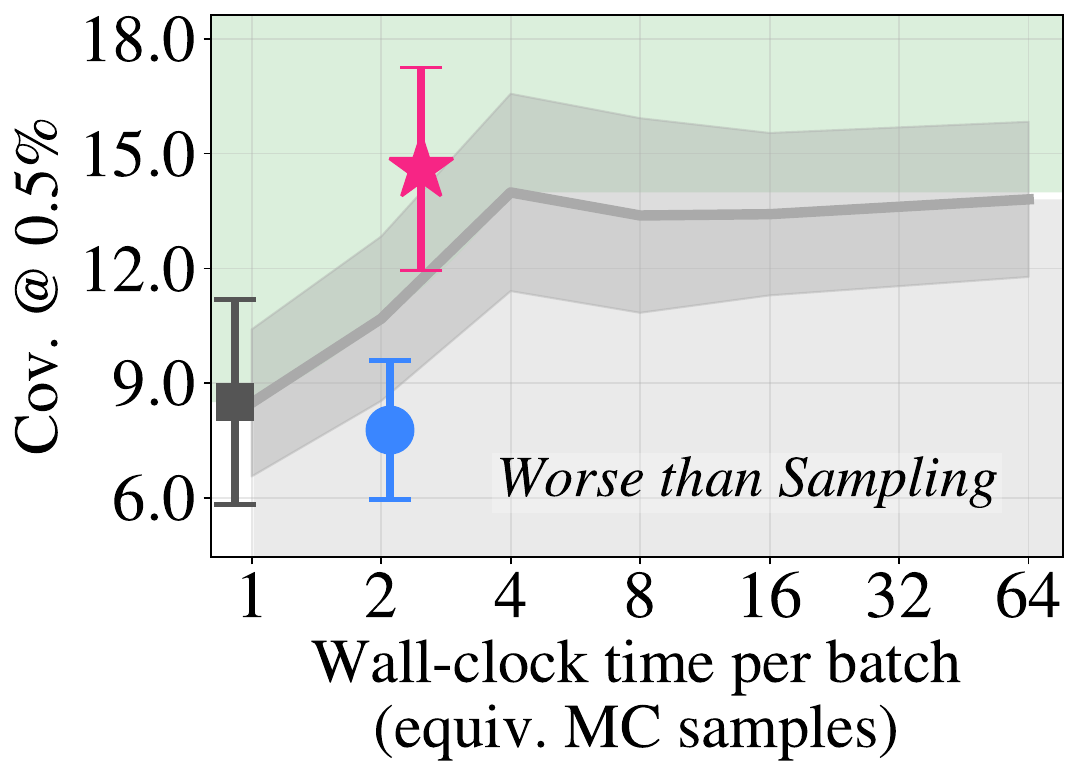}
    \end{subfigure}
    \hfill
    \begin{subfigure}[t]{0.32\textwidth}
        \centering
        \includegraphics[width=\textwidth]{new_figures/metrics/nlvr/C_0.01_errors.pdf}
    \end{subfigure}
    \hfill
    \begin{subfigure}[t]{0.32\textwidth}
        \centering
        \includegraphics[width=\textwidth]{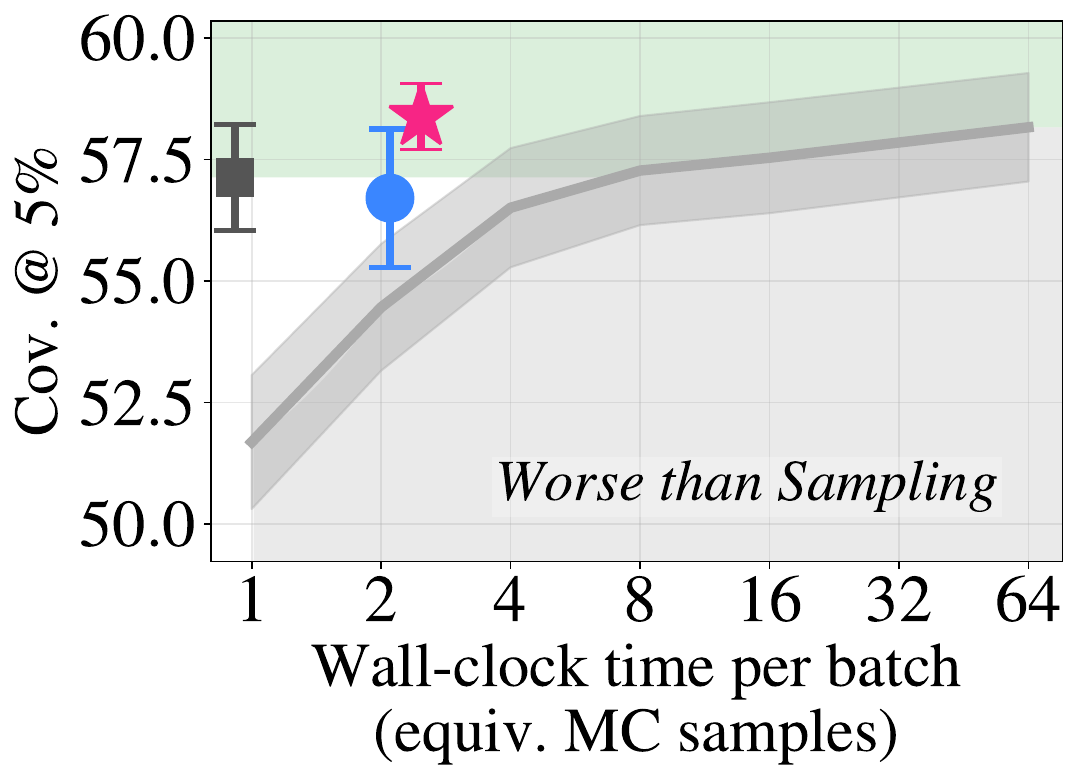}
    \end{subfigure}

    \vspace{1em}

    \begin{subfigure}[t]{0.32\textwidth}
        \centering
        \includegraphics[width=\textwidth]{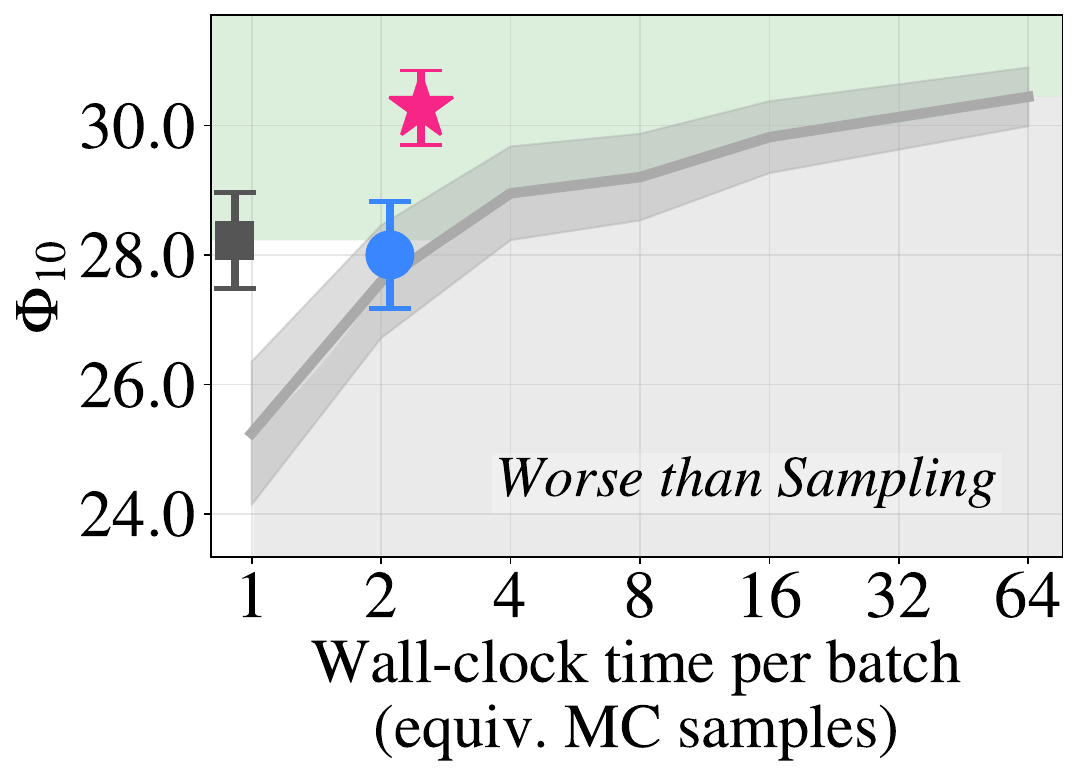}
    \end{subfigure}
    \hfill
    \begin{subfigure}[t]{0.32\textwidth}
        \centering
        \includegraphics[width=\textwidth]{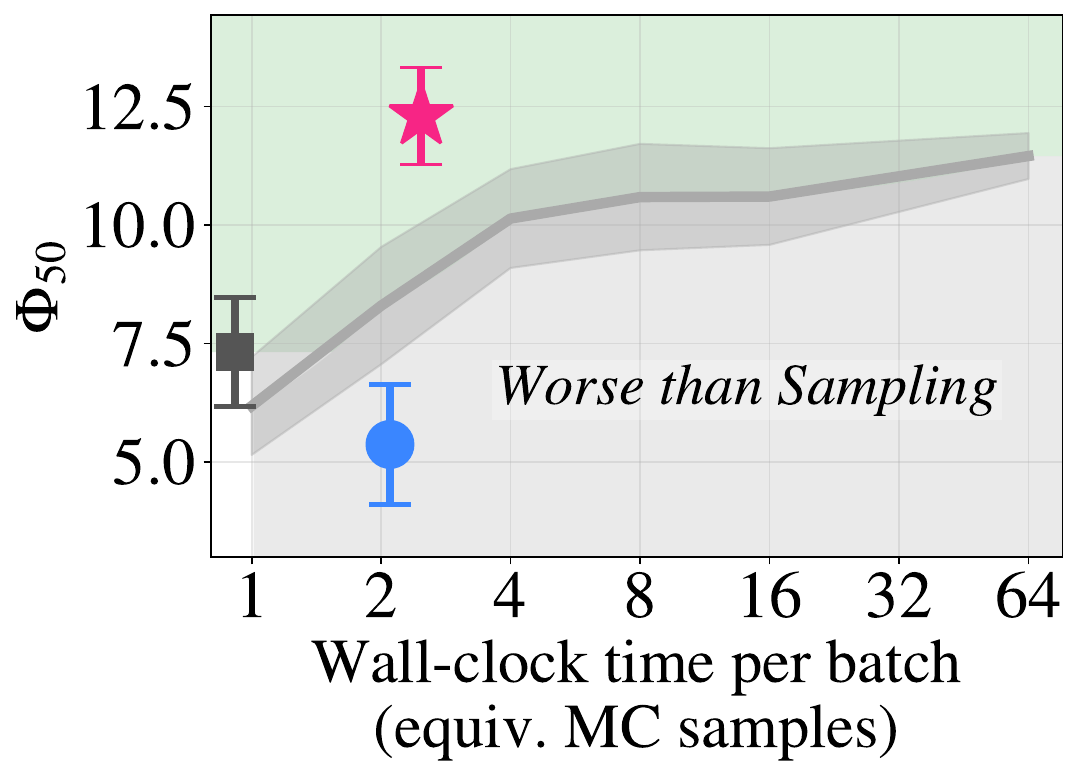}
    \end{subfigure}
    \hfill
    \begin{subfigure}[t]{0.32\textwidth}
        \centering
        \includegraphics[width=\textwidth]{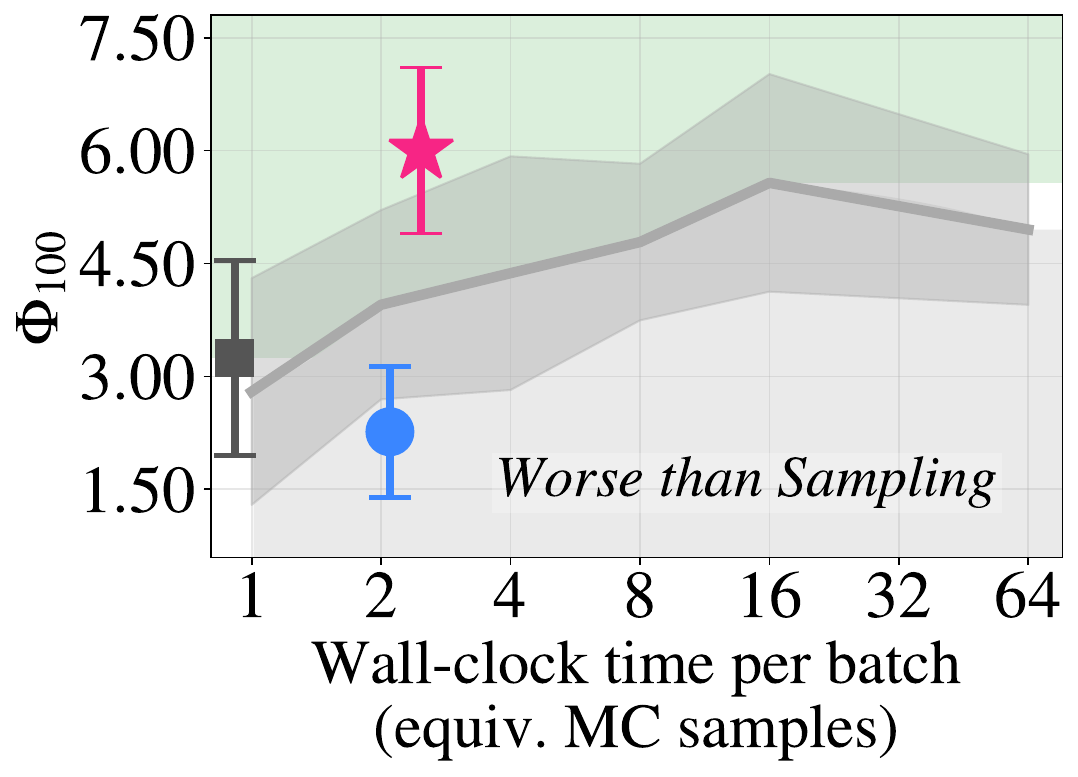}
    \end{subfigure}

    \vspace{0.3em}
    \includegraphics[width=0.6\textwidth]{new_figures/metrics/legend_horizontal.pdf}
    \caption{Results on Accuracy, Calibration, Loss, the Brier Score and Selective Prediction for BEiT-Base on NLVR2. The \colorbox{Paretogreen!20}{Pareto-dominating region} (vs. the mean network and MC Sampling) is highlighted in green. SEM (Standard Error of the Mean) across 5 random seeds is shown.}
    \label{fig:full_nlvr}
\end{figure}

\begin{figure}[t]
    \centering
    \begin{subfigure}[t]{0.32\textwidth}
        \centering
        \includegraphics[width=\textwidth]{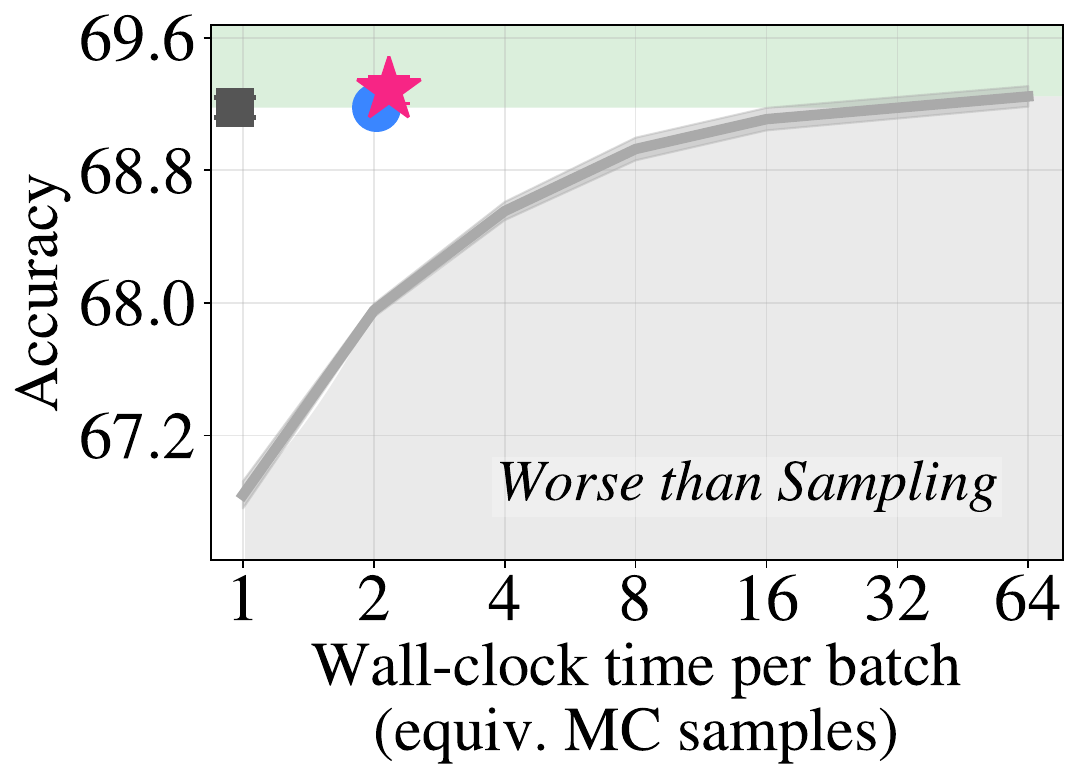}
    \end{subfigure}
    \hfill
    \begin{subfigure}[t]{0.32\textwidth}
        \centering
        \includegraphics[width=\textwidth]{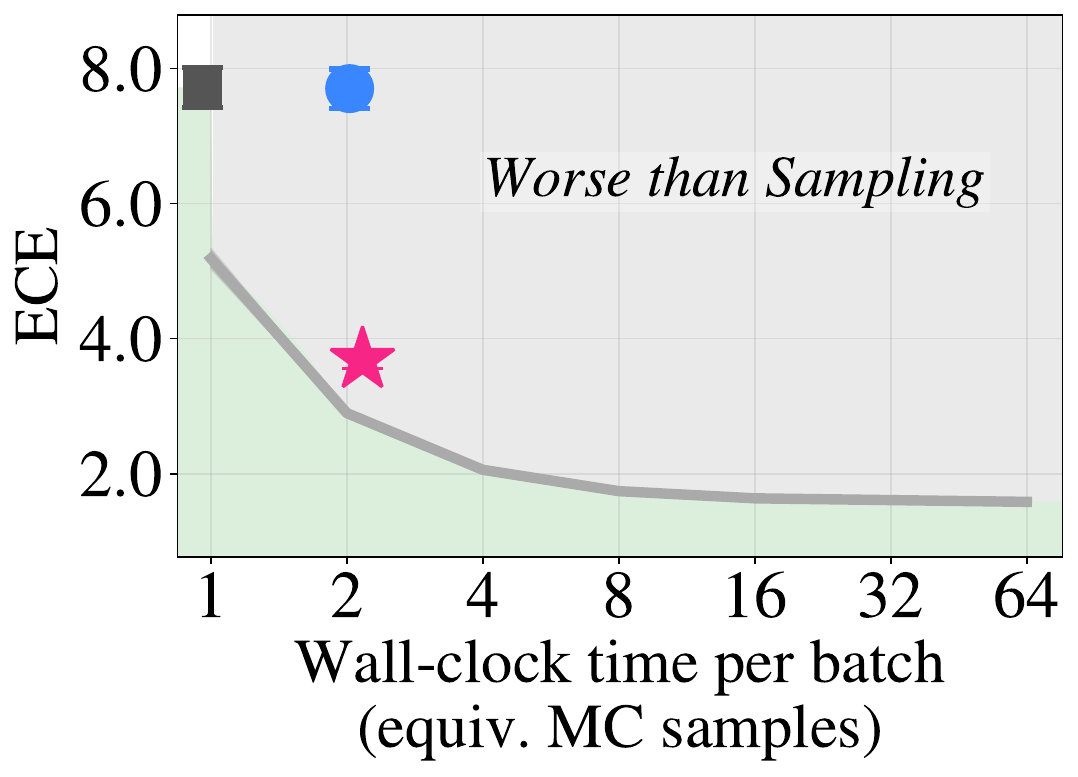}
    \end{subfigure}
    \hfill
    \begin{subfigure}[t]{0.32\textwidth}
        \centering
        \includegraphics[width=\textwidth]{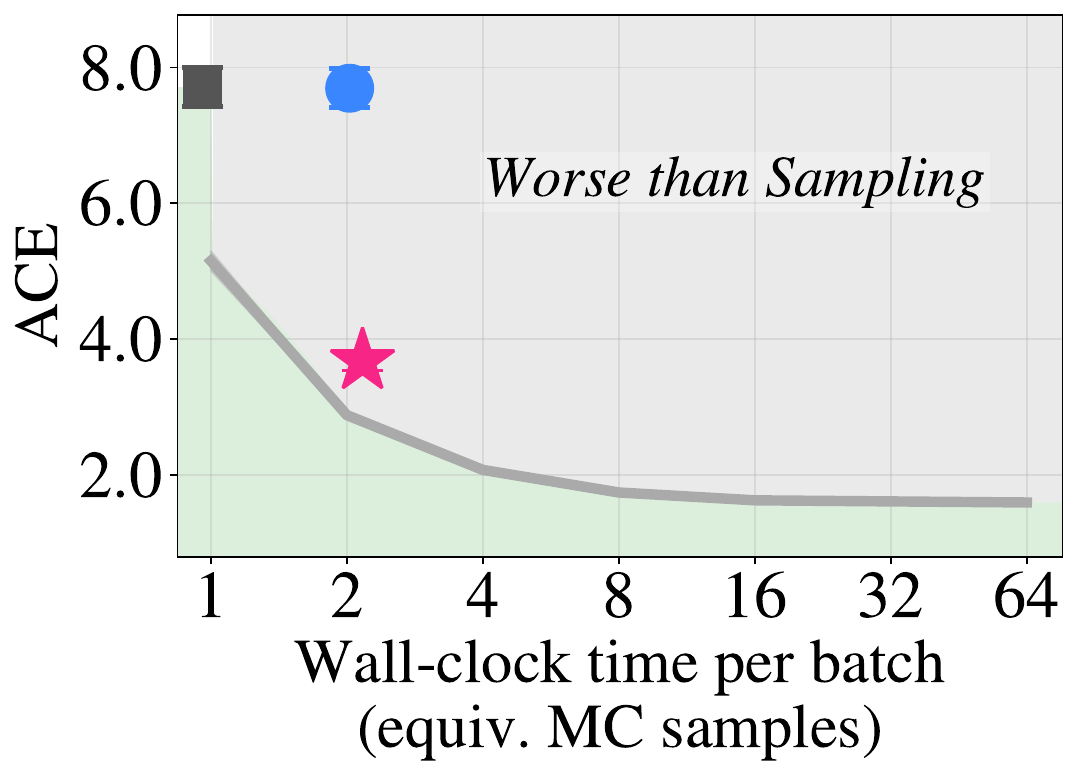}
    \end{subfigure}

    \vspace{1em}  %

    \begin{subfigure}[t]{0.32\textwidth}
        \centering
        \includegraphics[width=\textwidth]{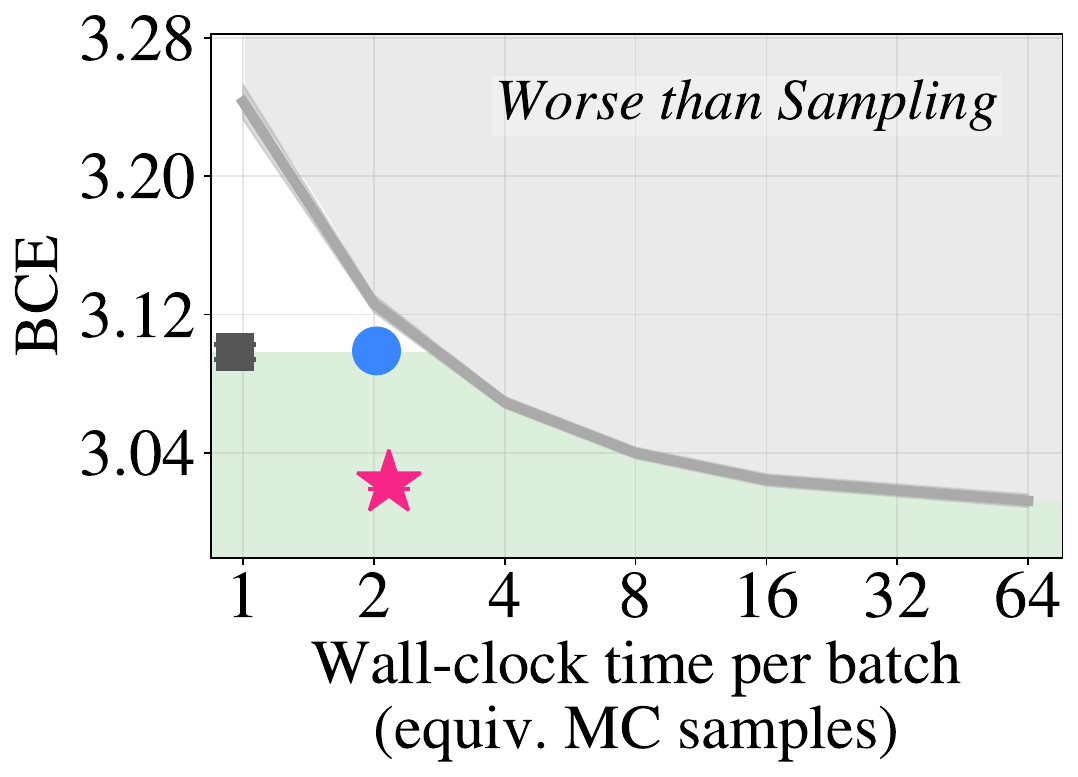}
    \end{subfigure}
    \hfill
    \begin{subfigure}[t]{0.32\textwidth}
        \centering
        \includegraphics[width=\textwidth]{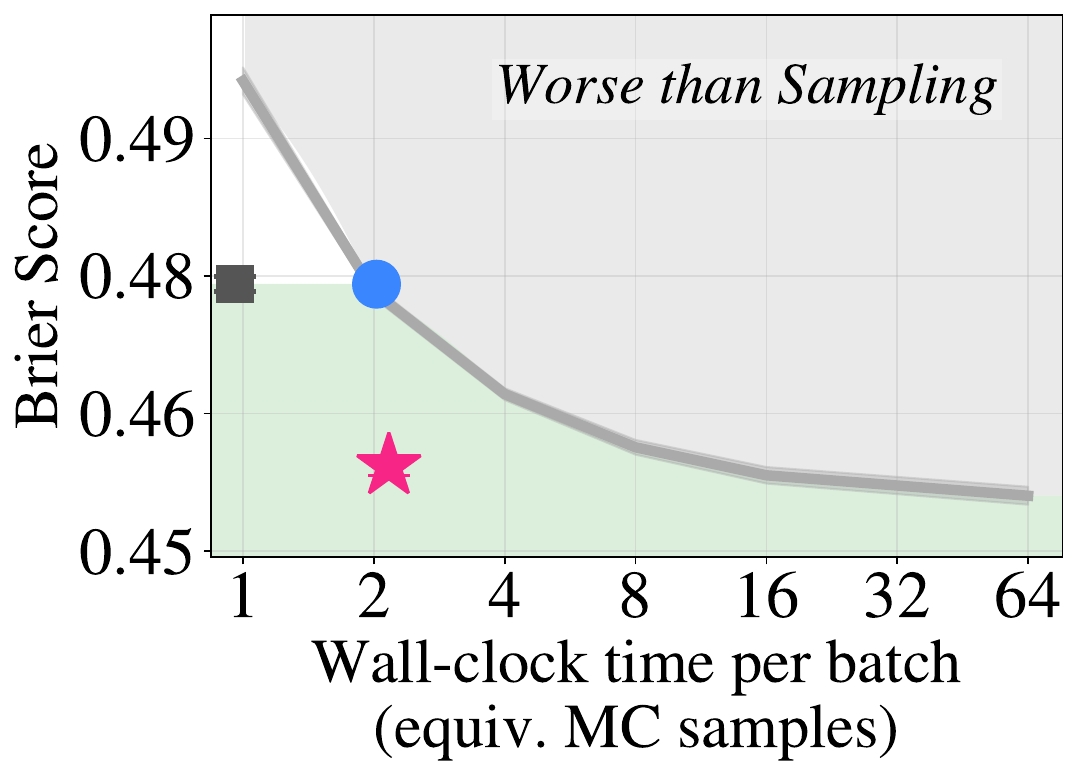}
    \end{subfigure}
    \hfill
    \begin{subfigure}[t]{0.32\textwidth}
        \centering
        \includegraphics[width=\textwidth]{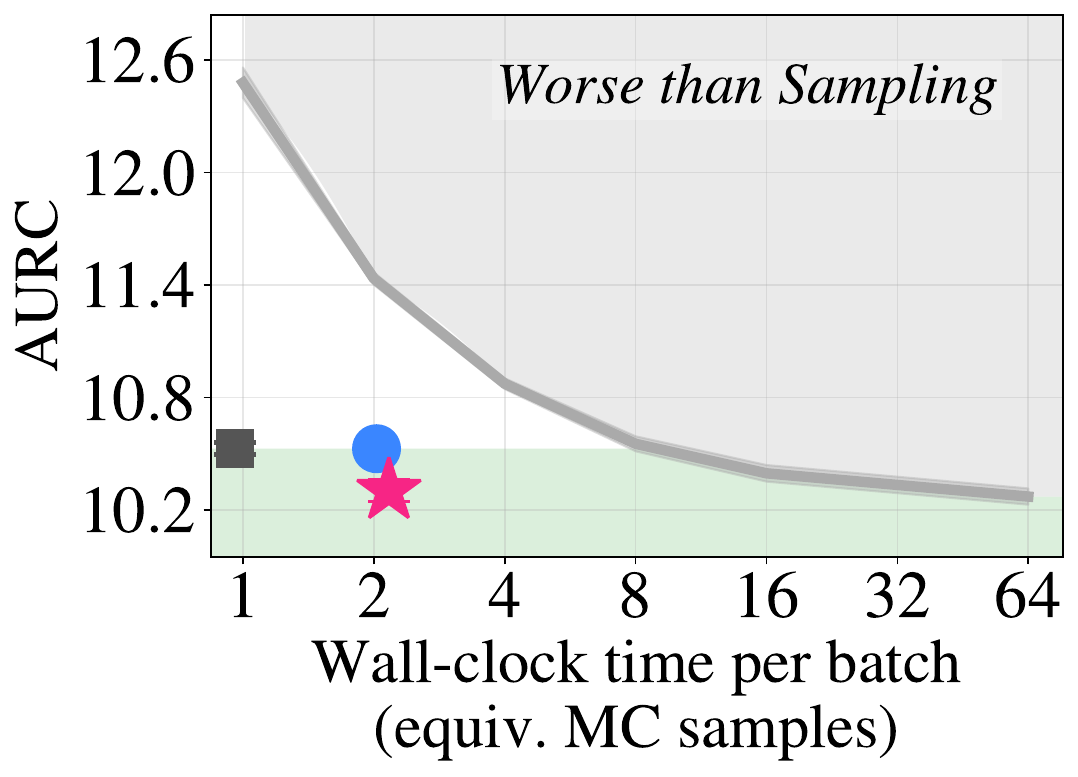}
    \end{subfigure}

    \vspace{1em}

    \begin{subfigure}[t]{0.32\textwidth}
        \centering
        \includegraphics[width=\textwidth]{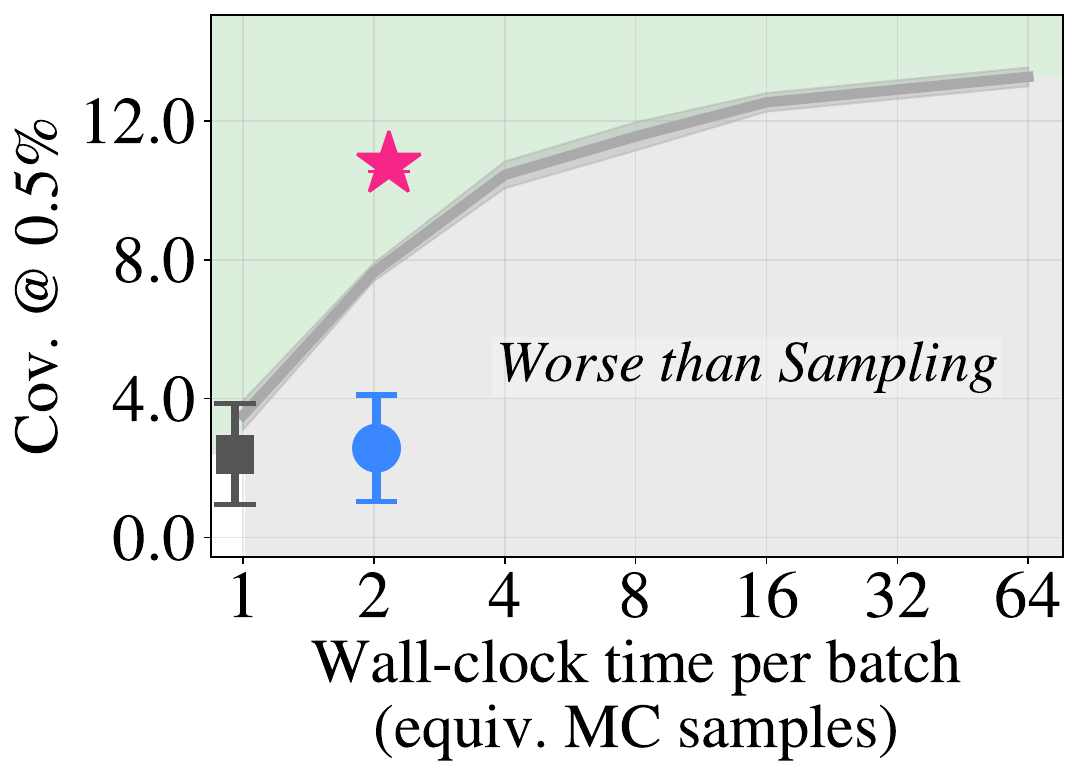}
    \end{subfigure}
    \hfill
    \begin{subfigure}[t]{0.32\textwidth}
        \centering
        \includegraphics[width=\textwidth]{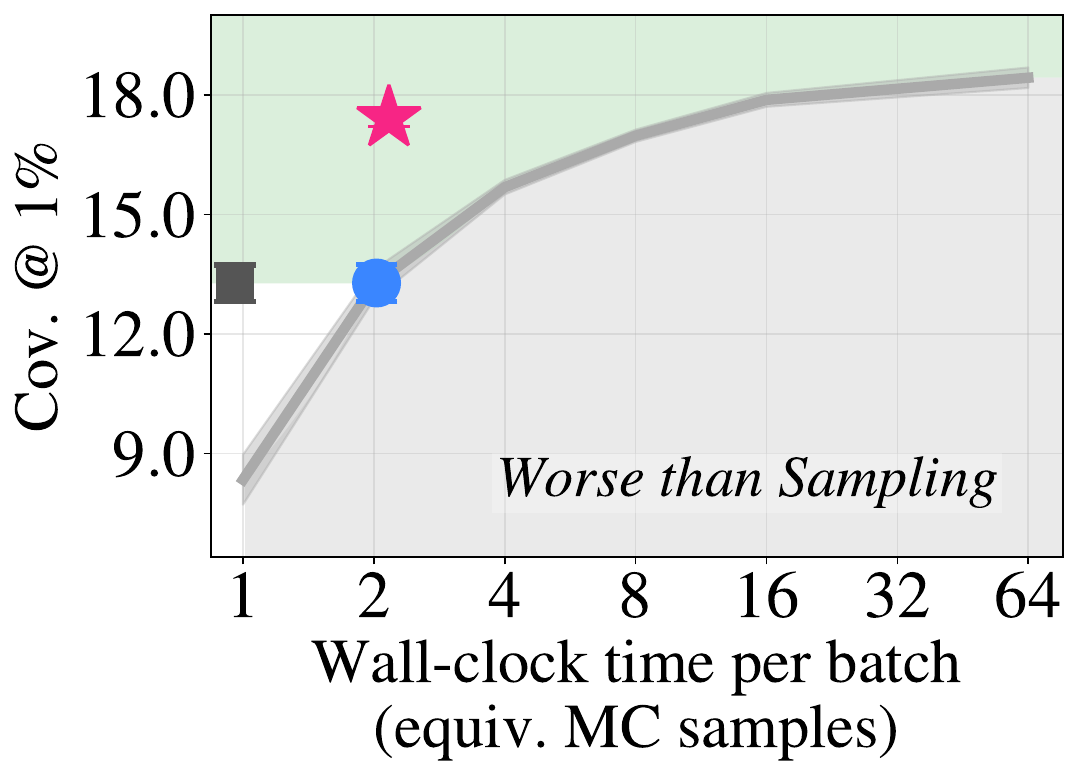}
    \end{subfigure}
    \hfill
    \begin{subfigure}[t]{0.32\textwidth}
        \centering
        \includegraphics[width=\textwidth]{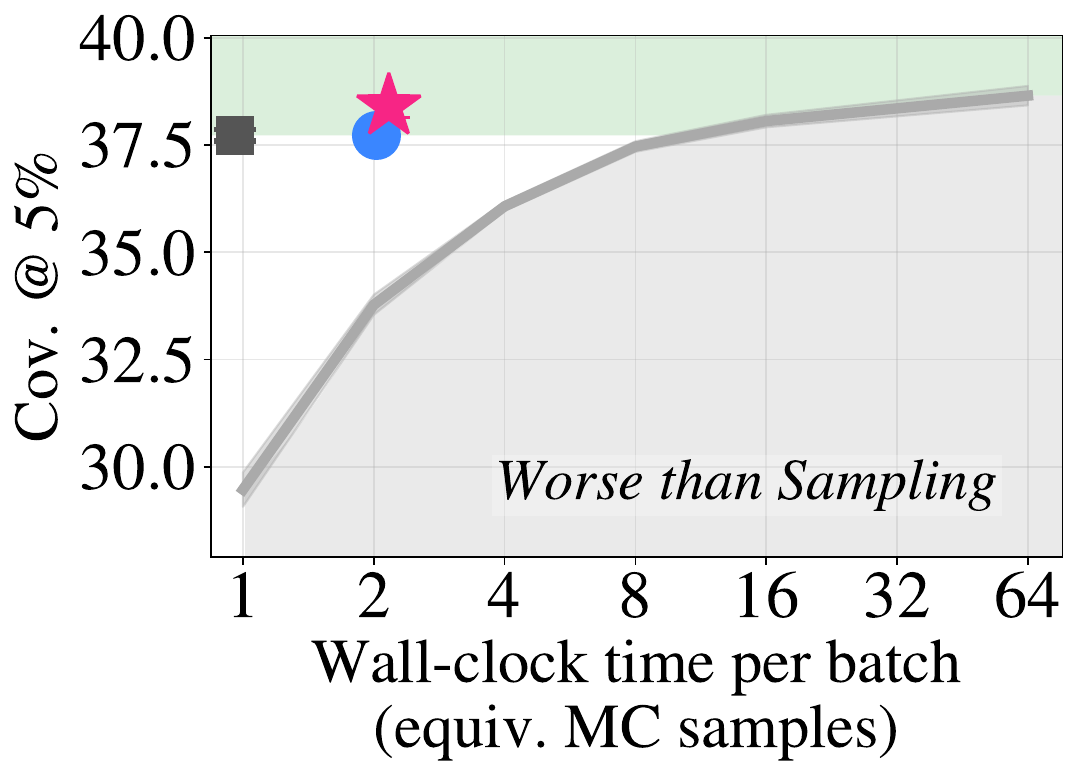}
    \end{subfigure}

    \vspace{1em}

    \begin{subfigure}[t]{0.32\textwidth}
        \centering
        \includegraphics[width=\textwidth]{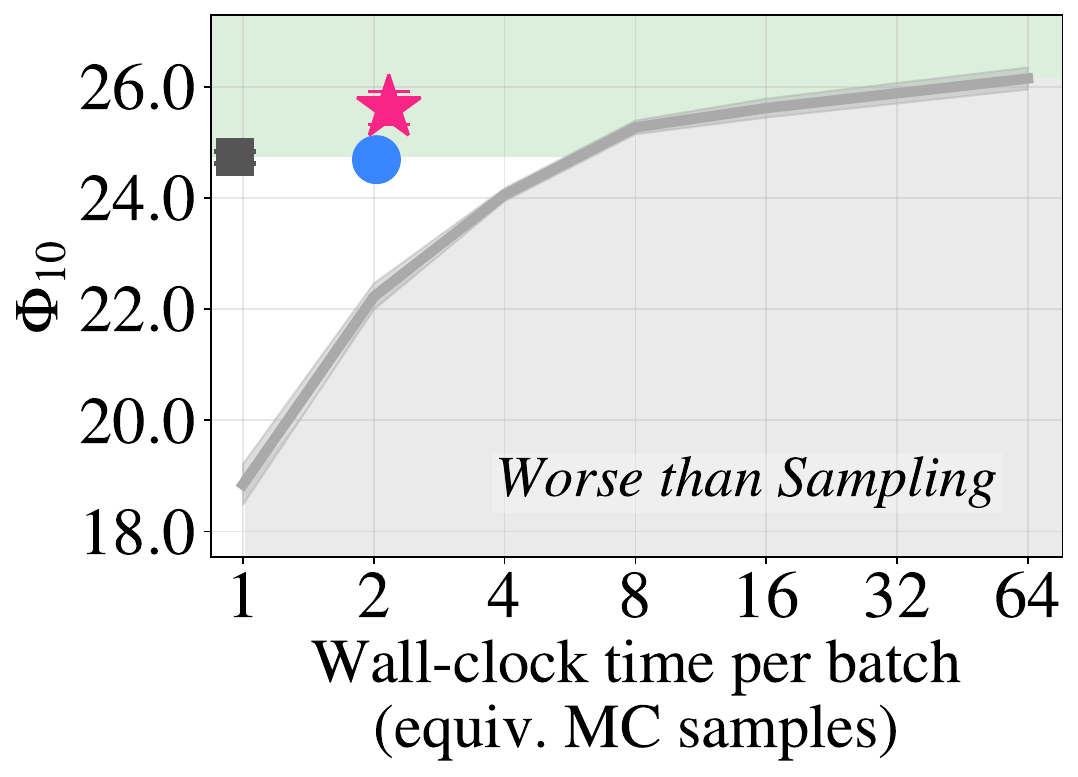}
    \end{subfigure}
    \hfill
    \begin{subfigure}[t]{0.32\textwidth}
        \centering
        \includegraphics[width=\textwidth]{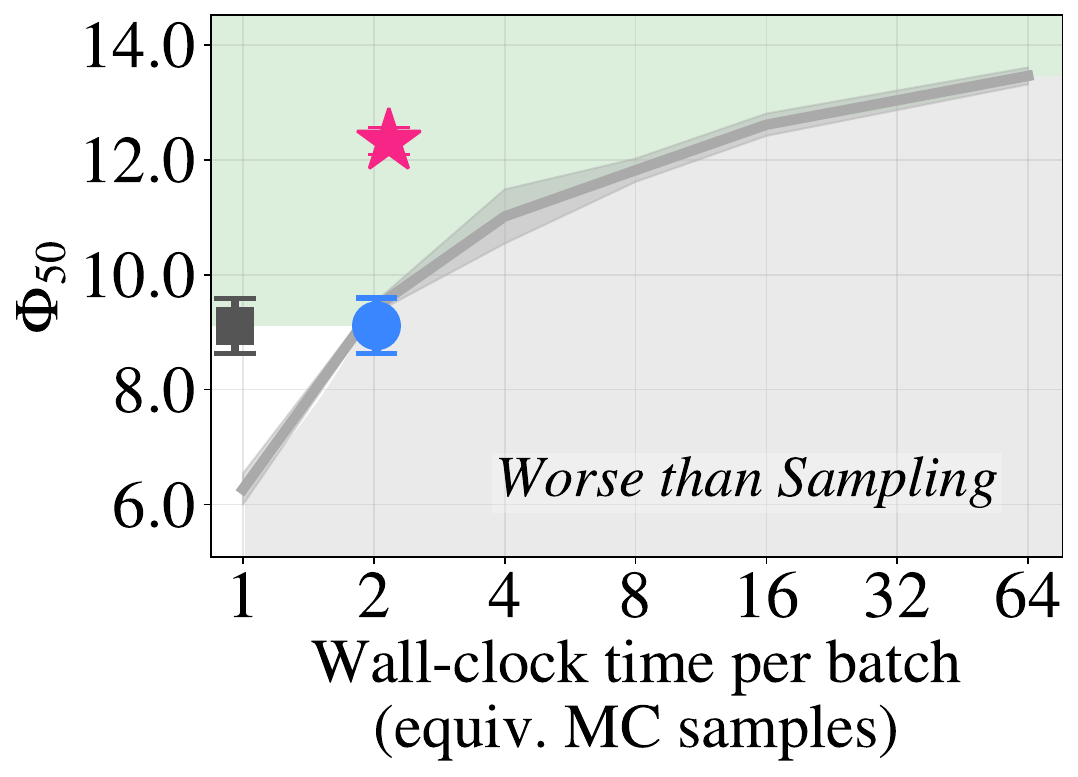}
    \end{subfigure}
    \hfill
    \begin{subfigure}[t]{0.32\textwidth}
        \centering
        \includegraphics[width=\textwidth]{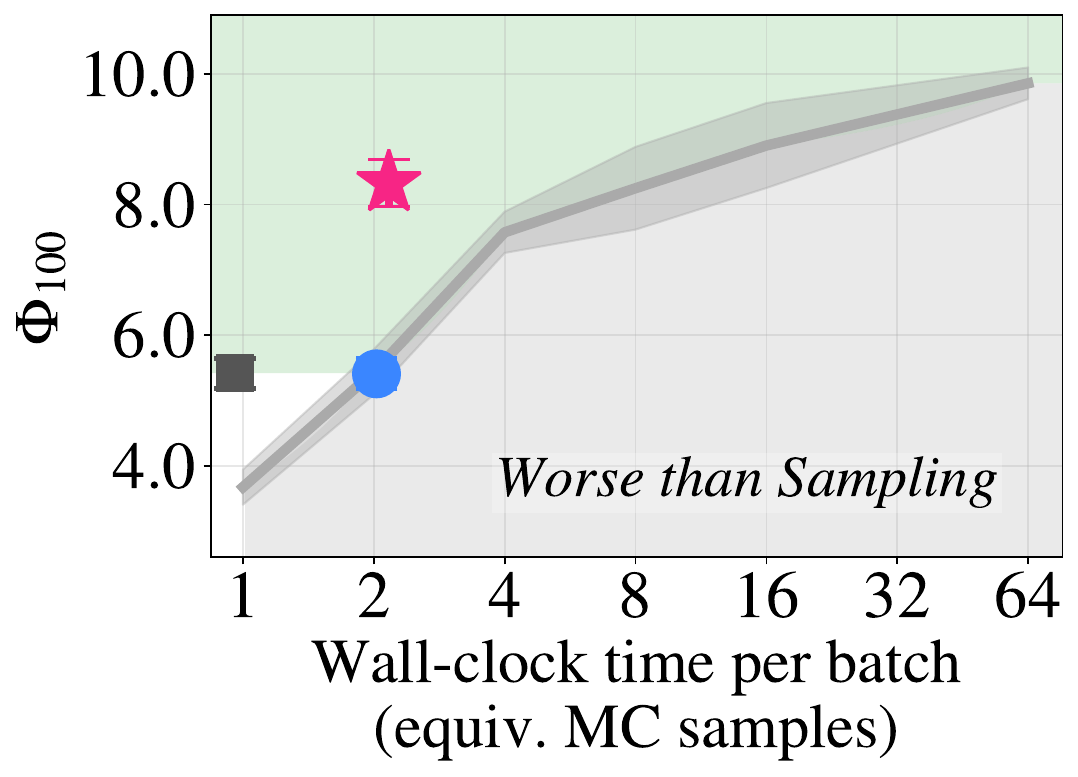}
    \end{subfigure}

    \vspace{0.3em}
    \includegraphics[width=0.6\textwidth]{new_figures/metrics/legend_horizontal.pdf}
    \caption{Results on Accuracy, Calibration, Loss, the Brier Score and Selective Prediction for ViLT on VQAv2. The \colorbox{Paretogreen!20}{Pareto-dominating region} (vs. the mean network and MC Sampling) is highlighted in green. SEM (Standard Error of the Mean) across 5 random seeds is shown.}
    \label{fig:full_viltvqa}
\end{figure}

\begin{figure}[t]
    \centering
    \begin{subfigure}[t]{0.32\textwidth}
        \centering
        \includegraphics[width=\textwidth]{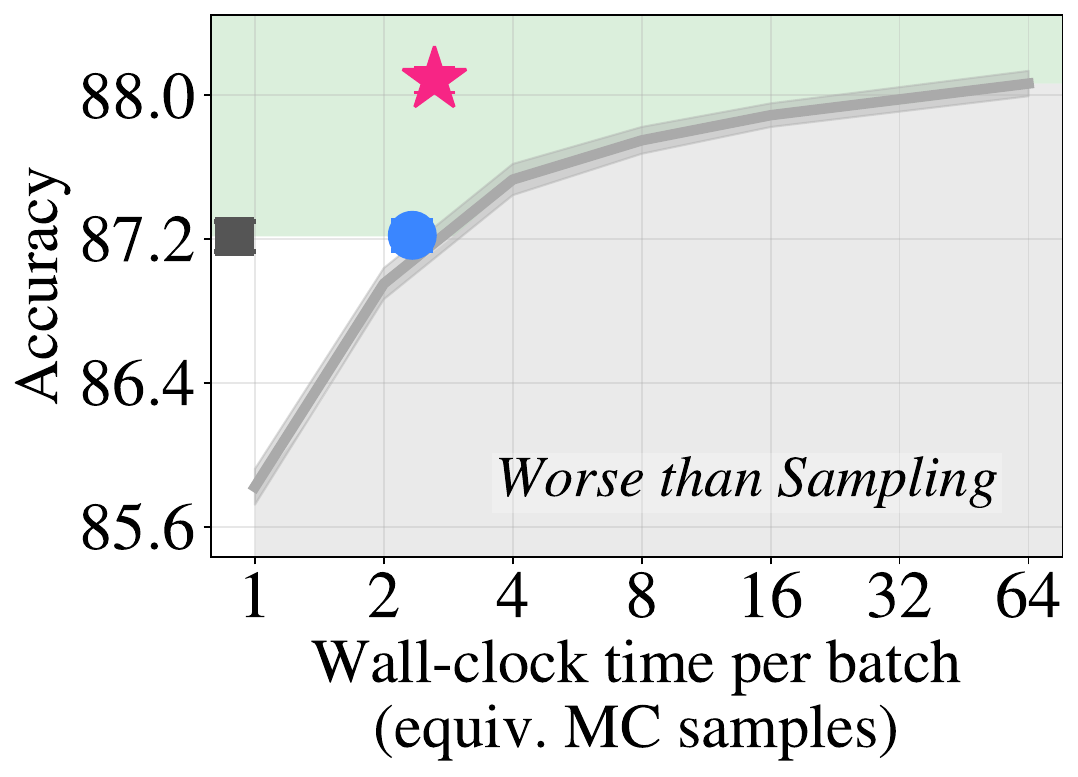}
    \end{subfigure}
    \hfill
    \begin{subfigure}[t]{0.32\textwidth}
        \centering
        \includegraphics[width=\textwidth]{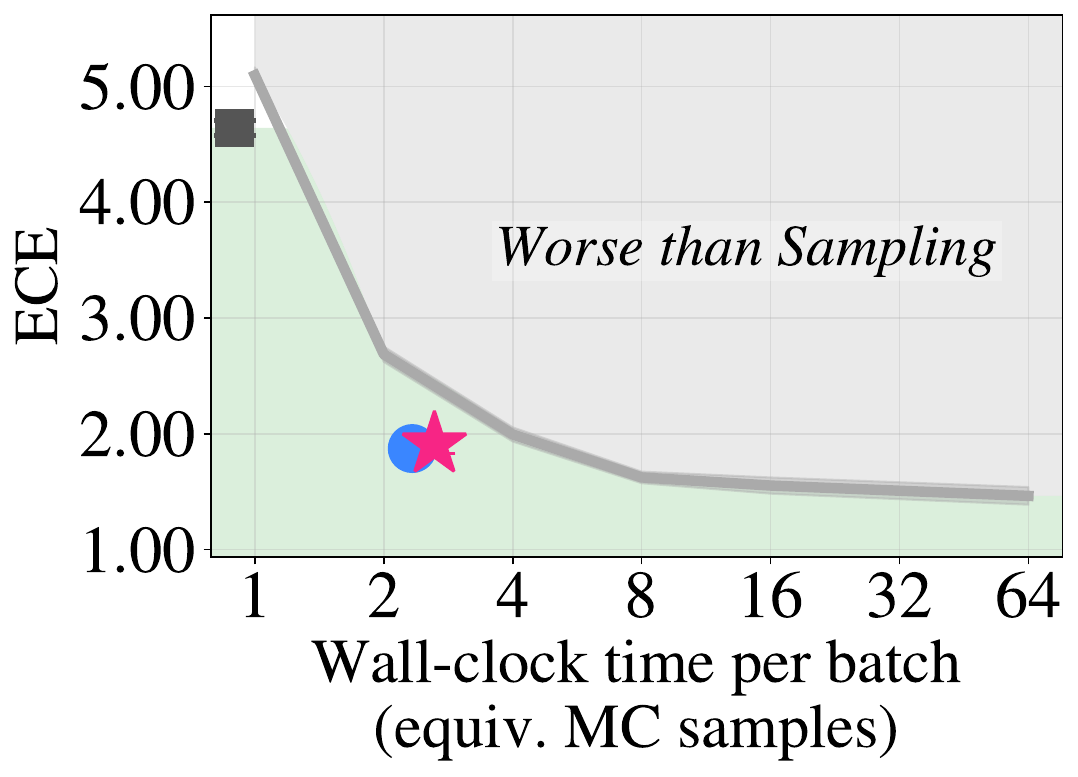}
    \end{subfigure}
    \hfill
    \begin{subfigure}[t]{0.32\textwidth}
        \centering
        \includegraphics[width=\textwidth]{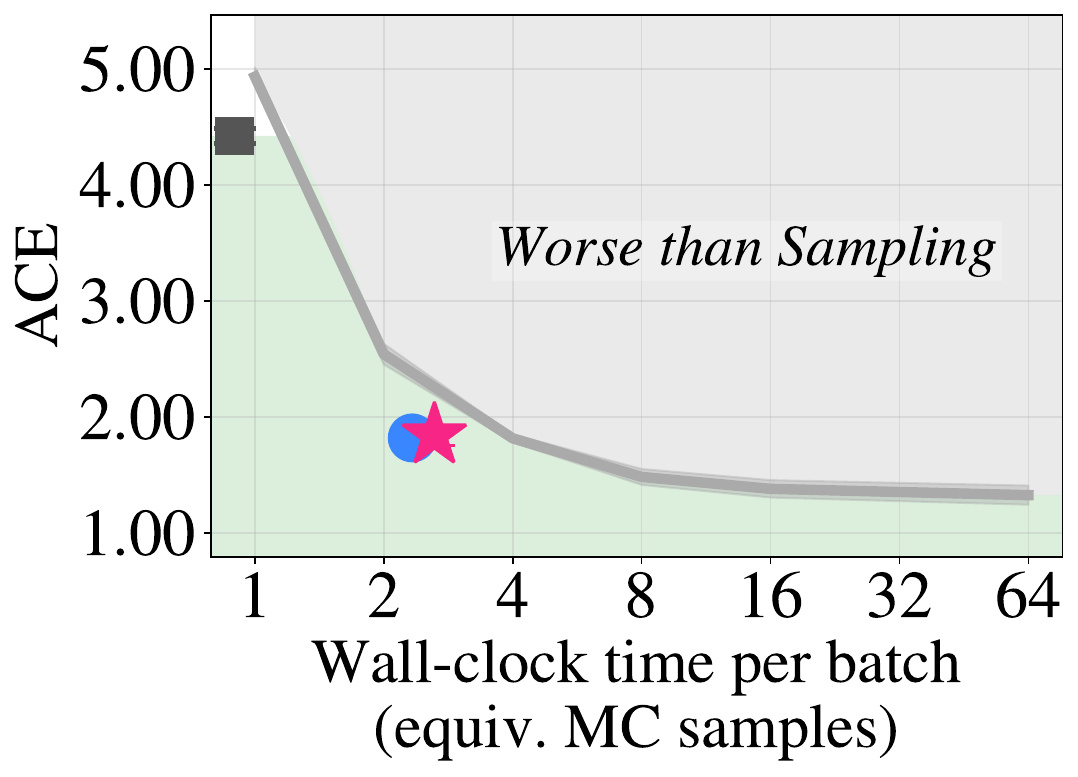}
    \end{subfigure}

    \vspace{1em}  %

    \begin{subfigure}[t]{0.32\textwidth}
        \centering
        \includegraphics[width=\textwidth]{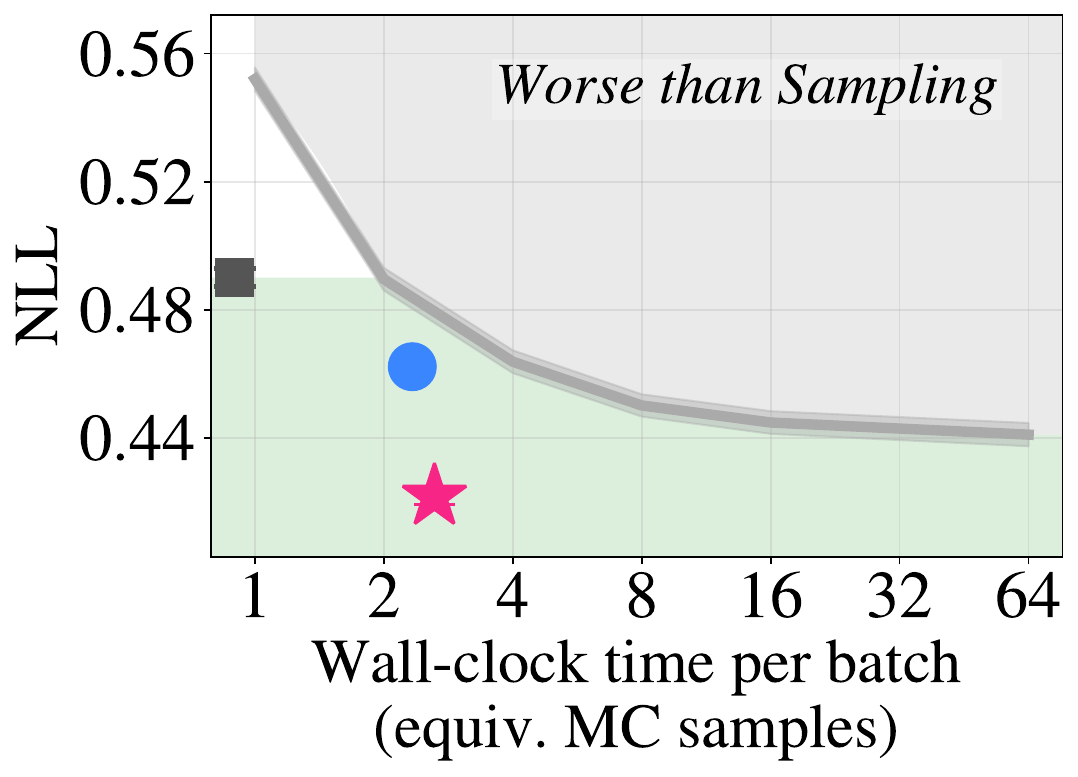}
    \end{subfigure}
    \hfill
    \begin{subfigure}[t]{0.32\textwidth}
        \centering
        \includegraphics[width=\textwidth]{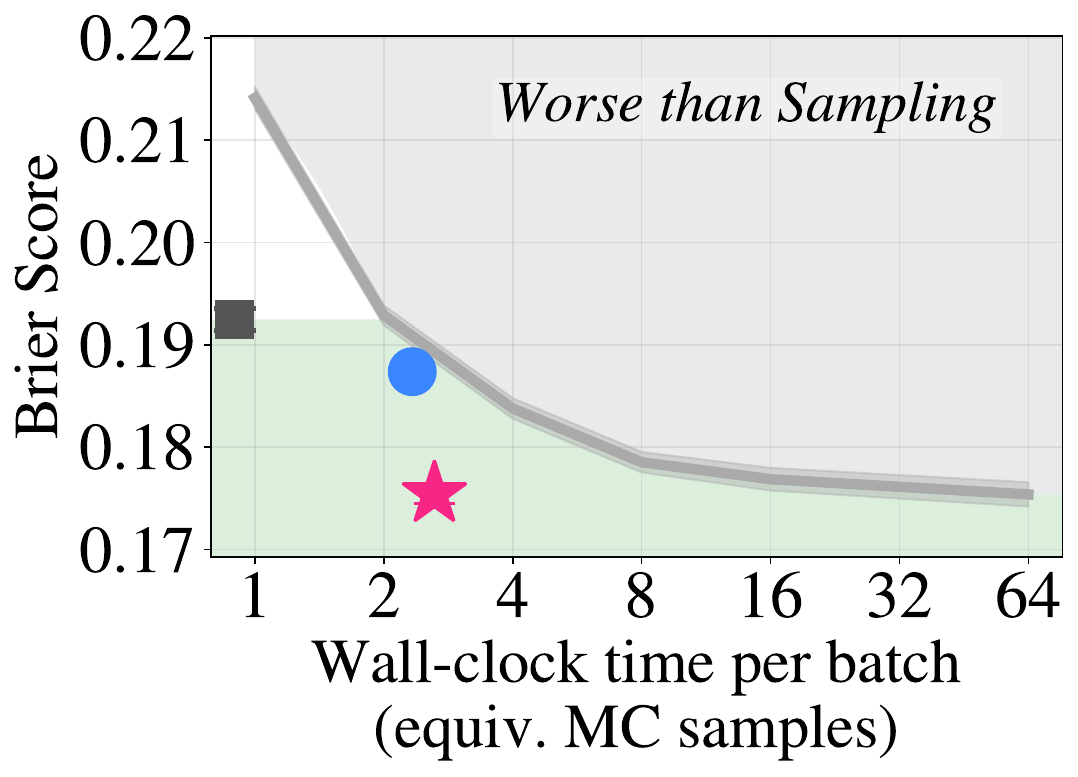}
    \end{subfigure}
    \hfill
    \begin{subfigure}[t]{0.32\textwidth}
        \centering
        \includegraphics[width=\textwidth]{new_figures/metrics/cifar100/auc_errors.pdf}
    \end{subfigure}

    \vspace{1em}

    \begin{subfigure}[t]{0.32\textwidth}
        \centering
        \includegraphics[width=\textwidth]{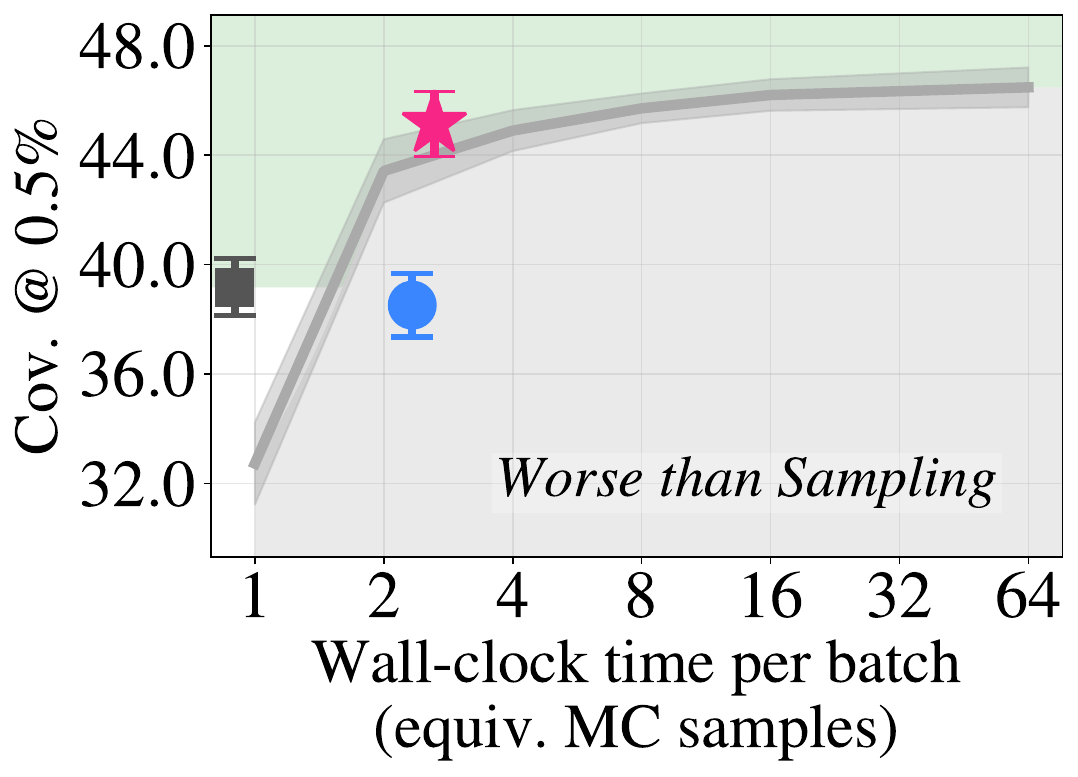}
    \end{subfigure}
    \hfill
    \begin{subfigure}[t]{0.32\textwidth}
        \centering
        \includegraphics[width=\textwidth]{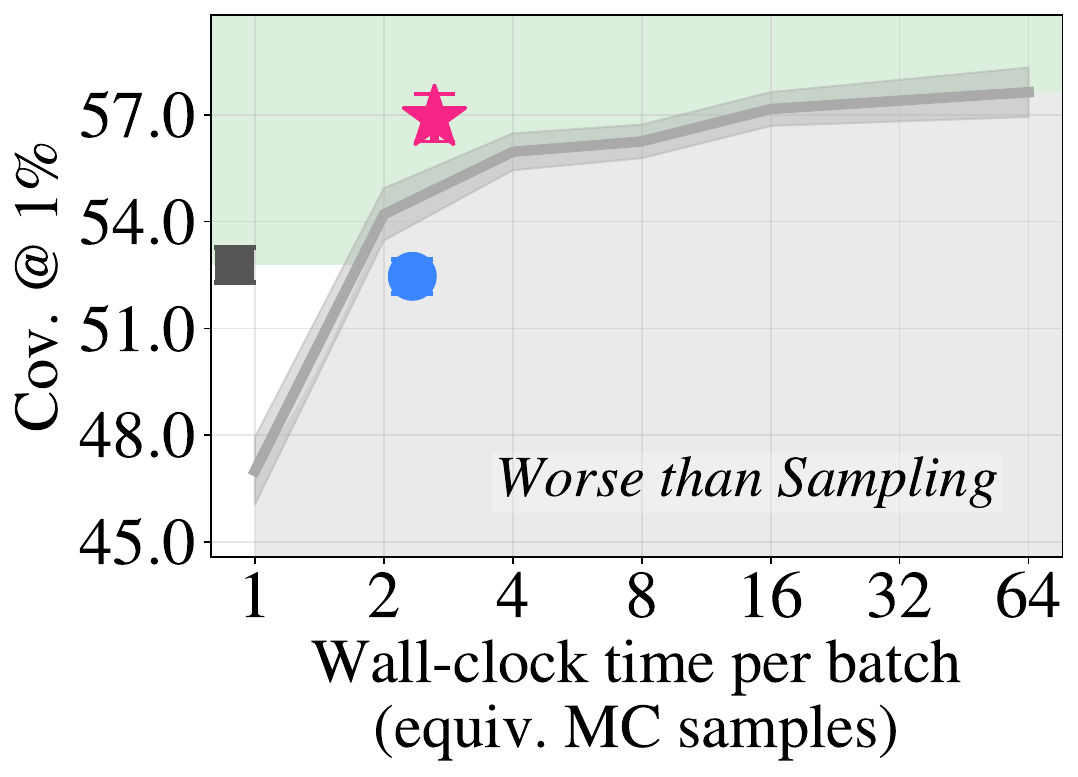}
    \end{subfigure}
    \hfill
    \begin{subfigure}[t]{0.32\textwidth}
        \centering
        \includegraphics[width=\textwidth]{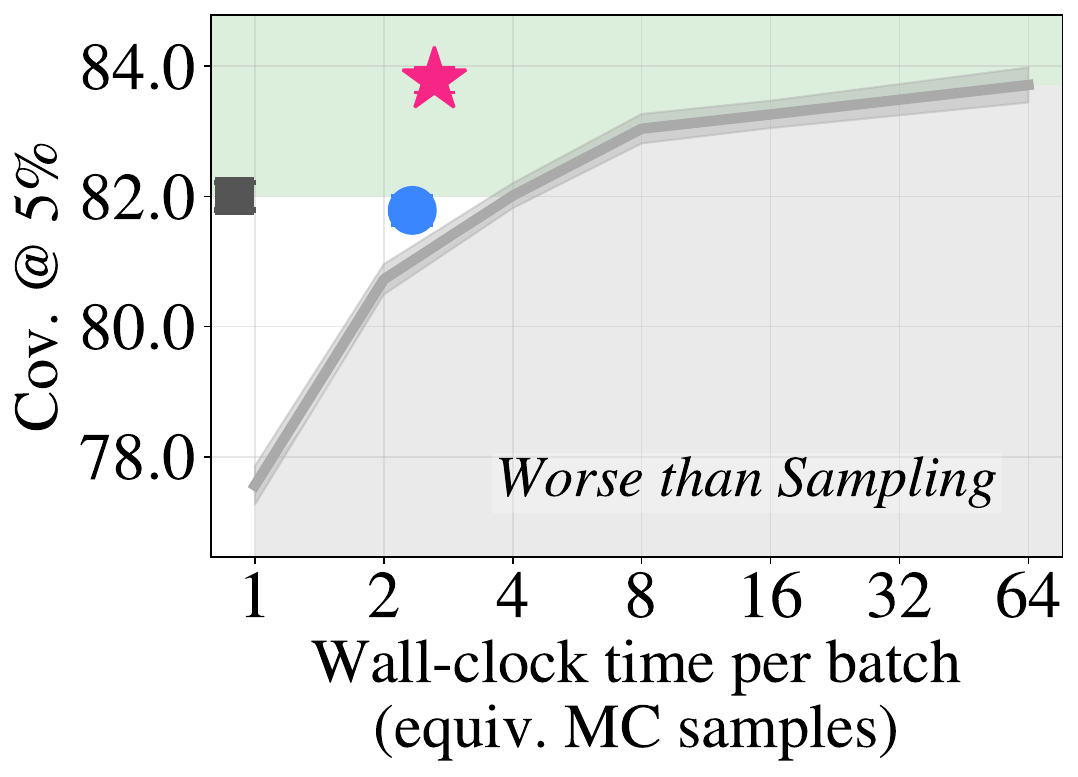}
    \end{subfigure}

    \vspace{1em}

    \begin{subfigure}[t]{0.32\textwidth}
        \centering
        \includegraphics[width=\textwidth]{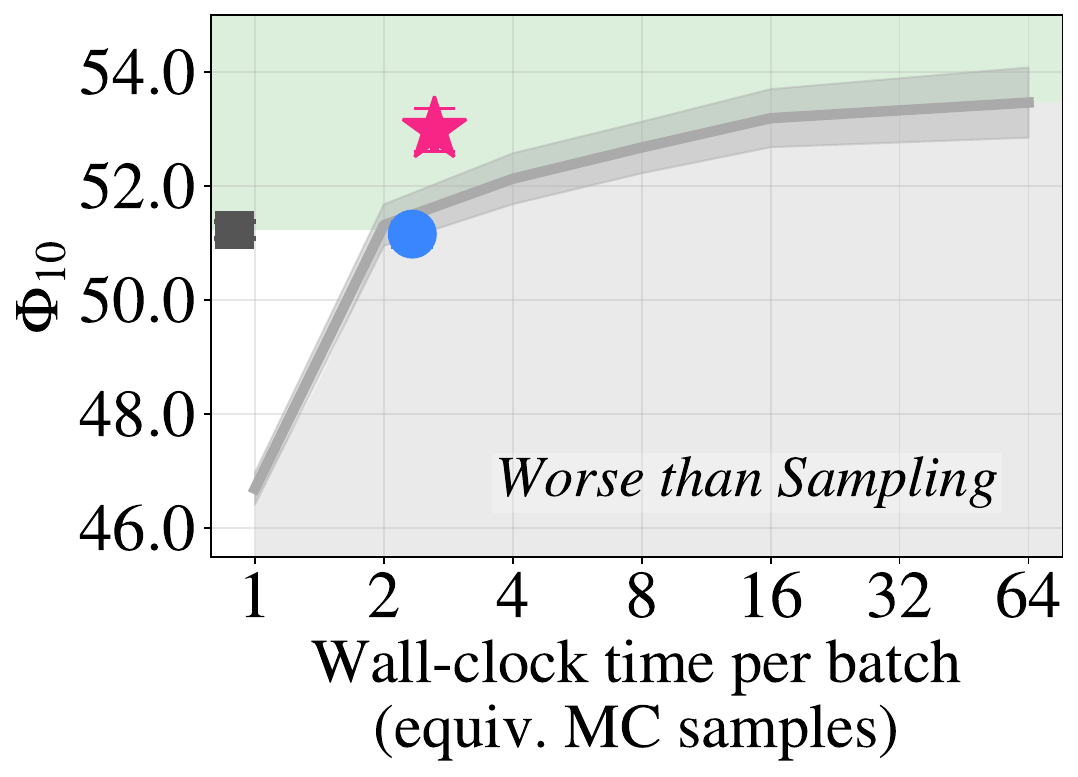}
    \end{subfigure}
    \hfill
    \begin{subfigure}[t]{0.32\textwidth}
        \centering
        \includegraphics[width=\textwidth]{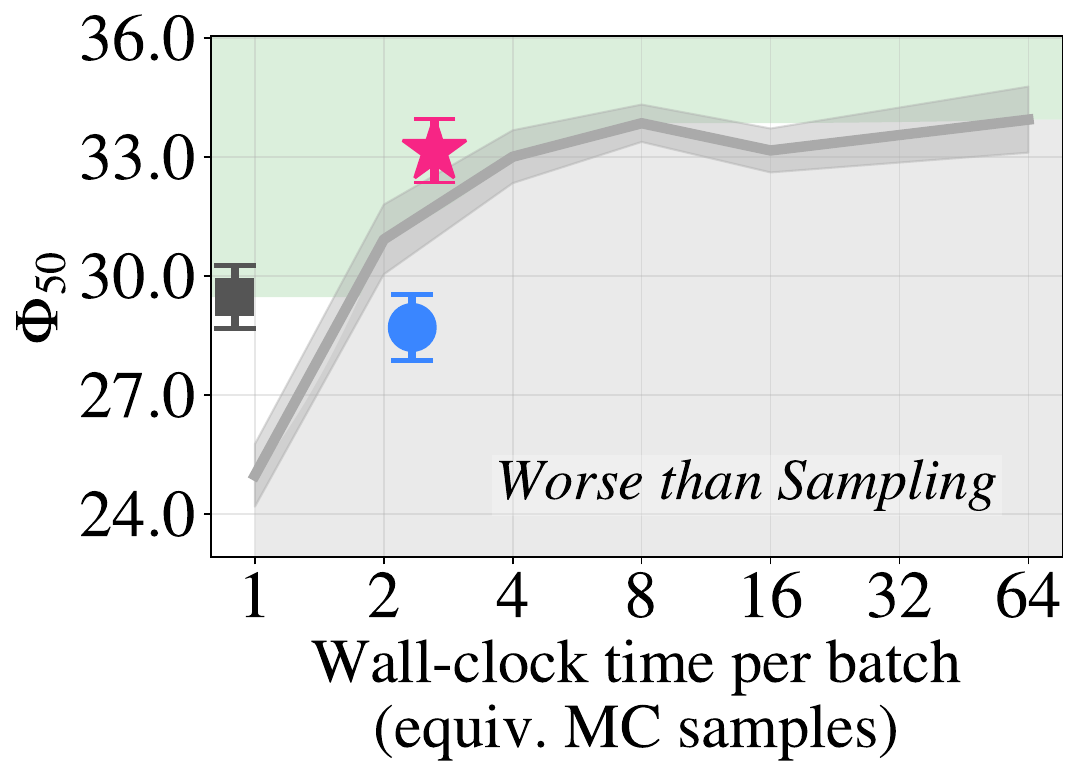}
    \end{subfigure}
    \hfill
    \begin{subfigure}[t]{0.32\textwidth}
        \centering
        \includegraphics[width=\textwidth]{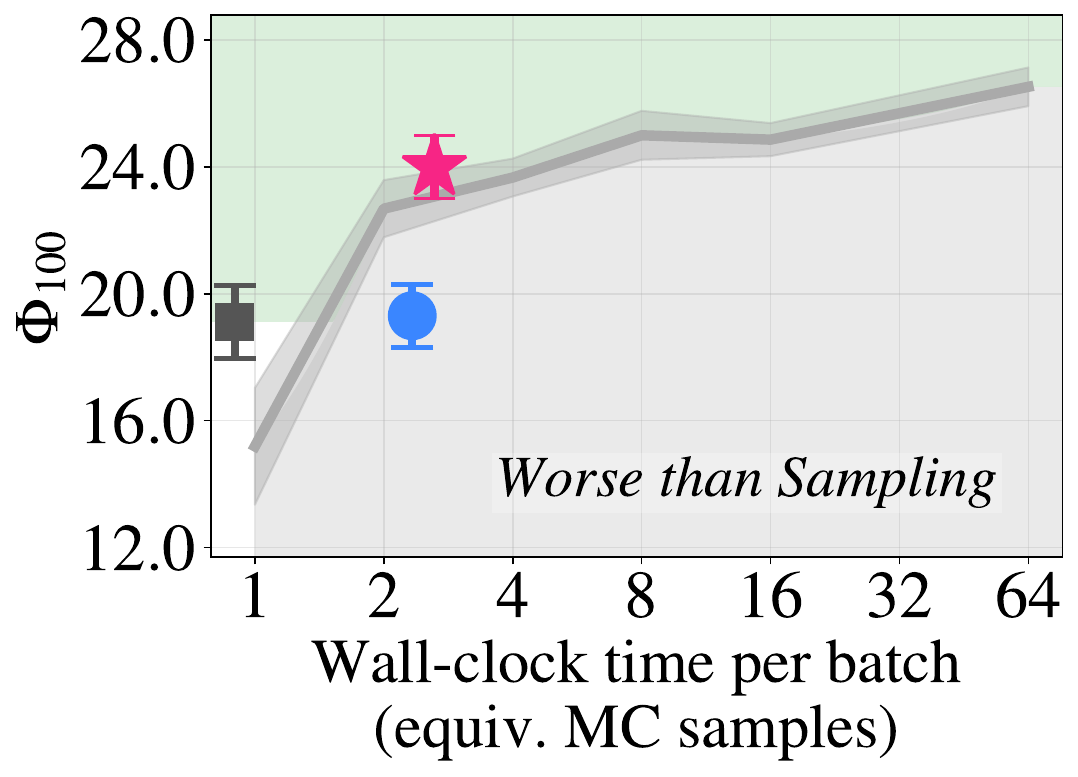}
    \end{subfigure}

    \vspace{0.3em}
    \includegraphics[width=0.6\textwidth]{new_figures/metrics/legend_horizontal.pdf}
    \caption{Results on Accuracy, Calibration, Loss, the Brier Score and Selective Prediction for ViT-Base on CIFAR-100. The \colorbox{Paretogreen!20}{Pareto-dominating region} (vs. the mean network and MC Sampling) is highlighted in green. SEM (Standard Error of the Mean) across 10 random seeds is shown.}
    \label{fig:full_cifar100}
\end{figure}

\begin{figure}[t]
    \centering
    \begin{subfigure}[t]{0.32\textwidth}
        \centering
        \includegraphics[width=\textwidth]{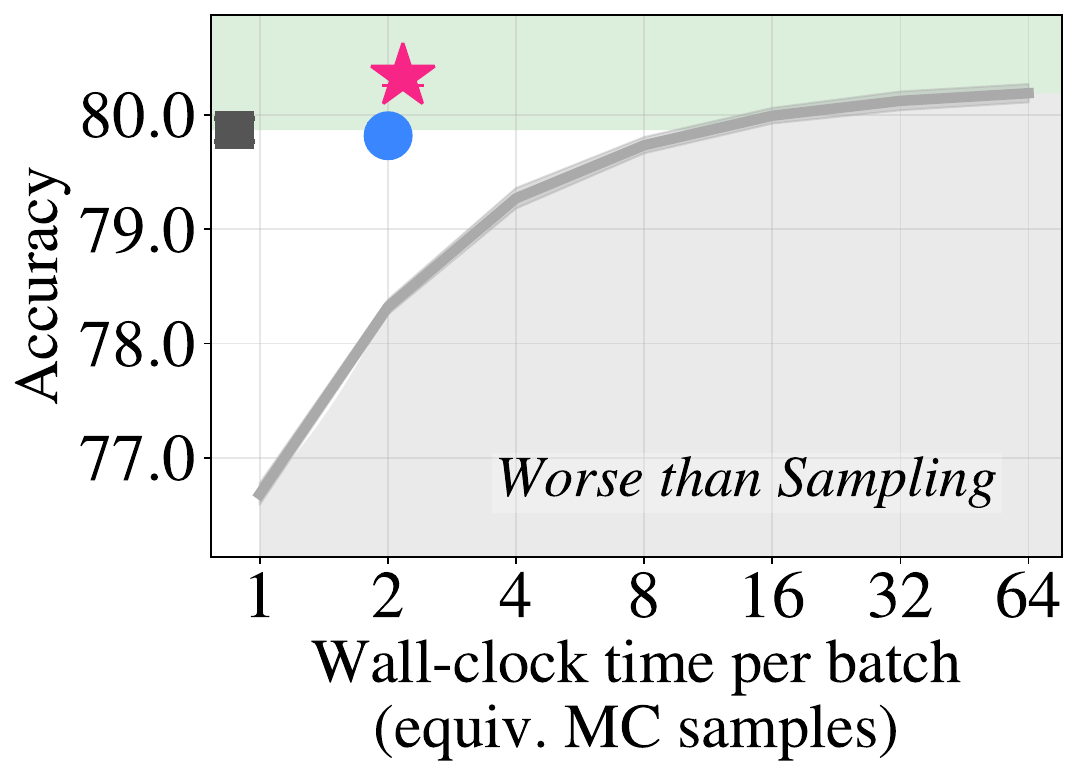}
    \end{subfigure}
    \hfill
    \begin{subfigure}[t]{0.32\textwidth}
        \centering
        \includegraphics[width=\textwidth]{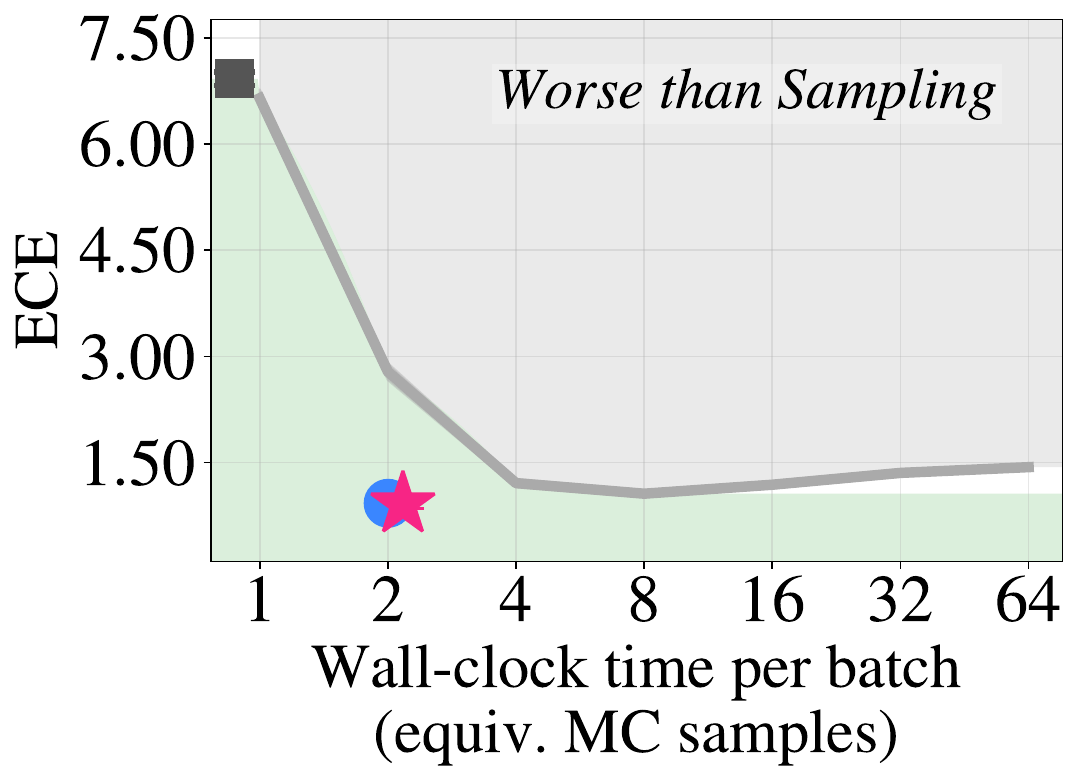}
    \end{subfigure}
    \hfill
    \begin{subfigure}[t]{0.32\textwidth}
        \centering
        \includegraphics[width=\textwidth]{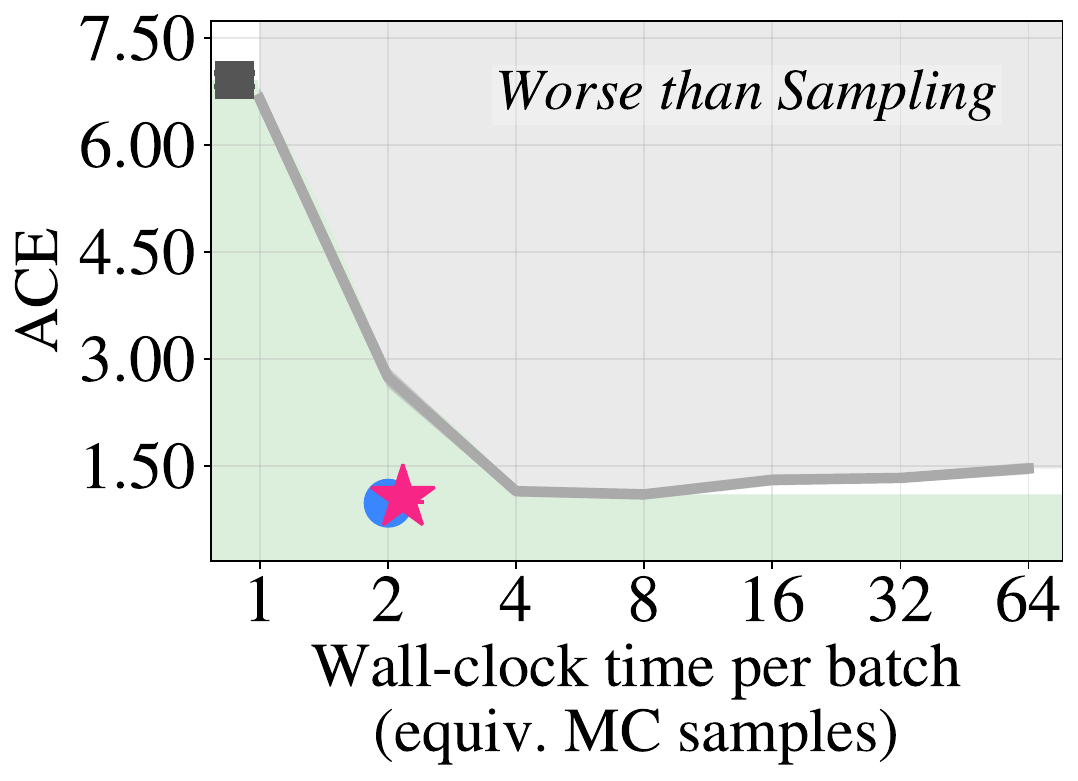}
    \end{subfigure}

    \vspace{1em}  %

    \begin{subfigure}[t]{0.32\textwidth}
        \centering
        \includegraphics[width=\textwidth]{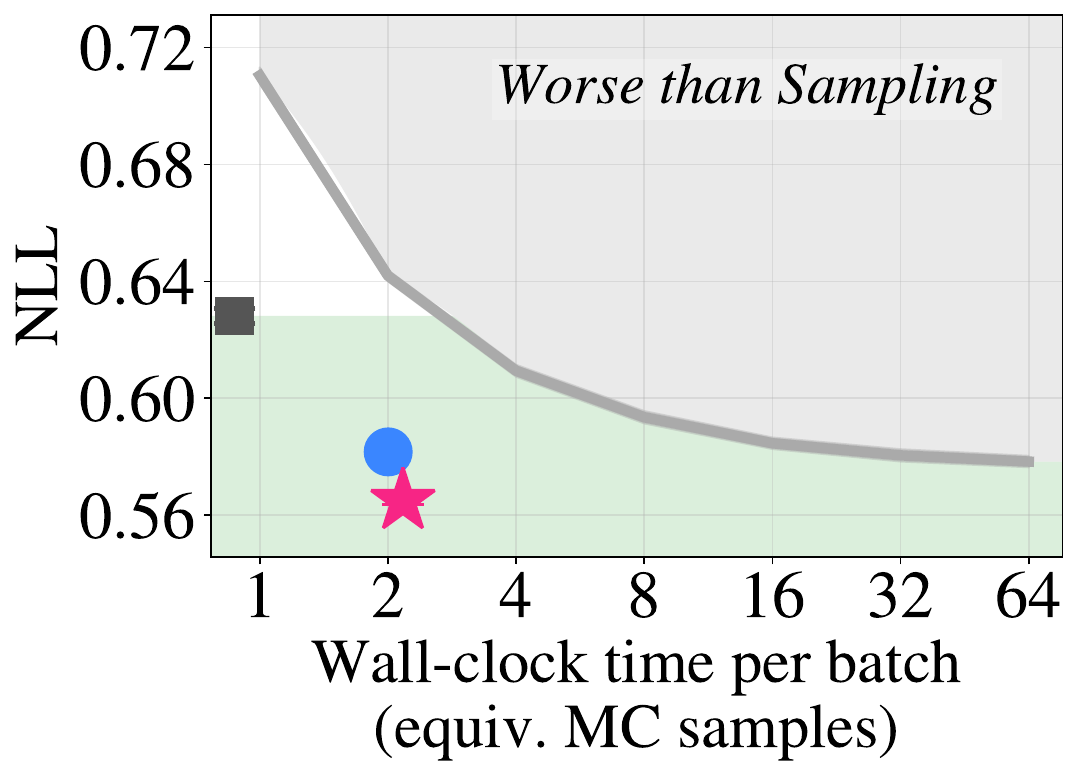}
    \end{subfigure}
    \hfill
    \begin{subfigure}[t]{0.32\textwidth}
        \centering
        \includegraphics[width=\textwidth]{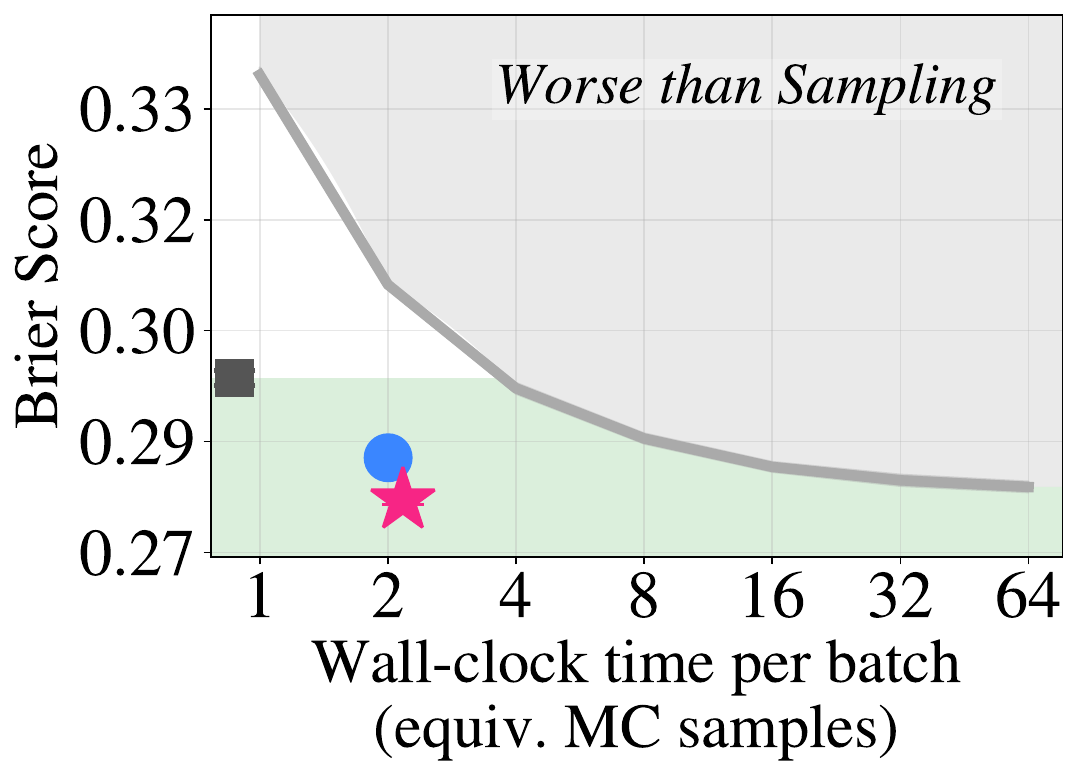}
    \end{subfigure}
    \hfill
    \begin{subfigure}[t]{0.32\textwidth}
        \centering
        \includegraphics[width=\textwidth]{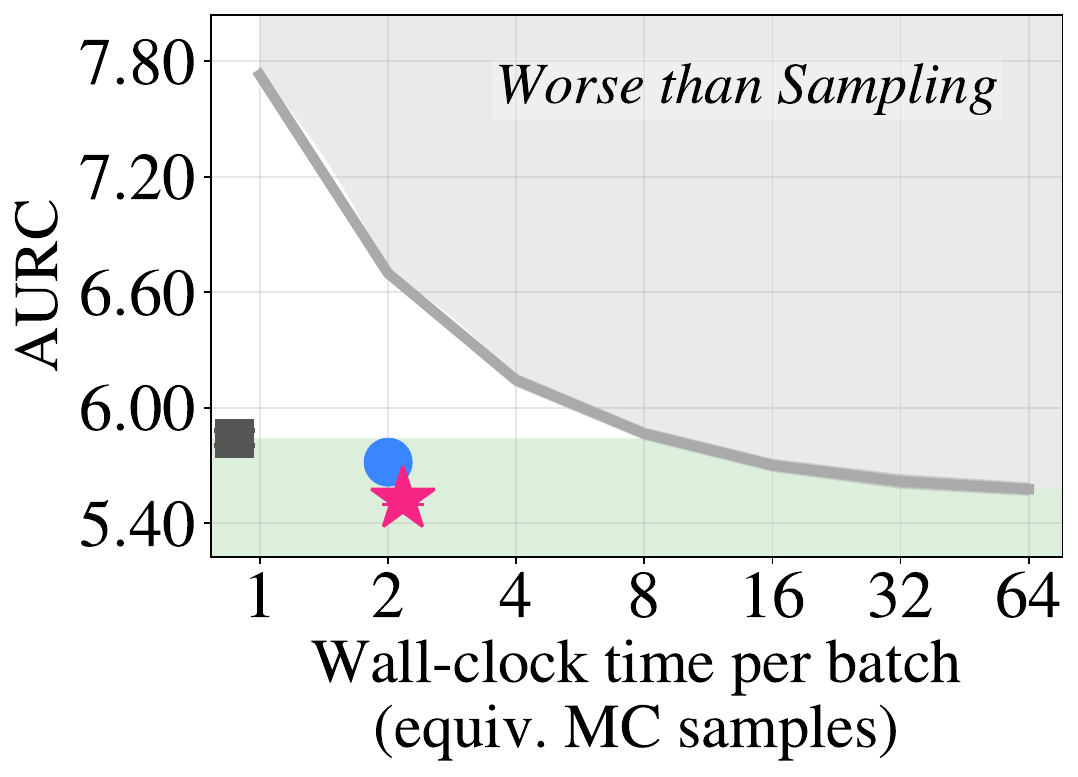}
    \end{subfigure}

    \vspace{1em}

    \begin{subfigure}[t]{0.32\textwidth}
        \centering
        \includegraphics[width=\textwidth]{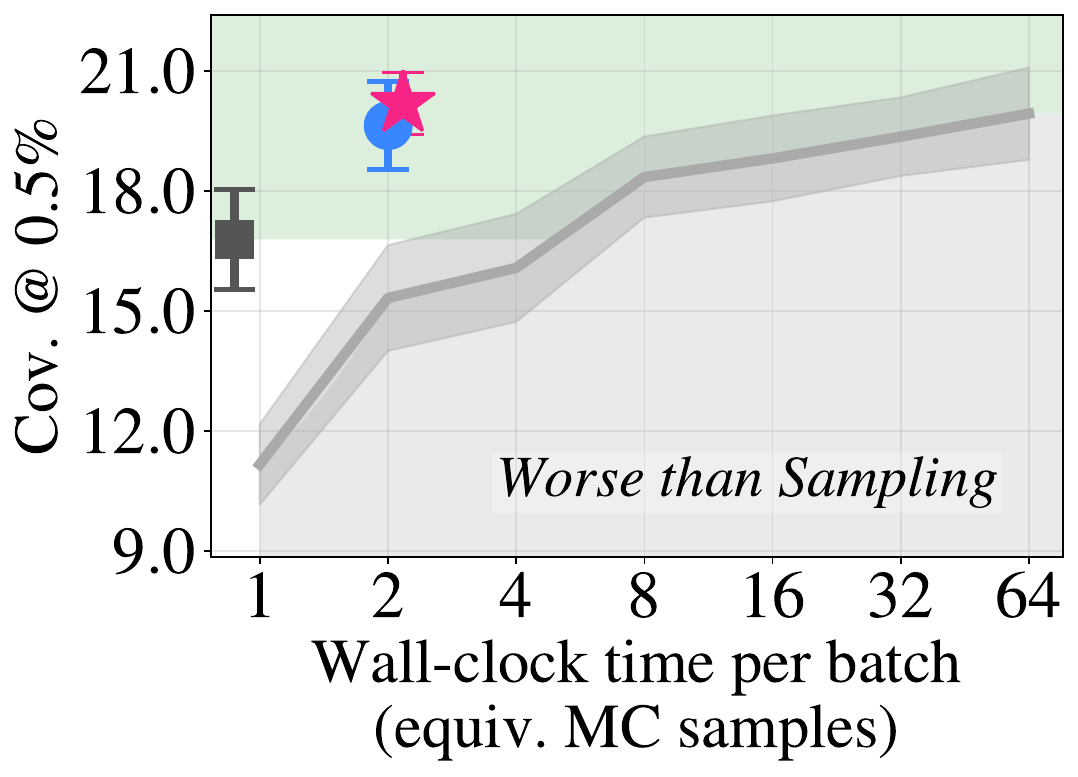}
    \end{subfigure}
    \hfill
    \begin{subfigure}[t]{0.32\textwidth}
        \centering
        \includegraphics[width=\textwidth]{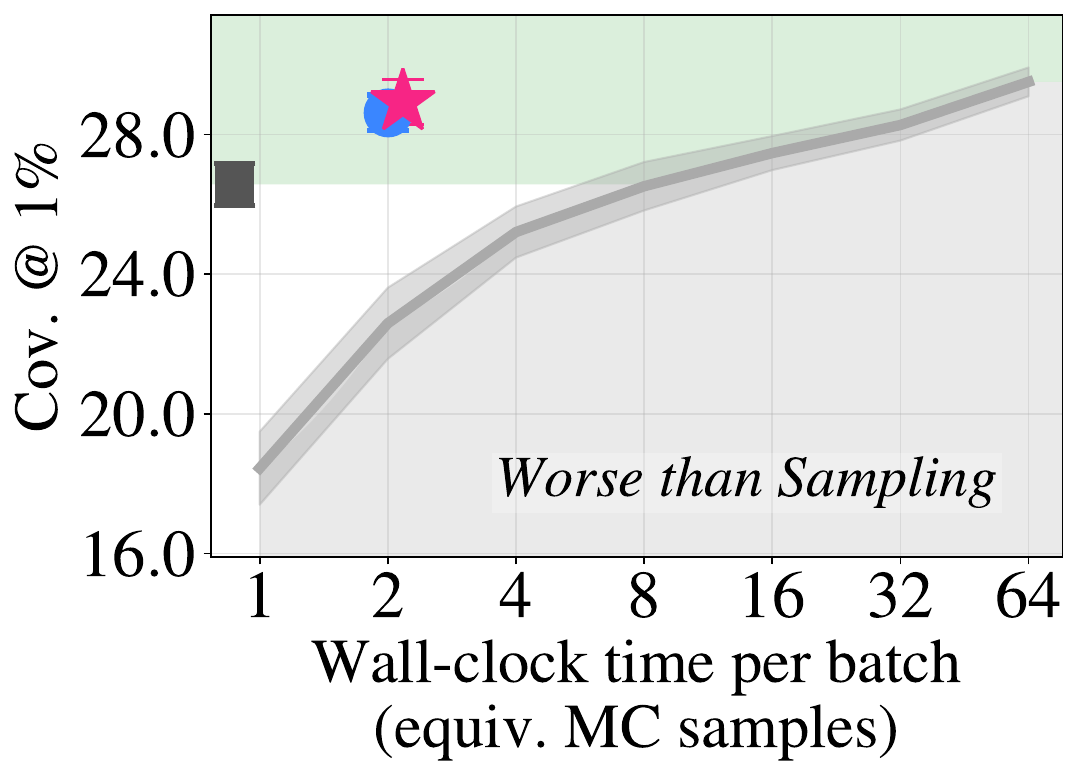}
    \end{subfigure}
    \hfill
    \begin{subfigure}[t]{0.32\textwidth}
        \centering
        \includegraphics[width=\textwidth]{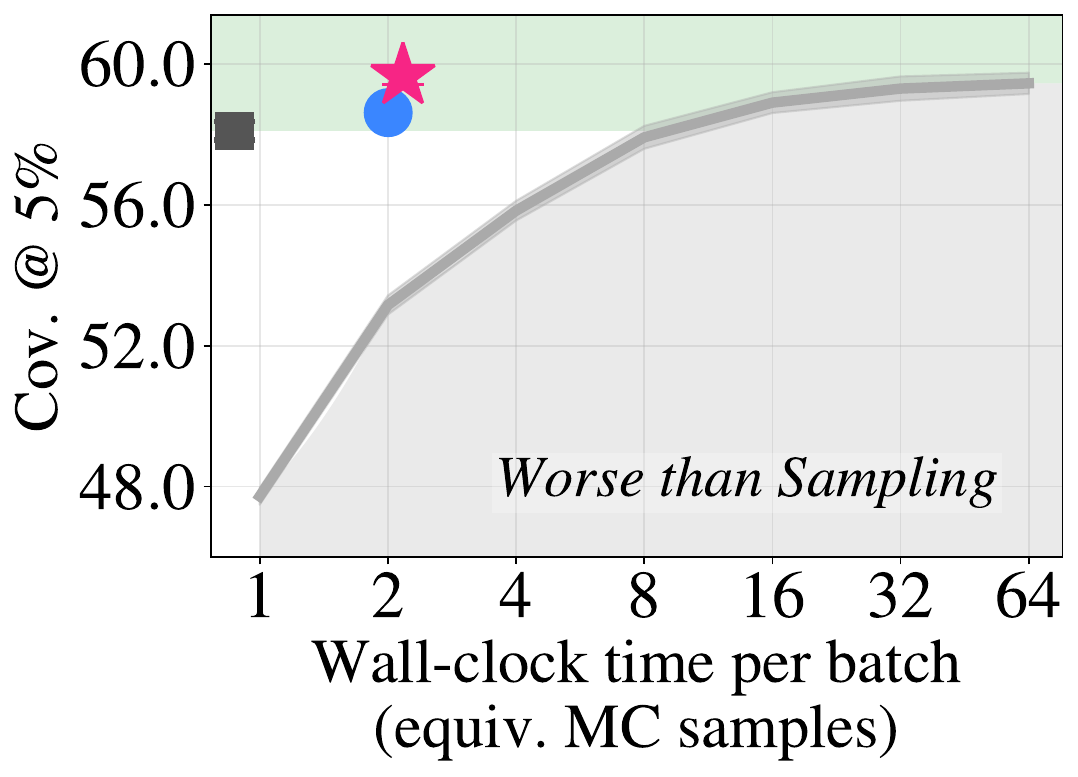}
    \end{subfigure}

    \vspace{1em}

    \begin{subfigure}[t]{0.32\textwidth}
        \centering
        \includegraphics[width=\textwidth]{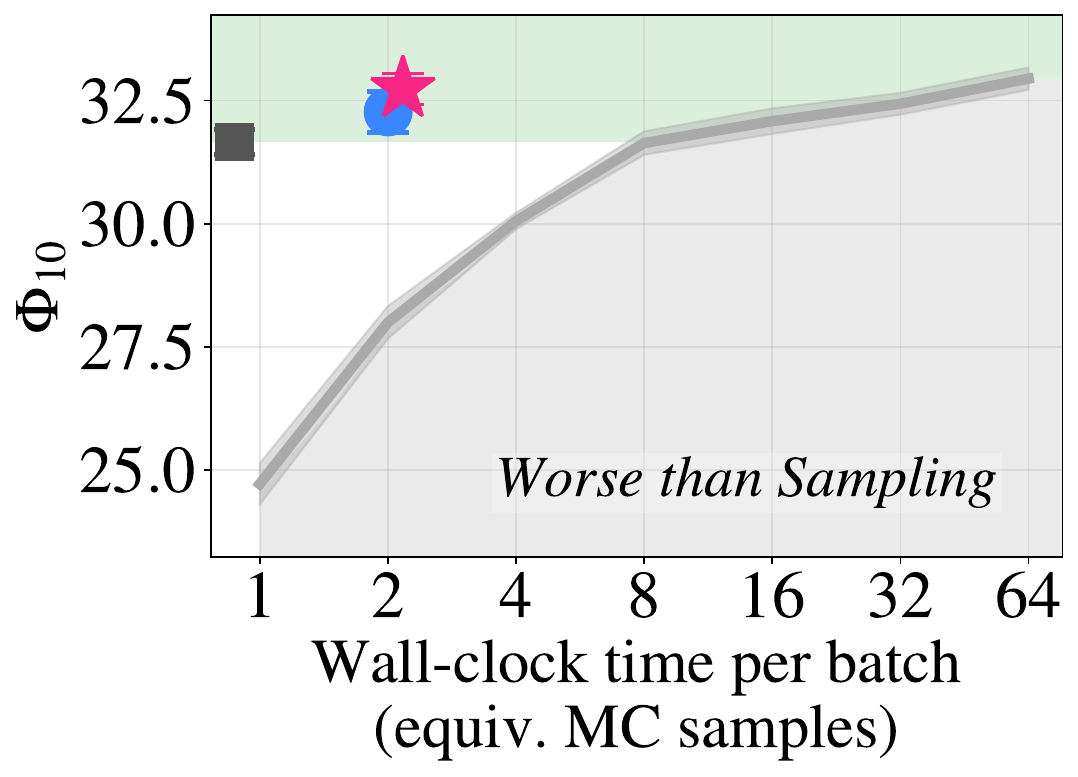}
    \end{subfigure}
    \hfill
    \begin{subfigure}[t]{0.32\textwidth}
        \centering
        \includegraphics[width=\textwidth]{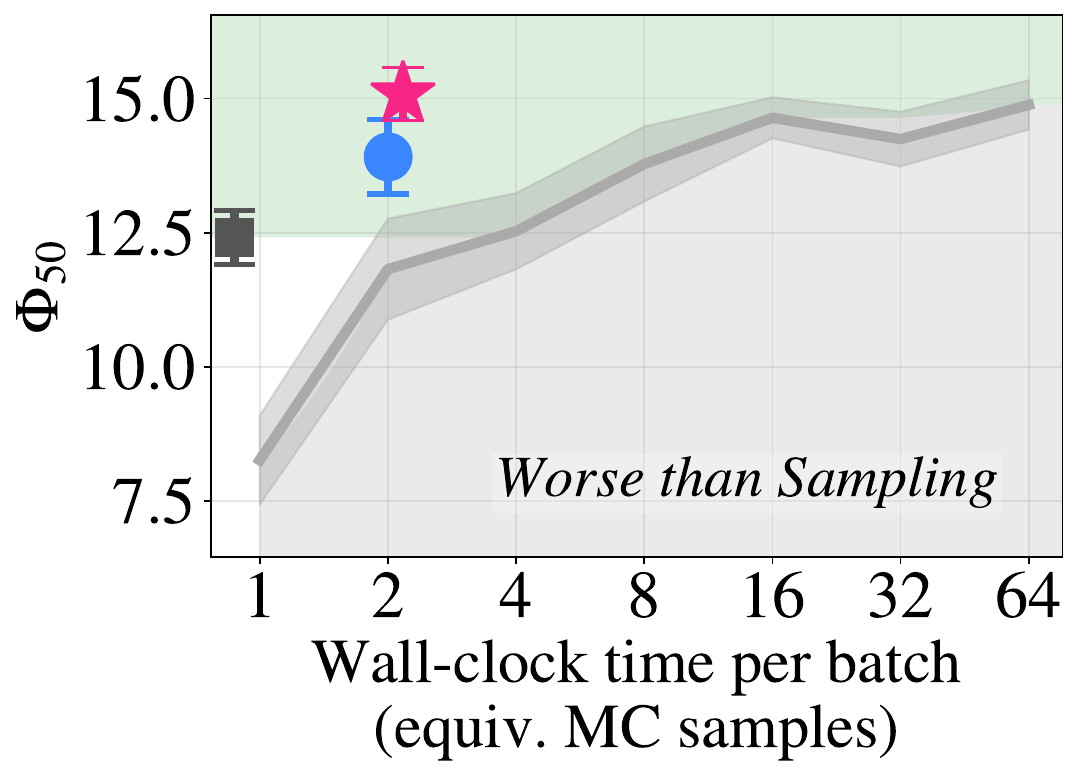}
    \end{subfigure}
    \hfill
    \begin{subfigure}[t]{0.32\textwidth}
        \centering
        \includegraphics[width=\textwidth]{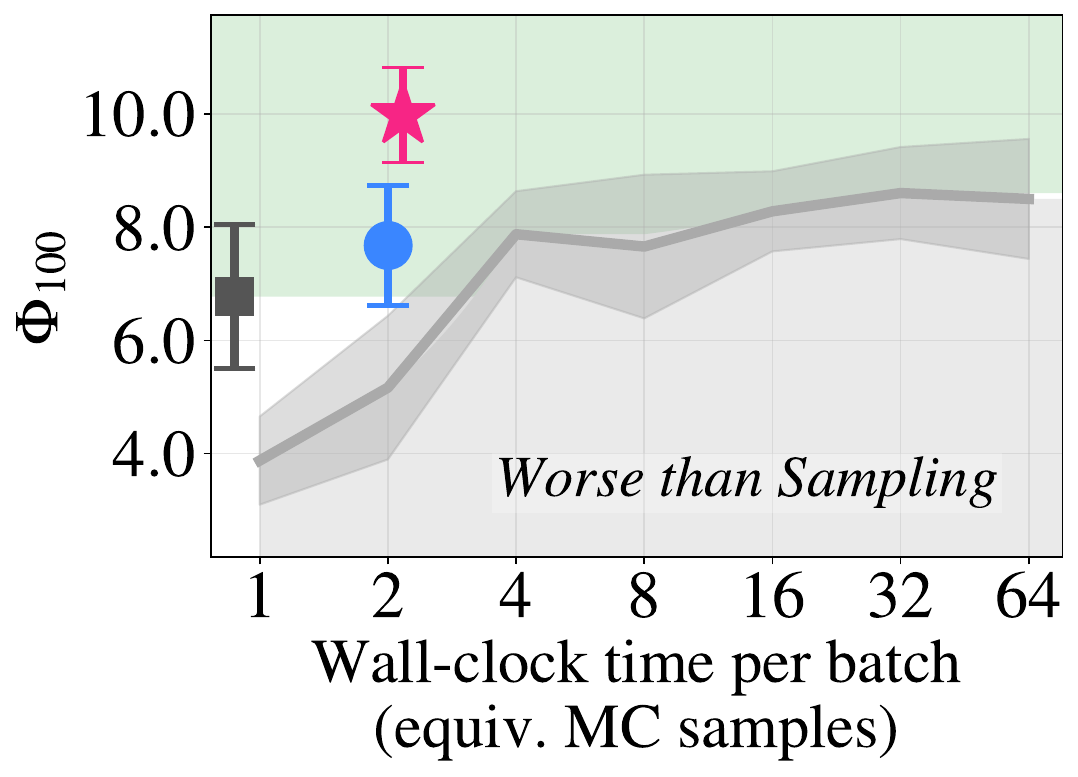}
    \end{subfigure}

    \vspace{0.3em}
    \includegraphics[width=0.6\textwidth]{new_figures/metrics/legend_horizontal.pdf}
    \caption{Results on Accuracy, Calibration, Loss, the Brier Score and Selective Prediction for the "sub-tiny" ViT on CIFAR-10. The \colorbox{Paretogreen!20}{Pareto-dominating region} (vs. the mean network and MC Sampling) is highlighted in green. SEM (Standard Error of the Mean) across 10 random seeds is shown.}
    \label{fig:full_cifar10}
\end{figure}

We show additional figures in the style of \cref{fig:teaser} on Accuracy, Calibration (ECE \cite{ece}, ACE \cite{ace}), the Brier Score \cite{brier1950} and Selective Prediction (Coverage at Risk, and Effective Reliability \cite{reliable_vqa}) for our results on image classification, VQA and Visual Reasoning from the main paper. With very few exceptions (such as ECE and ACE on VQAv2), our CVP method Pareto-dominates the mean network and MC sampling on the efficiency-reliability tradeoff, while Streamlining often does not improve upon the results of the mean network - particularly in accuracy and selective prediction. \Cref{fig:full_beitvqa} shows the plots for BEiT-3 on VQAv2, \cref{fig:full_nlvr} for BEiT-3 on NLVR2, \cref{fig:full_viltvqa} for ViLT on VQAv2, \cref{fig:full_cifar100} for ViT-Base on CIFAR-100 and \cref{fig:full_cifar10} for the sub-tiny ViT on CIFAR-10.

\clearpage

\section{Experimental Details}\label{sec:app_hyperparams}

This section reports training and calibration hyperparameters (\cref{tab:hyperparams:ivon}), hardware, and licenses of the existing assets used in our experiments.

\myparagraph{IVON training} Hyperparameters were optimized for selective prediction, using AURC on the validation set as the objective and MC sampling with $M=8$ for validation predictions. The full set of IVON hyperparameters is reported in \cref{tab:hyperparams:ivon}. For the hyperparameters of the ResNet-50 checkpoints, see \cite{ivon}.

\begin{table}[ht]
  \centering
  \renewcommand{\arraystretch}{1.3}
  \setlength{\tabcolsep}{3.5pt}
  \footnotesize
  \caption{IVON training hyperparameters and model sizes. $\eta$: learning rate; $\lambda$: effective sample size; $h_0$: Hessian initialization; $\delta$: weight decay; $\beta_1$: gradient momentum; $\beta_2$: Hessian momentum; $r$: per-parameter update clipping radius (IVON); Grad clip: gradient norm clipping applied before the optimizer. $\beta_1 = 0.9$ and $m = 1$ MC sample per gradient step are identical across all experiments.}
  \label{tab:hyperparams:ivon}
  \begin{tabular}{ll c c c c c c c c c}
    \toprule
    \multirow{2}{*}{Dataset} & \multirow{2}{*}{Model} & \multirow{2}{*}{\#Params} & \multirow{2}{*}{$\eta$} & \multirow{2}{*}{$\lambda$} & \multirow{2}{*}{$h_0$} & \multirow{2}{*}{$\beta_2$} & \multirow{2}{*}{$\delta$} & \multirow{2}{*}{$r$} & \multirow{2}{*}{Grad clip} & \multirow{2}{*}{Epochs} \\
     & & & & & & & & & & \\
    \midrule
    CIFAR-10 & Custom ViT & 1.8M & $2$ & $3.5\cdot10^{6}$ & $2\cdot10^{-2}$ & $0.9999$ & $1\cdot10^{-4}$ & $1\cdot10^{-3}$ & $1$ & $100$ \\
    CIFAR-100 & ViT-B/16 & 85.9M & $1$ & $5\cdot10^{6}$ & $2\cdot10^{-3}$ & $0.9998$ & $1\cdot10^{-4}$ & $1\cdot10^{-3}$ & $1$ & $50$ \\
    VQAv2 & BEiT-3 & 228.1M & $0.02$ & $5\cdot10^{5}$ & $5\cdot10^{-1}$ & $0.99995$ & $5\cdot10^{-5}$ & $1\cdot10^{-3}$ & $25$ & $10$ \\
    NLVR2 & BEiT-3 & 225.7M & $100$ & $1\cdot10^{8}$ & $5\cdot10^{-4}$ & $0.9999$ & $1\cdot10^{-5}$ & $1\cdot10^{-3}$ & $25$ & $10$ \\
    VQAv2 & ViLT & 117.6M & $0.5$ & $5\cdot10^{5}$ & $2\cdot10^{-2}$ & $0.99995$ & $5\cdot10^{-5}$ & $1\cdot10^{-3}$ & --- & $50$ \\
    \bottomrule
  \end{tabular}
\end{table}

\myparagraph{Calibration} The per-layer scaling parameters $\alpha_l$ are fitted by minimizing NLL on the validation set, with all other model parameters frozen. We use the Adam optimizer with a learning rate of $\eta = 0.03$, 10 epochs, and an effective batch size of 256 across all experiments.

\myparagraph{Hardware} All experiments were conducted on NVIDIA A100 GPUs with 80\,GB of memory. Since CVP is a test-time method that operates on already-trained checkpoints, training compute reflects standard fine-tuning of the underlying architectures (\cf~\cite{ivon, varvqa} for the IVON-trained checkpoints) rather than additional cost introduced by our method. Inference cost is reported relative to a single MC forward pass throughout \cref{sec:experiments}.

\subsection{Licenses of Existing Assets}\label{sec:app_licenses}

The following lists the existing assets used in this work and their licenses. All assets were used in compliance with their respective terms.

\myparagraph{Datasets}
\begin{itemize}
    \item \textbf{CIFAR-10} and \textbf{CIFAR-100} \cite{krizhevsky2009}: released for research use; no formal license attached. Available at \url{https://www.cs.toronto.edu/~kriz/cifar.html}.
    \item \textbf{VQAv2} \cite{vqav2}: annotations licensed under CC-BY 4.0 by the VQA Consortium; images sourced from MS COCO under the COCO Terms of Use.
    \item \textbf{NLVR2} \cite{nlvr2}: sentence and label annotations licensed under CC-BY 4.0; image URLs are released under Cornell's NLVR2 Terms of Service, which limit use to non-commercial research and educational purposes.
\end{itemize}

\myparagraph{Models and code}
\begin{itemize}
    \item \textbf{Vision Transformer (ViT)} \cite{vit}: code and pre-trained checkpoints released by \texttt{google-research/vision\_transformer} under the Apache 2.0 License.
    \item \textbf{ViLT} \cite{vilt}: code and pre-trained checkpoints released by \texttt{dandelin/ViLT} under the Apache 2.0 License.
    \item \textbf{BEiT-3} \cite{beit3}: code and pre-trained checkpoints released as part of \texttt{microsoft/unilm} under the MIT License.
    \item \textbf{IVON optimizer} \cite{ivon}: PyTorch implementation \texttt{ivon-opt} released by \texttt{team-approx-bayes/ivon} under the GNU General Public License v3 (GPLv3+).
\end{itemize}

\section{Learned Variance Scaling Parameters}\label{sec:app_scalingparams}

\begin{figure}[ht]
    \centering
    \setlength{\tabcolsep}{2pt}

    \hspace{2em}%
    \makebox[0.43\linewidth]{\large CVP}\hfill
    \makebox[0.45\linewidth]{\large Streamlining}
    \vspace{2pt}
    \hfill
    \begin{subfigure}{\linewidth}
        \centering
        \includegraphics[width=0.48\linewidth]{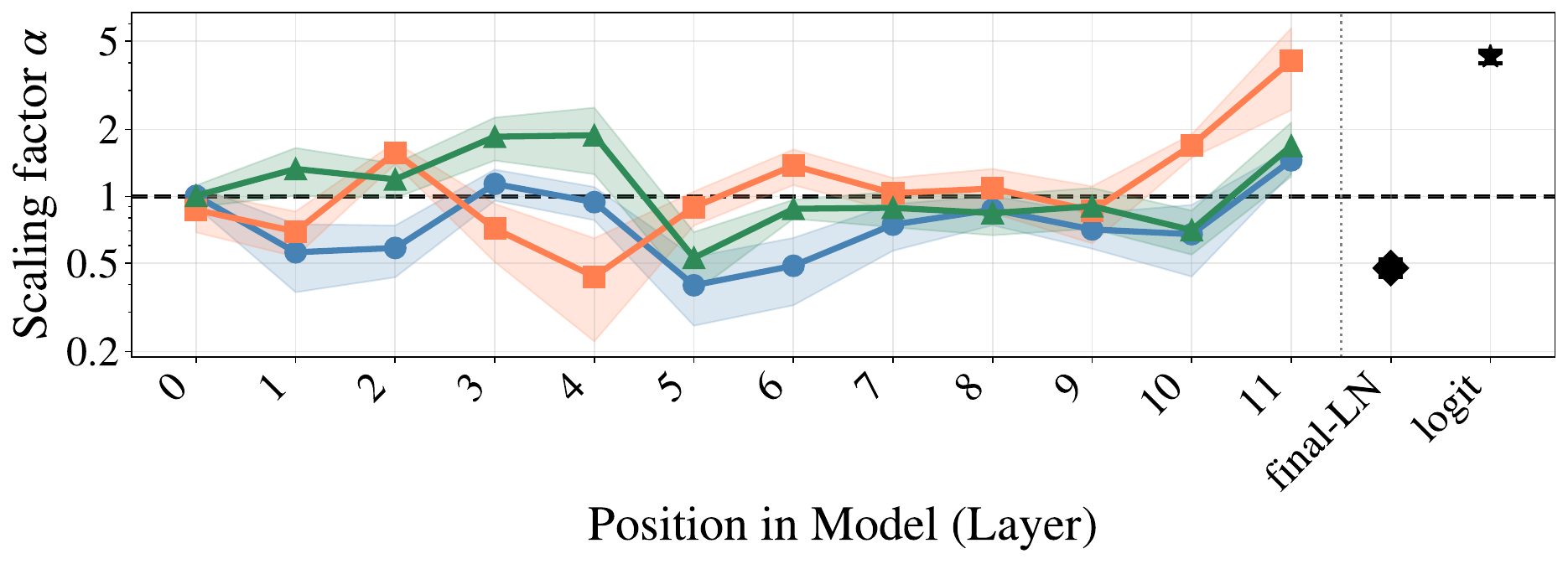}\hfill
        \includegraphics[width=0.48\linewidth]{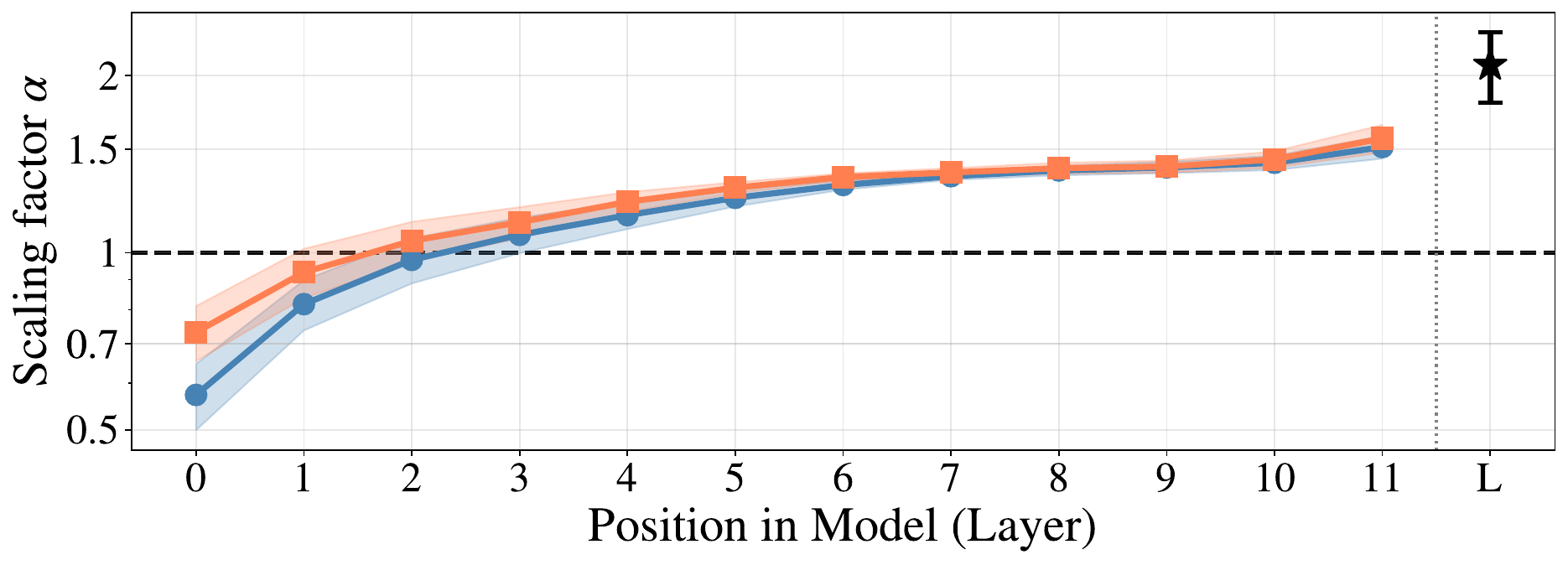}
        \caption{CIFAR-100 (ViT-Base)}
    \end{subfigure}
    \hfill
    \begin{subfigure}{\linewidth}
        \centering
        \includegraphics[width=0.48\linewidth]{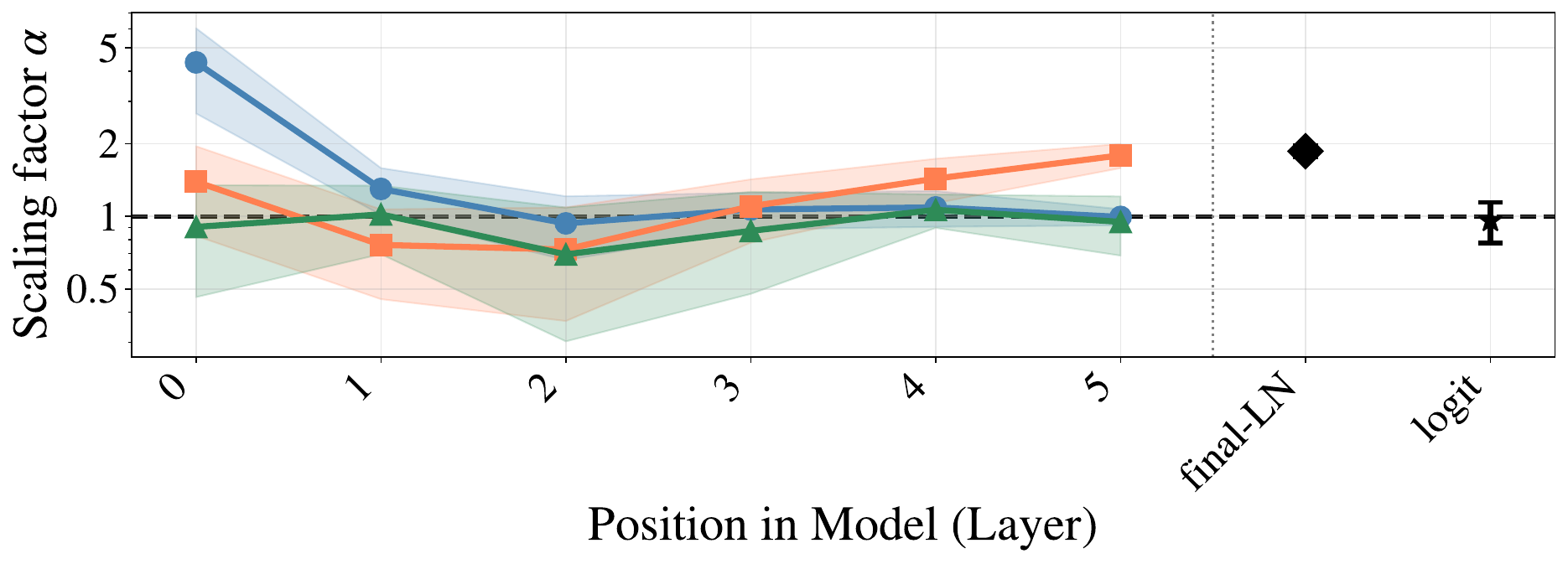}\hfill
        \includegraphics[width=0.48\linewidth]{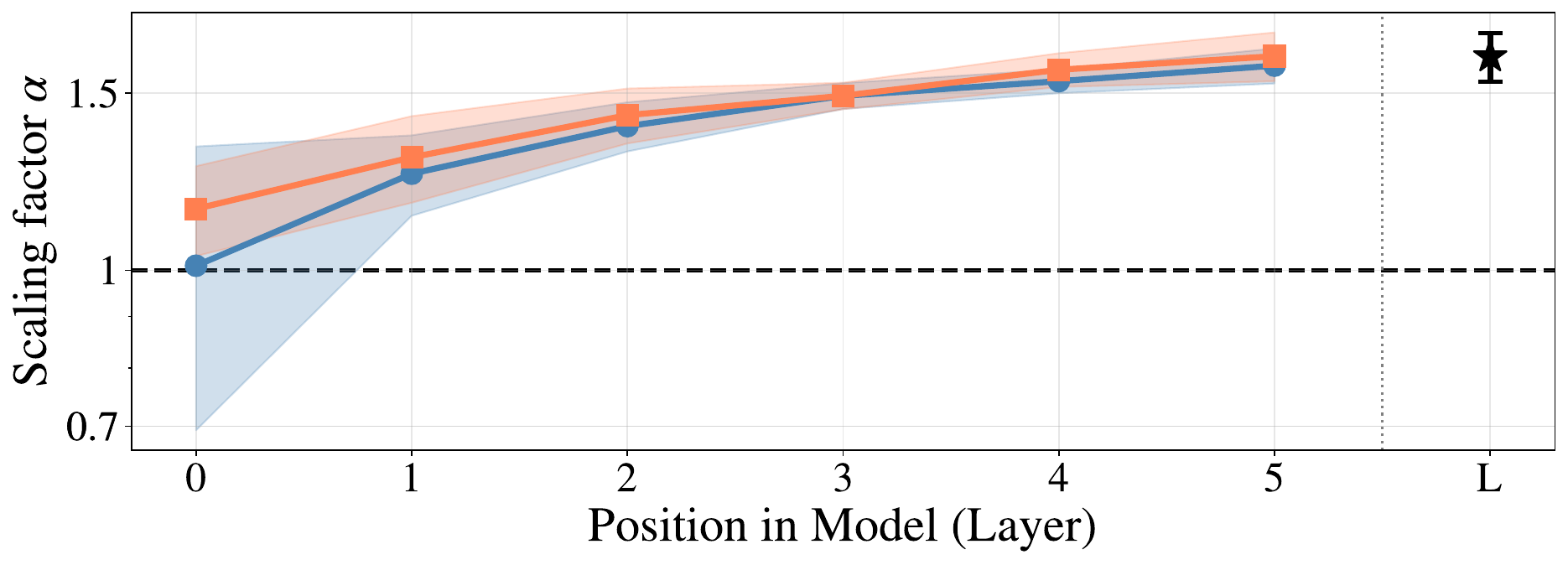}
        \caption{CIFAR-10 (sub-tiny ViT)}
    \end{subfigure}
    \hfill
    \begin{subfigure}{\linewidth}
        \centering
        \includegraphics[width=0.48\linewidth]{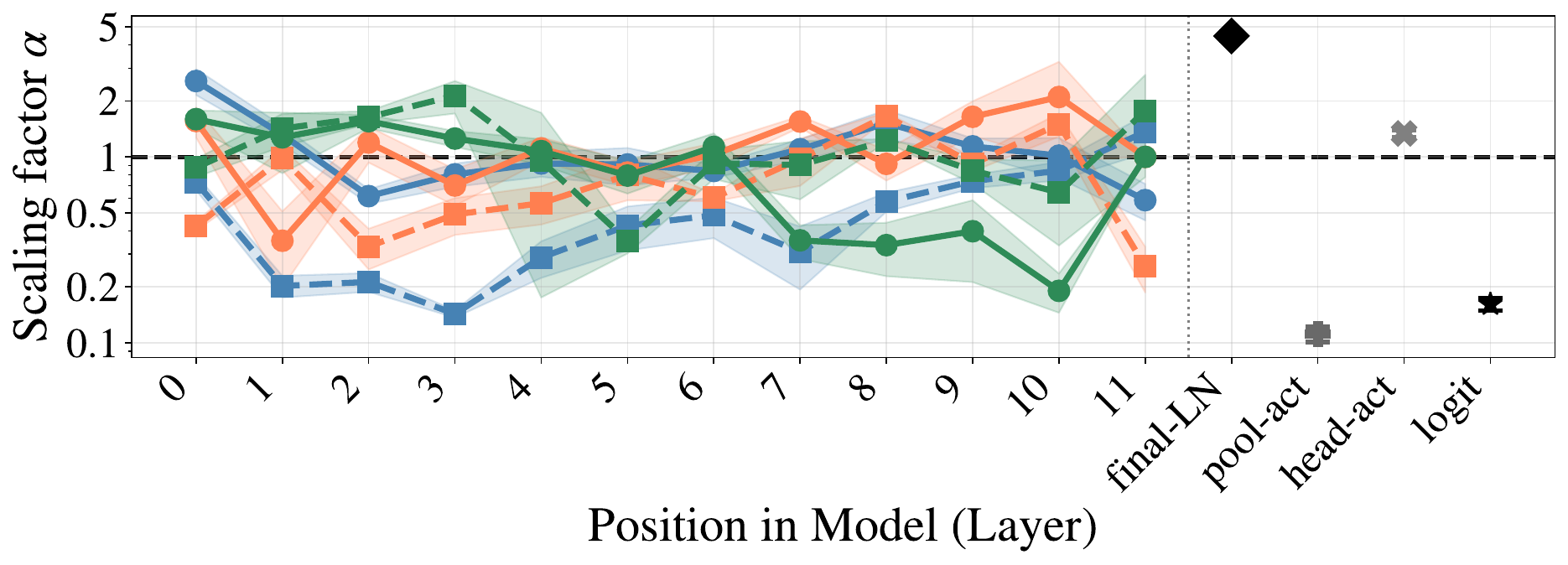}\hfill
        \includegraphics[width=0.48\linewidth]{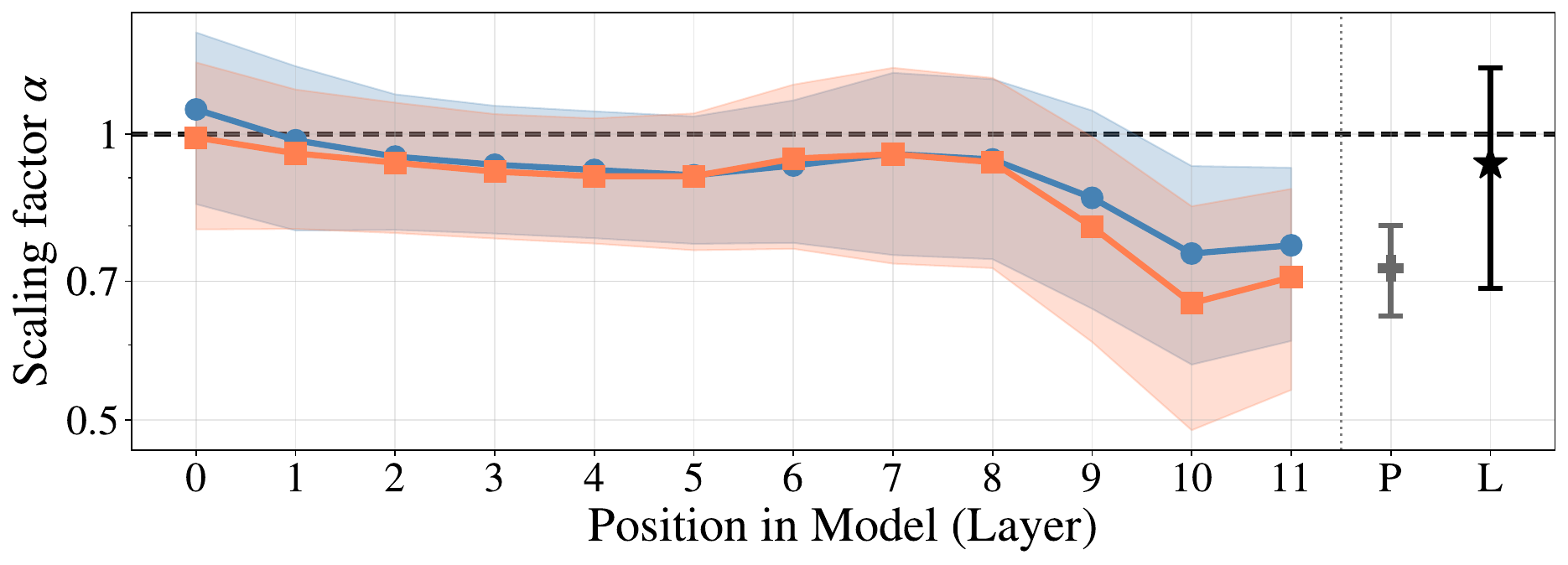}
        \caption{VQAv2 (BEiT-3-base)}
    \end{subfigure}
    \hfill
    \begin{subfigure}{\linewidth}
        \centering
        \includegraphics[width=0.48\linewidth]{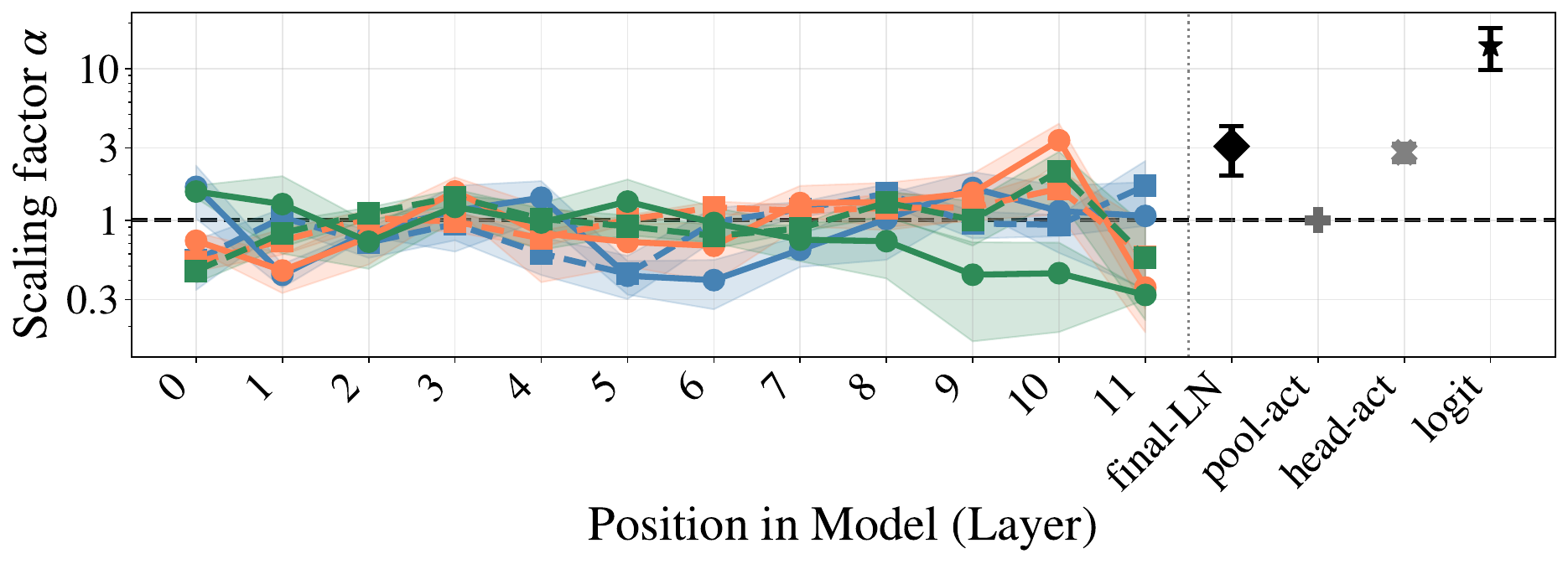}\hfill
        \includegraphics[width=0.48\linewidth]{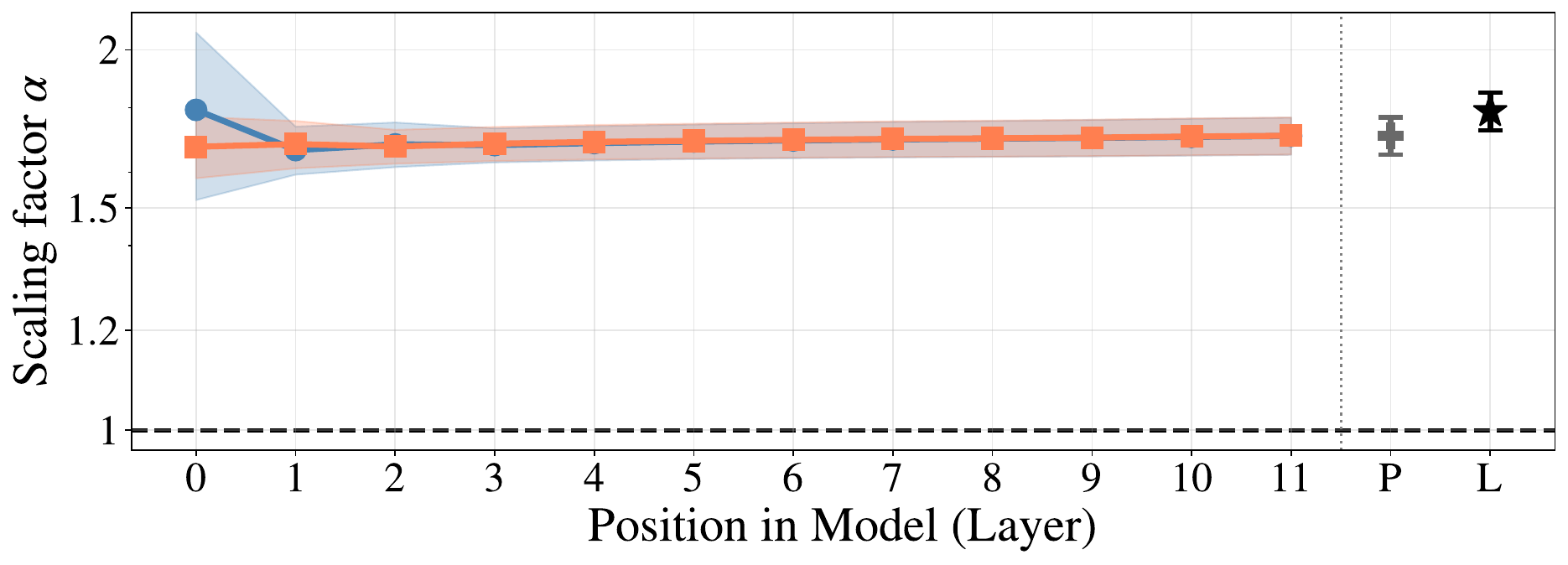}
        \caption{NLVR2 (BEiT-3-base)}
    \end{subfigure}
    \hfill
    \begin{subfigure}{\linewidth}
        \centering
        \includegraphics[width=0.48\linewidth]{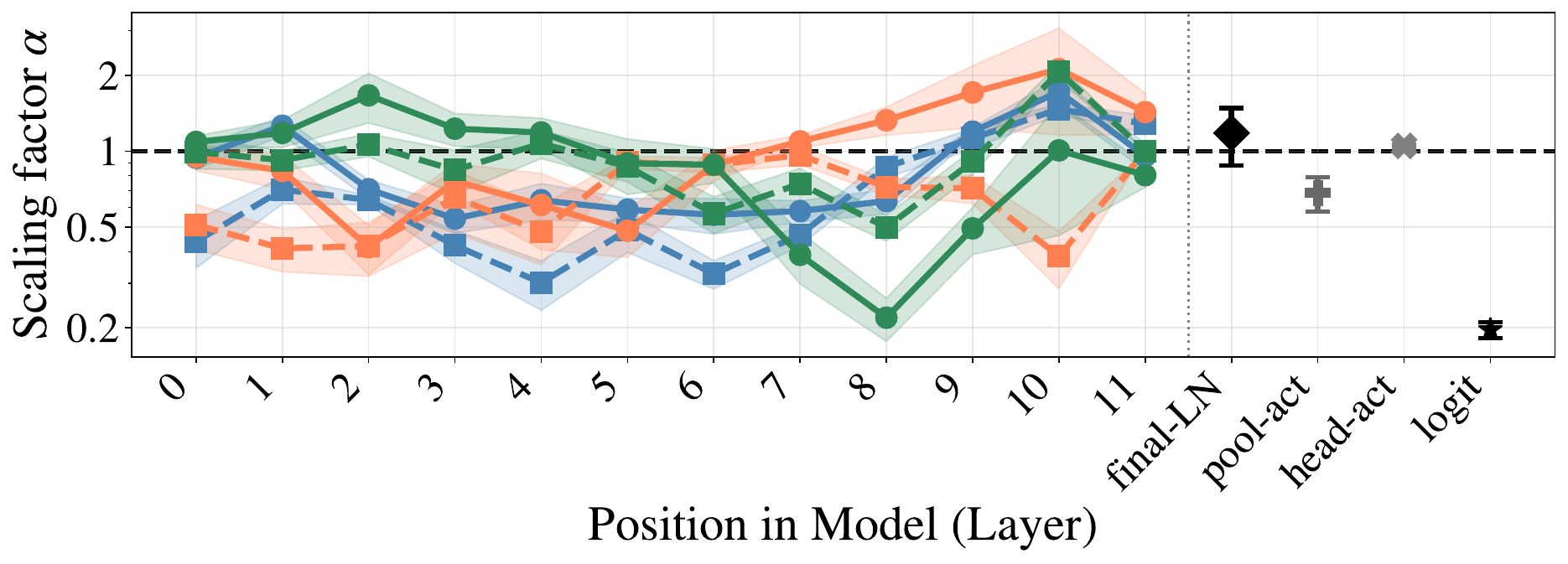}\hfill
        \includegraphics[width=0.48\linewidth]{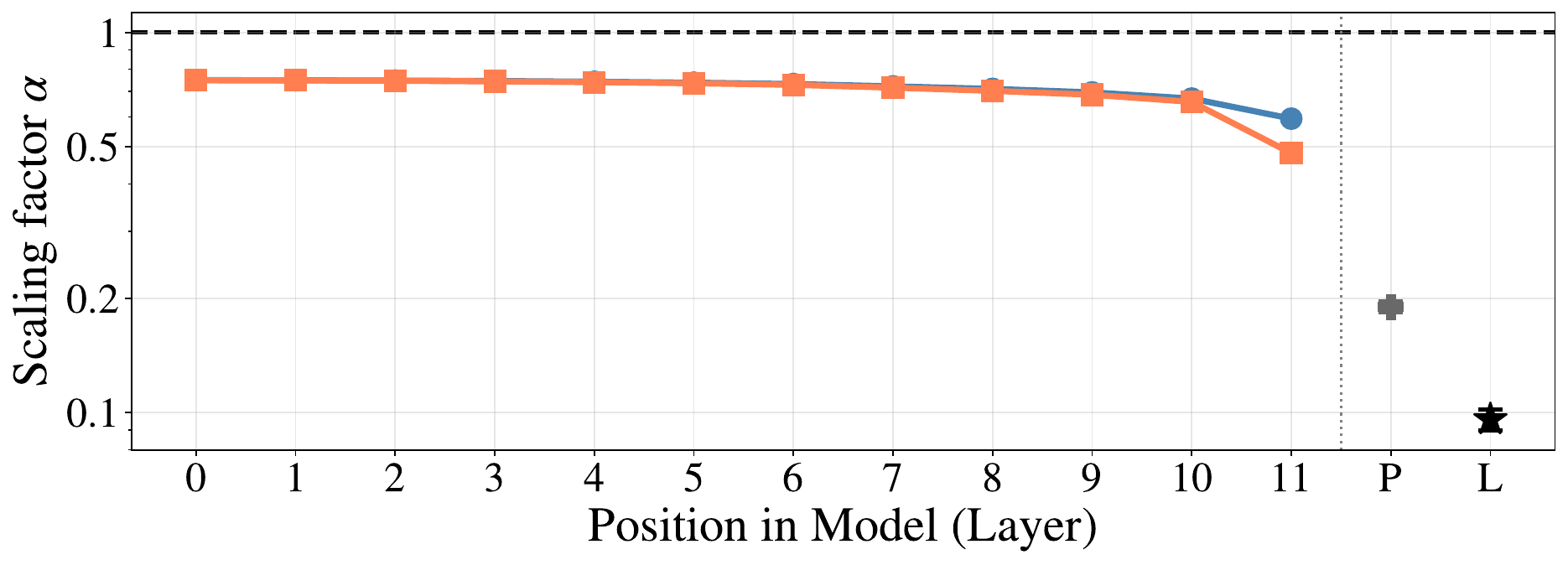}
        \caption{VQAv2 (ViLT)}
    \end{subfigure}
    \hfill
    \begin{subfigure}{\linewidth}
        \centering
        \includegraphics[width=0.48\linewidth]{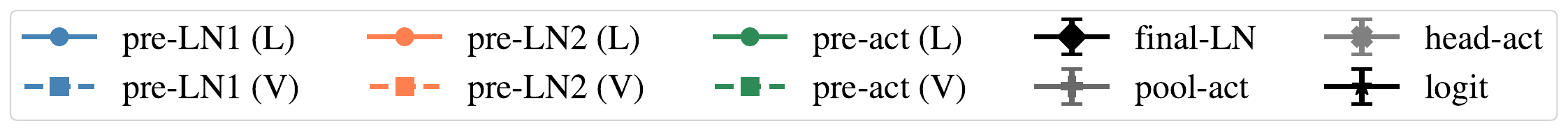}\hfill
        \includegraphics[width=0.2\linewidth]{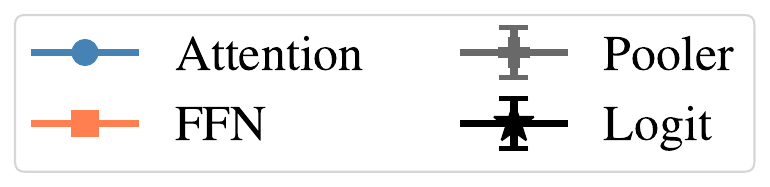}
    \end{subfigure}

    \caption{Learned scaling factors $\alpha$ per transformer block (0--11), pooler (P), and logit (L) for CVP (left) and Streamlining (right) across five benchmarks. The dashed line marks $\alpha = 1$ (no rescaling). Shaded bands show standard deviation across seeds.}
    \label{fig:scaling_factors}
\end{figure}

We report the learned variance scaling parameters for all our trained models in \Cref{fig:scaling_factors}. Compared to Streamlining, the learned factors show more variability between layers, and often increase for the last 1-2 layers (BEiT-3 NLVR2, BEiT-3 VQA, CIFAR-10 sub-tiny ViT), particularly when measured using the geometric mean of the scaling factors within one block $\alpha_l^\textrm{geo}=\sqrt{\alpha_l^\textrm{Attn}\cdot\alpha_l^\textrm{FFN}}$. There also is more difference between the FFN and Attn scaling factors for CVP, while they are almost identical per transformer block for Streamlining. Finally, the logit scaling factors for CVP often taken extreme values, \eg $\alpha_\textrm{Logit}>100$ for BEiT-3 NLVR2 or $\alpha_\textrm{Logit}<0.1$ for BEiT-3 VQA, while those of Streamlining stay much closer to 1.

\section{Extended Metrics Tables}\label{sec:app_resultstables}

We present results on our six measured Image Classification / multimodal setups (\cf \cref{tab:default:vit_test_sem,tab:default:multimodal_test_sem} from the main paper) for additional reliability metrics in \Cref{tab:default_extra:combined_test_sem}. We also report the macro-average across all six Image Classification / multimodal setups in \cref{tab:default:avg_test_sem,tab:default_extra:avg_test_sem}.

\begin{table*}[ht]
  \centering
  \renewcommand{\arraystretch}{1.3}
  \setlength{\tabcolsep}{3.5pt}
  \footnotesize
  \caption{CVP and methods of similar runtime - evaluated on additional reliability metrics. Reported are mean and standard error of the mean (SEM) over 10 seeds (ViT) and 5 seeds (multimodal, ResNet); bold marks the best result per metric. Wall-clock time is relative to a single MC sample.}
  \label{tab:default_extra:combined_test_sem}
  \begin{tabular}{l c r r r r r r}
    \toprule
    \multirow{2}{*}{Method} & Rel. & \multicolumn{1}{c}{\multirow{2}{*}{C@2\%}} & \multicolumn{1}{c}{\multirow{2}{*}{C@5\%}} & \multicolumn{1}{c}{\multirow{2}{*}{ACE ($\downarrow$)}} & \multicolumn{1}{c}{\multirow{2}{*}{$\Phi_{10}$}} & \multicolumn{1}{c}{\multirow{2}{*}{$\Phi_{50}$}} & \multicolumn{1}{c}{\multirow{2}{*}{$\Phi_{100}$}} \\
     & Time &  &  &  &  &  &  \\
    \midrule
    \multicolumn{8}{l}{\textit{\textbf{ImageNet (ResNet-50)}}} \\
    Mean Network & 1.0$\times$ & \vpm{38.7}{0.1} & \vpm{58.4}{0.1} & \vpm{3.47}{0.04} & \vpm{31.4}{0.1} & \bvpm{11.2}{0.2} & \vpm{3.3}{0.4} \\
    \quad + Temp. Sc. & 1.0$\times$ & \vpm{38.5}{0.1} & \vpm{58.3}{0.1} & \vpm{1.31}{0.02} & \vpm{31.3}{0.1} & \vpm{10.9}{0.2} & \vpm{3.3}{0.5} \\
    \hdashline
    2 samples & 2.0$\times$ & \vpm{36.1}{0.2} & \vpm{56.0}{0.1} & \bvpm{1.15}{0.05} & \vpm{30.1}{0.1} & \vpm{10.0}{0.2} & \vpm{3.8}{0.4} \\
    4 samples & 4.0$\times$ & \vpm{37.2}{0.1} & \vpm{56.8}{0.1} & \vpm{1.60}{0.03} & \vpm{30.4}{0.1} & \vpm{10.8}{0.1} & \bvpm{4.3}{0.4} \\
    \hdashline
    \textcolor{Streamlining}{Streamlining} & 2.6$\times$ & \vpm{38.5}{0.1} & \vpm{58.2}{0.1} & \vpm{1.41}{0.02} & \vpm{31.3}{0.1} & \vpm{11.0}{0.2} & \vpm{3.1}{0.5} \\
    \hdashline
    \textcolor{CVP}{CVP} & 2.9$\times$ & \bvpm{38.8}{0.1} & \bvpm{58.6}{0.1} & \vpm{1.40}{0.06} & \bvpm{31.8}{0.1} & \vpm{10.7}{0.3} & \vpm{4.0}{0.3} \\
    \midrule
    \multicolumn{8}{l}{\textit{\textbf{CIFAR-100 (ViT-Base)}}} \\
    Mean Network & 0.9$\times$ & \vpm{66.1}{0.3} & \vpm{82.0}{0.2} & \vpm{4.42}{0.07} & \vpm{51.2}{0.2} & \vpm{29.5}{0.8} & \vpm{19.1}{1.1} \\
    \quad + Temp. Sc. & 0.9$\times$ & \vpm{65.3}{0.3} & \vpm{81.8}{0.2} & \vpm{2.70}{0.06} & \vpm{50.9}{0.2} & \vpm{28.4}{0.7} & \vpm{18.0}{1.0} \\
    \hdashline
    2 samples & 2.0$\times$ & \vpm{65.8}{0.4} & \vpm{80.7}{0.2} & \vpm{2.54}{0.09} & \vpm{51.3}{0.4} & \vpm{30.9}{0.9} & \vpm{22.7}{0.9} \\
    4 samples & 4.0$\times$ & \vpm{66.7}{0.5} & \vpm{82.0}{0.2} & \bvpm{1.82}{0.03} & \vpm{52.1}{0.4} & \vpm{33.0}{0.7} & \vpm{23.7}{0.6} \\
    \hdashline
    \textcolor{Streamlining}{Streamlining} & 2.3$\times$ & \vpm{65.1}{0.3} & \vpm{81.7}{0.2} & \vpm{2.86}{0.07} & \vpm{50.8}{0.2} & \vpm{27.9}{0.7} & \vpm{17.8}{0.9} \\
    \hdashline
    \textcolor{CVP}{CVP} & 2.6$\times$ & \bvpm{68.6}{0.4} & \bvpm{83.8}{0.2} & \vpm{1.84}{0.09} & \bvpm{53.0}{0.4} & \bvpm{33.2}{0.8} & \bvpm{24.0}{1.0} \\
    \midrule
    \multicolumn{8}{l}{\textit{\textbf{CIFAR-10 (sub-tiny ViT)}}} \\
    Mean Network & 0.9$\times$ & \vpm{39.2}{0.5} & \vpm{58.1}{0.3} & \vpm{6.91}{0.10} & \vpm{31.7}{0.3} & \vpm{12.4}{0.5} & \vpm{6.8}{1.3} \\
    \quad + Temp. Sc. & 0.9$\times$ & \vpm{39.6}{0.6} & \vpm{58.2}{0.3} & \vpm{1.18}{0.06} & \vpm{31.7}{0.3} & \vpm{13.5}{0.7} & \vpm{6.4}{1.2} \\
    \hdashline
    2 samples & 2.0$\times$ & \vpm{34.2}{0.8} & \vpm{53.2}{0.3} & \vpm{2.75}{0.13} & \vpm{28.0}{0.3} & \vpm{11.8}{0.9} & \vpm{5.2}{1.3} \\
    4 samples & 4.0$\times$ & \vpm{37.4}{0.4} & \vpm{55.8}{0.3} & \vpm{1.15}{0.06} & \vpm{30.1}{0.2} & \vpm{12.5}{0.7} & \vpm{7.9}{0.8} \\
    \hdashline
    \textcolor{Streamlining}{Streamlining} & 2.0$\times$ & \vpm{40.4}{0.4} & \vpm{58.6}{0.3} & \vpm{1.13}{0.09} & \vpm{32.2}{0.3} & \vpm{14.6}{0.7} & \vpm{8.7}{1.1} \\
    \hdashline
    \textcolor{CVP}{CVP} & 2.2$\times$ & \bvpm{41.6}{0.2} & \bvpm{59.7}{0.2} & \bvpm{1.06}{0.06} & \bvpm{32.7}{0.3} & \bvpm{15.1}{0.5} & \bvpm{10.0}{0.8} \\
    \midrule
    \multicolumn{8}{l}{\textit{\textbf{VQAv2 (BEiT-3-base)}}} \\
    Mean Network & 1.0$\times$ & \vpm{32.1}{0.2} & \vpm{49.4}{0.1} & \vpm{3.82}{0.09} & \vpm{34.6}{0.1} & \vpm{16.9}{0.1} & \vpm{11.2}{0.4} \\
    \quad + Temp. Sc. & 1.0$\times$ & \vpm{32.1}{0.2} & \vpm{49.4}{0.1} & \vpm{3.72}{0.09} & \vpm{34.6}{0.1} & \vpm{16.9}{0.1} & \vpm{11.2}{0.4} \\
    \hdashline
    2 samples & 2.0$\times$ & \vpm{31.2}{0.2} & \vpm{47.9}{0.1} & \vpm{2.51}{0.07} & \vpm{33.4}{0.1} & \vpm{16.0}{0.2} & \vpm{9.3}{0.6} \\
    4 samples & 4.0$\times$ & \vpm{32.4}{0.1} & \vpm{48.7}{0.1} & \bvpm{2.15}{0.05} & \vpm{34.1}{0.1} & \vpm{17.0}{0.2} & \vpm{10.7}{0.6} \\
    \hdashline
    \textcolor{Streamlining}{Streamlining} & 2.0$\times$ & \vpm{32.1}{0.2} & \vpm{49.4}{0.1} & \vpm{3.77}{0.09} & \vpm{34.5}{0.1} & \vpm{16.9}{0.1} & \vpm{11.2}{0.4} \\
    \hdashline
    \textcolor{CVP}{CVP} & 2.3$\times$ & \bvpm{33.5}{0.2} & \bvpm{49.7}{0.1} & \vpm{3.12}{0.07} & \bvpm{34.9}{0.1} & \bvpm{18.2}{0.3} & \bvpm{11.7}{0.6} \\
    \midrule
    \multicolumn{8}{l}{\textit{\textbf{NLVR2 (BEiT-3-base)}}} \\
    Mean Network & 0.9$\times$ & \vpm{31.1}{1.4} & \vpm{57.1}{1.1} & \vpm{5.84}{0.67} & \vpm{28.2}{0.7} & \vpm{7.3}{1.2} & \vpm{3.2}{1.3} \\
    \quad + Temp. Sc. & 0.9$\times$ & \vpm{31.2}{1.4} & \vpm{57.1}{1.1} & \bvpm{1.19}{0.12} & \vpm{28.2}{0.7} & \vpm{7.3}{1.2} & \vpm{3.2}{1.3} \\
    \hdashline
    2 samples & 2.0$\times$ & \vpm{32.7}{1.5} & \vpm{54.5}{1.3} & \vpm{4.47}{0.66} & \vpm{27.6}{0.9} & \vpm{8.3}{1.2} & \vpm{4.0}{1.3} \\
    4 samples & 4.0$\times$ & \vpm{34.7}{1.2} & \vpm{56.5}{1.2} & \vpm{3.86}{0.57} & \vpm{29.0}{0.7} & \vpm{10.1}{1.0} & \vpm{4.4}{1.6} \\
    \hdashline
    \textcolor{Streamlining}{Streamlining} & 2.1$\times$ & \vpm{30.9}{1.2} & \vpm{56.6}{1.4} & \vpm{1.50}{0.12} & \vpm{28.4}{1.0} & \vpm{6.7}{1.3} & \vpm{2.7}{1.9} \\
    \hdashline
    \textcolor{CVP}{CVP} & 2.5$\times$ & \bvpm{36.9}{1.0} & \bvpm{58.4}{0.7} & \vpm{1.35}{0.13} & \bvpm{30.3}{0.6} & \bvpm{12.3}{1.0} & \bvpm{6.0}{1.1} \\
    \midrule
    \multicolumn{8}{l}{\textit{\textbf{VQAv2 (ViLT)}}} \\
    Mean Network & 1.0$\times$ & \vpm{22.5}{0.2} & \vpm{37.7}{0.1} & \vpm{7.71}{0.29} & \vpm{24.7}{0.1} & \vpm{9.1}{0.5} & \vpm{5.4}{0.2} \\
    \quad + Temp. Sc. & 1.0$\times$ & \vpm{22.5}{0.2} & \vpm{37.7}{0.1} & \vpm{7.67}{0.28} & \vpm{24.7}{0.1} & \vpm{9.1}{0.5} & \vpm{5.4}{0.2} \\
    \hdashline
    2 samples & 2.0$\times$ & \vpm{20.4}{0.3} & \vpm{33.8}{0.2} & \vpm{2.88}{0.03} & \vpm{22.2}{0.2} & \vpm{9.4}{0.1} & \vpm{5.4}{0.3} \\
    4 samples & 4.0$\times$ & \vpm{22.6}{0.2} & \vpm{36.1}{0.1} & \bvpm{2.07}{0.02} & \vpm{24.1}{0.1} & \vpm{11.0}{0.5} & \vpm{7.6}{0.3} \\
    \hdashline
    \textcolor{Streamlining}{Streamlining} & 2.0$\times$ & \vpm{22.5}{0.2} & \vpm{37.7}{0.1} & \vpm{7.69}{0.29} & \vpm{24.7}{0.1} & \vpm{9.1}{0.5} & \vpm{5.4}{0.2} \\
    \hdashline
    \textcolor{CVP}{CVP} & 2.2$\times$ & \bvpm{24.7}{0.1} & \bvpm{38.4}{0.3} & \vpm{3.68}{0.13} & \bvpm{25.6}{0.3} & \bvpm{12.3}{0.2} & \bvpm{8.3}{0.4} \\
    \bottomrule
  \end{tabular}
\end{table*}

\begin{table}[ht]
  \centering
  \renewcommand{\arraystretch}{1.3}
  \setlength{\tabcolsep}{3.5pt}
  \footnotesize
  \caption{Macro-average over all six benchmarks. SEMs combined assuming independence across benchmarks.}
  \label{tab:default:avg_test_sem}
  \begin{tabular}{l c r r r r r r r}
    \toprule
    \multirow{2}{*}{Method} & Rel. & \multicolumn{1}{c}{\multirow{2}{*}{Acc.}} & \multicolumn{1}{c}{\multirow{2}{*}{AURC ($\downarrow$)}} & \multicolumn{1}{c}{\multirow{2}{*}{C@$\tfrac{1}{2}$\%}} & \multicolumn{1}{c}{\multirow{2}{*}{C@1\%}} & \multicolumn{1}{c}{\multirow{2}{*}{NLL / BCE ($\downarrow$)}} & \multicolumn{1}{c}{\multirow{2}{*}{ECE ($\downarrow$)}} & \multicolumn{1}{c}{\multirow{2}{*}{Brier ($\downarrow$)}} \\
     & Time &  &  &  &  &  &  &  \\
    \midrule
    Mean Network & 0.9$\times$ & \vpm{78.4}{0.0} & \vpm{6.39}{0.02} & \vpm{14.6}{0.6} & \vpm{25.9}{0.3} & \vpm{1.37}{0.00} & \vpm{5.42}{0.13} & \vpm{0.32}{0.00} \\
    \quad + Temp. Sc. & 0.9$\times$ & \vpm{78.4}{0.0} & \vpm{6.39}{0.02} & \vpm{14.9}{0.6} & \vpm{25.9}{0.3} & \vpm{1.35}{0.00} & \vpm{2.98}{0.05} & \vpm{0.32}{0.00} \\
    \hdashline
    2 samples & 2.0$\times$ & \vpm{77.6}{0.0} & \vpm{6.84}{0.03} & \vpm{16.7}{0.5} & \vpm{25.7}{0.4} & \vpm{1.38}{0.00} & \vpm{2.76}{0.11} & \vpm{0.33}{0.00} \\
    4 samples & 4.0$\times$ & \vpm{78.1}{0.0} & \vpm{6.52}{0.03} & \vpm{18.6}{0.5} & \vpm{27.4}{0.3} & \vpm{1.36}{0.00} & \vpm{2.16}{0.10} & \vpm{0.32}{0.00} \\
    \hdashline
    \textcolor{Streamlining}{Streamlining} & 2.2$\times$ & \vpm{78.4}{0.0} & \vpm{6.39}{0.03} & \vpm{15.2}{0.6} & \vpm{25.5}{0.4} & \vpm{1.35}{0.00} & \vpm{3.06}{0.06} & \vpm{0.32}{0.00} \\
    \hdashline
    \textcolor{CVP}{CVP} & 2.3$\times$ & \bvpm{78.8}{0.0} & \bvpm{6.16}{0.02} & \bvpm{19.8}{0.5} & \bvpm{29.2}{0.2} & \bvpm{1.32}{0.00} & \bvpm{2.06}{0.04} & \bvpm{0.31}{0.00} \\
    \bottomrule
  \end{tabular}
\end{table}

\begin{table}[ht]
  \centering
  \renewcommand{\arraystretch}{1.3}
  \setlength{\tabcolsep}{3.5pt}
  \footnotesize
  \caption{Macro-average over all six benchmarks. SEMs combined assuming independence across benchmarks.}
  \label{tab:default_extra:avg_test_sem}
  \begin{tabular}{l c r r r r r r}
    \toprule
    \multirow{2}{*}{Method} & Rel. & \multicolumn{1}{c}{\multirow{2}{*}{C@2\%}} & \multicolumn{1}{c}{\multirow{2}{*}{C@5\%}} & \multicolumn{1}{c}{\multirow{2}{*}{ACE ($\downarrow$)}} & \multicolumn{1}{c}{\multirow{2}{*}{$\Phi_{10}$}} & \multicolumn{1}{c}{\multirow{2}{*}{$\Phi_{50}$}} & \multicolumn{1}{c}{\multirow{2}{*}{$\Phi_{100}$}} \\
     & Time &  &  &  &  &  &  \\
    \midrule
    Mean Network & 0.9$\times$ & \vpm{38.3}{0.3} & \vpm{57.1}{0.2} & \vpm{5.36}{0.12} & \vpm{33.6}{0.1} & \vpm{14.4}{0.3} & \vpm{8.2}{0.4} \\
    \quad + Temp. Sc. & 0.9$\times$ & \vpm{38.2}{0.3} & \vpm{57.1}{0.2} & \vpm{2.96}{0.06} & \vpm{33.6}{0.1} & \vpm{14.4}{0.3} & \vpm{7.9}{0.4} \\
    \hdashline
    2 samples & 2.0$\times$ & \vpm{36.7}{0.3} & \vpm{54.3}{0.2} & \vpm{2.72}{0.11} & \vpm{32.1}{0.2} & \vpm{14.4}{0.3} & \vpm{8.4}{0.4} \\
    4 samples & 4.0$\times$ & \vpm{38.5}{0.2} & \vpm{56.0}{0.2} & \vpm{2.11}{0.10} & \vpm{33.3}{0.1} & \vpm{15.7}{0.3} & \vpm{9.7}{0.3} \\
    \hdashline
    \textcolor{Streamlining}{Streamlining} & 2.2$\times$ & \vpm{38.3}{0.2} & \vpm{57.0}{0.2} & \vpm{3.06}{0.06} & \vpm{33.7}{0.2} & \vpm{14.4}{0.3} & \vpm{8.2}{0.4} \\
    \hdashline
    \textcolor{CVP}{CVP} & 2.3$\times$ & \bvpm{40.7}{0.2} & \bvpm{58.1}{0.1} & \bvpm{2.08}{0.04} & \bvpm{34.7}{0.1} & \bvpm{17.0}{0.2} & \bvpm{10.7}{0.3} \\
    \bottomrule
  \end{tabular}
\end{table}

\section{Extended Ablation Study}\label{sec:app_ablations}

We present the additional ablation results using macro-averages acorss the six setups (\cf \Cref{tab:ablations:avg_test_sem} from the main paper) in \Cref{tab:ablations_extra:avg_test_sem}.

\begin{table}[ht]
  \centering
  \renewcommand{\arraystretch}{1.3}
  \setlength{\tabcolsep}{3.5pt}
  \footnotesize
  \caption{Ablation study: macro-average over all six benchmarks. SEMs combined assuming independence across benchmarks.}
  \label{tab:ablations_extra:avg_test_sem}
  \begin{tabular}{l r r r r r r}
    \toprule
    \multirow{2}{*}{Method} & \multicolumn{1}{c}{\multirow{2}{*}{C@2\%}} & \multicolumn{1}{c}{\multirow{2}{*}{C@5\%}} & \multicolumn{1}{c}{\multirow{2}{*}{ACE ($\downarrow$)}} & \multicolumn{1}{c}{\multirow{2}{*}{$\Phi_{10}$}} & \multicolumn{1}{c}{\multirow{2}{*}{$\Phi_{50}$}} & \multicolumn{1}{c}{\multirow{2}{*}{$\Phi_{100}$}} \\
     &  &  &  &  &  &  \\
    \midrule
    \textcolor{CVP}{CVP} & \bvpm{40.4}{0.1} & \bvpm{57.9}{0.1} & \vpm{2.22}{0.04} & \bvpm{34.6}{0.1} & \bvpm{16.2}{0.2} & \bvpm{10.1}{0.3} \\
    \quad $-$ Normalization & \vpm{38.7}{0.2} & \vpm{57.1}{0.2} & \bvpm{2.09}{0.04} & \vpm{33.8}{0.1} & \vpm{14.4}{0.3} & \vpm{8.3}{0.3} \\
    \quad $-$ Exact Activation & \vpm{39.8}{0.1} & \vpm{57.4}{0.2} & \vpm{3.55}{0.10} & \vpm{34.4}{0.1} & \vpm{16.0}{0.2} & \vpm{9.2}{0.4} \\
    \quad $-$ Per-Layer Calib. & \vpm{38.3}{0.2} & \vpm{56.5}{0.2} & \vpm{2.22}{0.05} & \vpm{33.3}{0.3} & \vpm{14.8}{0.3} & \vpm{8.1}{0.4} \\
    \quad $-$ All Calib. & \vpm{38.6}{0.2} & \vpm{56.6}{0.3} & \vpm{4.01}{0.11} & \vpm{33.5}{0.2} & \vpm{15.6}{0.3} & \vpm{9.0}{0.4} \\
    \hdashline
    \textcolor{Streamlining}{Streamlining} & \vpm{38.0}{0.3} & \vpm{57.1}{0.2} & \vpm{2.86}{0.05} & \vpm{33.7}{0.2} & \vpm{14.2}{0.3} & \vpm{8.2}{0.3} \\
    \quad + Per-Layer Calib. & \vpm{39.1}{0.1} & \vpm{57.2}{0.2} & \vpm{2.85}{0.06} & \vpm{33.9}{0.1} & \vpm{15.1}{0.2} & \vpm{8.3}{0.3} \\
    \quad $-$ All Calib. & \vpm{38.3}{0.3} & \vpm{57.1}{0.2} & \vpm{5.18}{0.12} & \vpm{33.7}{0.1} & \vpm{14.5}{0.3} & \vpm{8.1}{0.4} \\
    \bottomrule
  \end{tabular}
\end{table}

\section{LayerNorm approximation}\label{sec:app_layernorm}

\begin{figure}[!thbp]
    \centering
    \begin{subfigure}[t]{\textwidth}
        \centering
        \includegraphics[width=\textwidth]{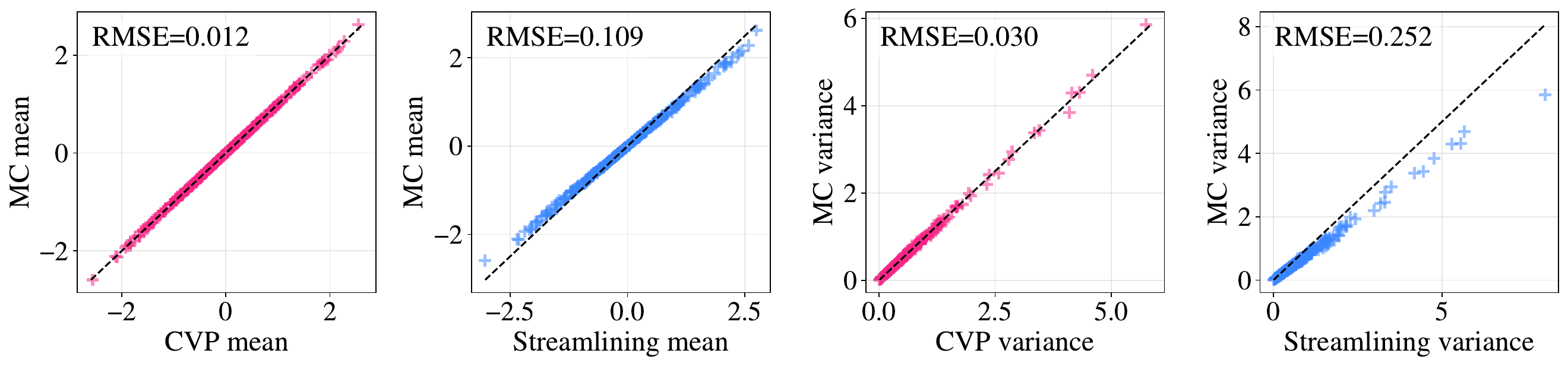}
        \caption{Synthetic Data with $\mu_x \sim \mathcal{N}(0, I)$ and $\Sigma_x\sim\mathcal{U}(0, 0.5)$.}
    \end{subfigure}
    \hfill
    \begin{subfigure}[t]{\textwidth}
        \centering
        \includegraphics[width=\textwidth]{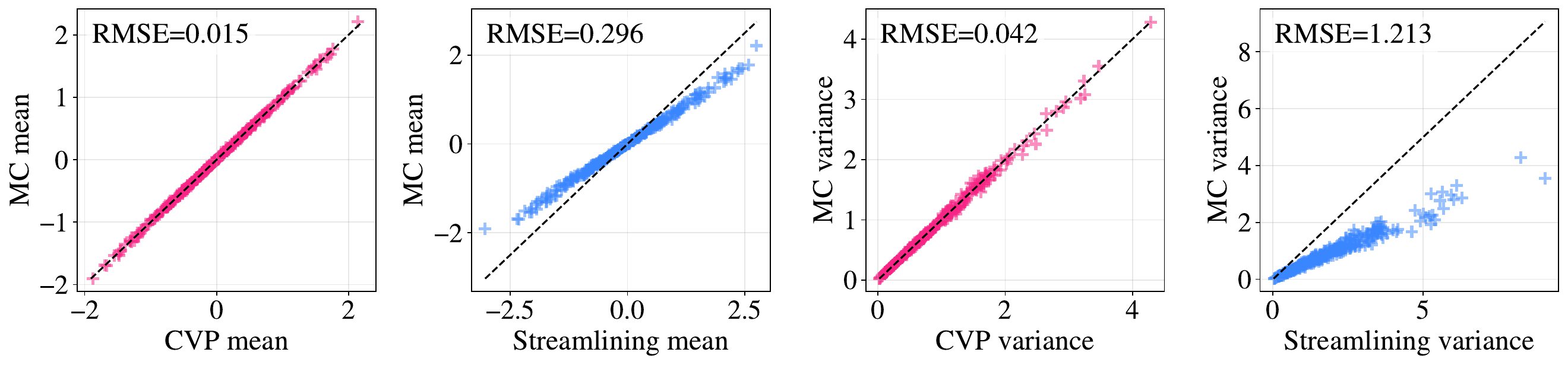}
        \caption{Synthetic Data with $\mu_x \sim \mathcal{N}(0, I)$ and $\Sigma_x\sim\mathcal{U}(0, 2)$.}
    \end{subfigure}
    \hfill
    \begin{subfigure}[t]{\textwidth}
        \centering
        \includegraphics[width=\textwidth]{new_figures/layernorm/scatter_iv50_wv1.pdf}
        \caption{Synthetic Data with $\mu_x \sim \mathcal{N}(0, I)$ and $\Sigma_x\sim\mathcal{U}(0, 5)$.}
    \end{subfigure}
    \hfill
    \begin{subfigure}[t]{\textwidth}
        \centering
        \includegraphics[width=\textwidth]{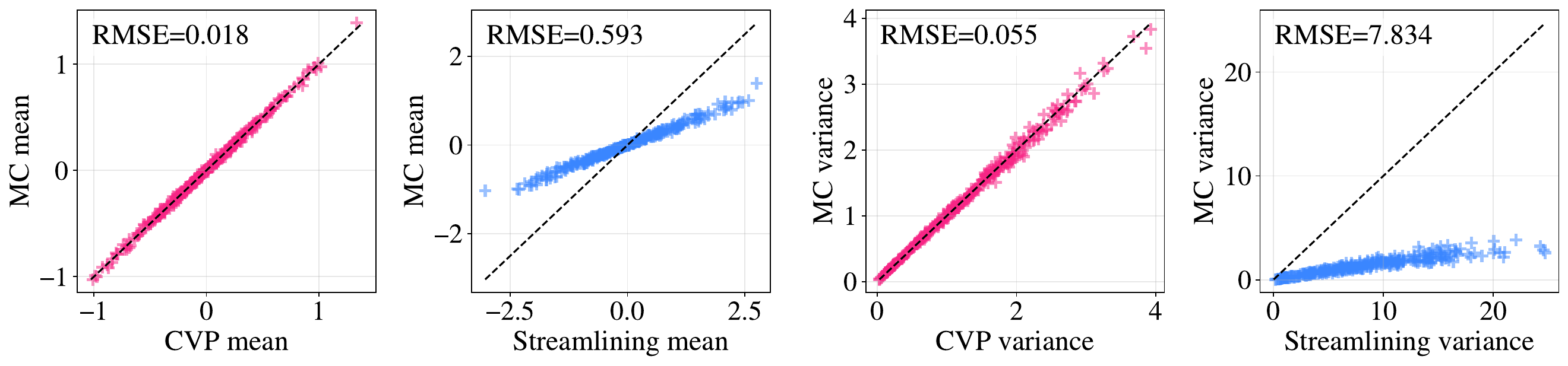}
        \caption{Synthetic Data with $\mu_x \sim \mathcal{N}(0, I)$ and $\Sigma_x\sim\mathcal{U}(0, 10)$.}
    \end{subfigure}

    \caption{Synthetic Data: With increasing input variance, Streamlining's propagated mean and variance through LayerNorm deviate more and more from the MC Sampling ground truth (variance and absolute value of the mean are overestimated), whereas our approximation (\cf \cref{eq:method_layernorm_s2}) shows only negligible deviations. Every scattered dot represents an activation.}
    \label{fig:layernorm_toy_all}
\end{figure}

\begin{figure}[!thbp]

    \begin{subfigure}[t]{0.9\textwidth}
        \centering
        \includegraphics[width=\textwidth]{new_figures/layernorm/cifar100_pooled.pdf}
        \label{fig:ln_real_cifar100}
        \caption{CIFAR-100 (ViT-Base)}
    \end{subfigure}
    \begin{subfigure}[t]{0.9\textwidth}
        \centering
        \includegraphics[width=\textwidth]{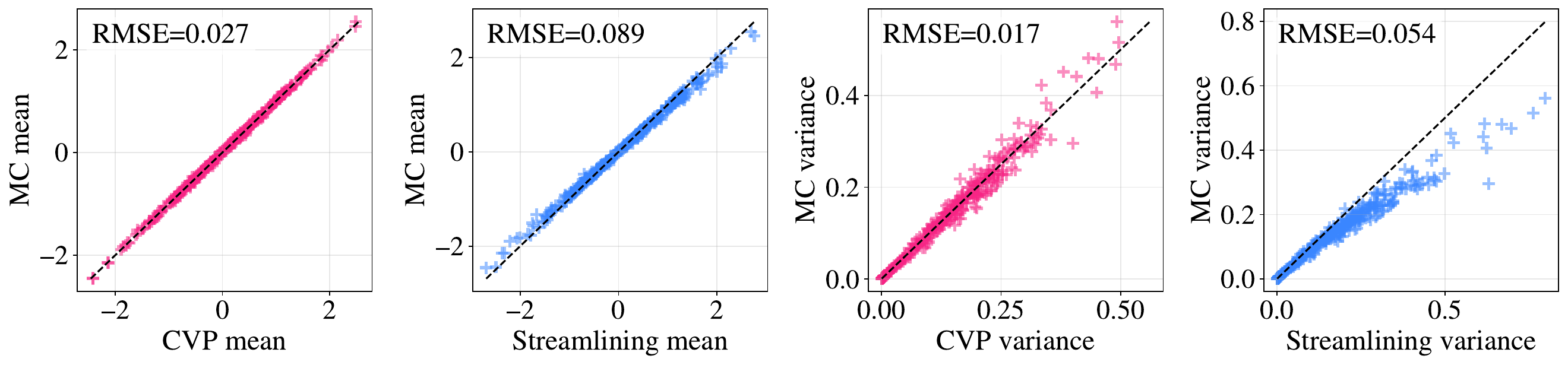}
        \label{fig:ln_real_cifar10}
        \caption{CIFAR-10 (sub-tiny ViT)}
    \end{subfigure}
    \begin{subfigure}[t]{0.9\textwidth}
        \centering
        \includegraphics[width=\textwidth]{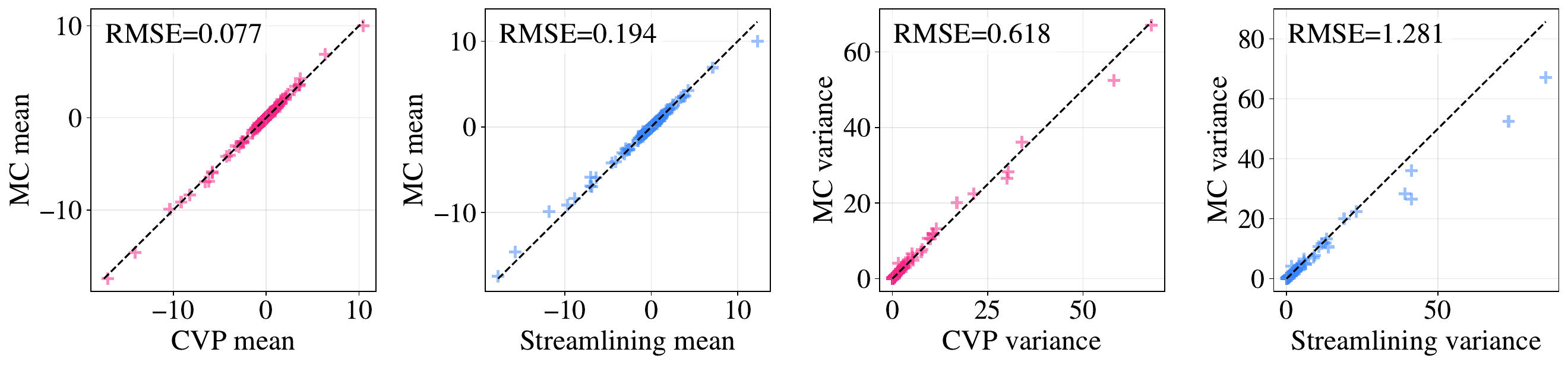}
        \label{fig:ln_real_vqabeit}
        \caption{VQAv2 (BEiT-3-base)}
    \end{subfigure}
    \begin{subfigure}[t]{0.9\textwidth}
        \centering
        \includegraphics[width=\textwidth]{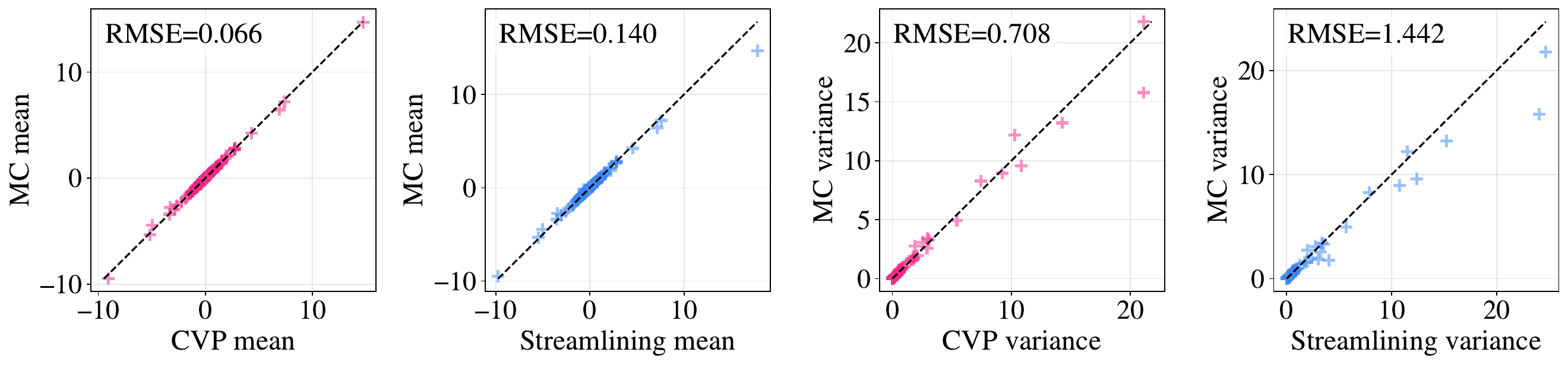}
        \label{fig:ln_real_nlvr2}
        \caption{NLVR2 (BEiT-3-base)}
    \end{subfigure}
    \begin{subfigure}[t]{0.9\textwidth}
        \centering
        \includegraphics[width=\textwidth]{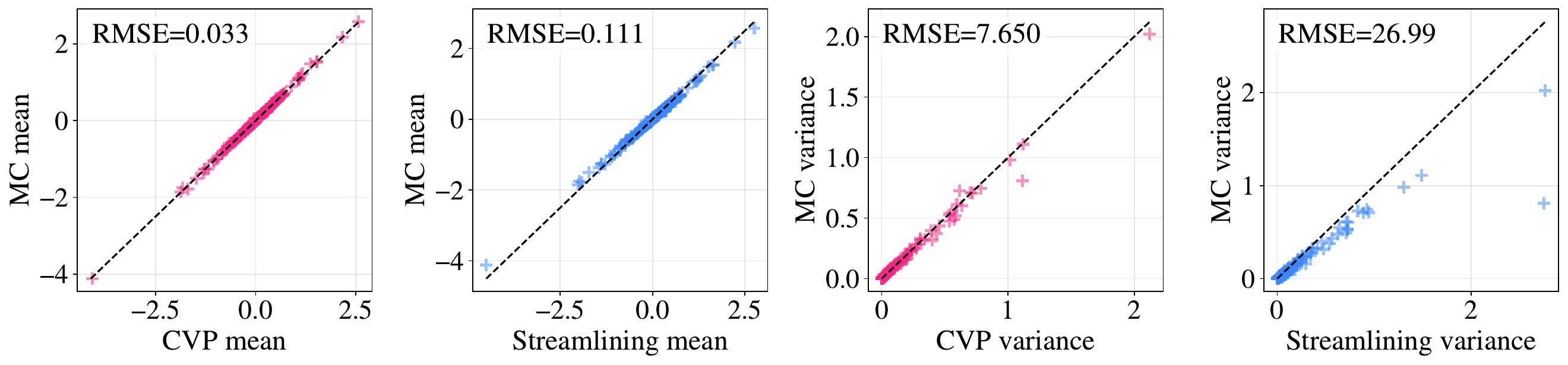}
        \label{fig:ln_real_vqavilt}
        \caption{VQAv2 (ViLT)}
    \end{subfigure}

    \caption{Real Data: Across all models, our corrected LayerNorm approximation shows consistent RSME reduction vs. Streamlining's linearized implementation. Every scattered dot represents an activation.}
    \label{fig:layernorm_real_all}
\end{figure}

We show extended results from our LayerNorm experiments (\cf \cref{sec:exp_layernorm} in the main paper). \Cref{fig:layernorm_toy_all} extends the toy experiments and \Cref{fig:layernorm_real_all} shows the real data from more settings.

\end{document}